%% file: emnlp-ijcnlp-2019.tex
\documentclass[11pt,a4paper]{article}
\usepackage[hyperref]{emnlp-ijcnlp-2019}
\usepackage{times}
\usepackage{latexsym}

\usepackage{url}

\usepackage{verbatim}
\usepackage{hyperref}
\usepackage{url}
\usepackage{multirow}
\usepackage{graphicx}
\usepackage{color,soul}
\usepackage{enumitem}
\usepackage{booktabs}       % professional-quality tables
\usepackage{subcaption}
\usepackage{appendix}
\usepackage{amsmath,amsfonts,bm}
\usepackage{float}
\usepackage{bibentry}

\aclfinalcopy % Uncomment this line for the final submission

\title{%How practical is active learning for NLP?
Practical Obstacles to Deploying Active Learning
}

\author{David Lowell \\
Northeastern University \\
  {\tt lowell.d@husky.neu.edu} \\\And
  Zachary C. Lipton \\
Carnegie Mellon University \\
  {\tt zlipton@cmu.edu} \\\AND
  Byron C. Wallace \\
Northeastern University \\
  {\tt b.wallace@northeastern.edu} \\}

\date{}

\begin{document}
\maketitle

\begin{abstract}

\input{sections/abstract.tex}
\end{abstract}

\section{Introduction}
\label{sec:intro}
\input{sections/intro.tex}

%\section{Empirical Study}
\section{Experimental Questions and Setup}
\label{section:experimental-setup}
\input{sections/experiments.tex}

\section{Tasks}
\label{sec:tasks}
\input{sections/tasks.tex}

\section{Results}
\label{sec:results}
\input{sections/results.tex}

\section{Discussion}
\label{sec:discussion}
\input{sections/discussion.tex}

\section{Conclusions}
\label{sec:conclusion}
\input{sections/conclusions.tex}

\section{Acknowledgements}

This work was supported in part by the Army Research Office (ARO), award W911NF1810328.

%\nobibliography{emnlp-ijcnlp-2019}
\bibliography{emnlp-ijcnlp-2019}
\bibliographystyle{acl_natbib}

\clearpage
\newpage
\appendix
\appendixpage
\setcounter{page}{1}

\section{\textbf{Experimental Results}} 
\input{sections/appendix_b.tex}

\end{document}

%% file: sections/abstract.tex
%
%  SETUP: What's the problem?
%
Active learning (AL) is a widely-used training strategy 
for maximizing predictive performance subject to a fixed annotation budget. 
In AL one iteratively selects training examples for annotation, 
% typically as a function of some measure of uncertainty.
often those for which the current model is most uncertain (by some measure).
The hope is that active sampling leads to better performance 
than would be achieved under independent and identically distributed (i.i.d.) random samples. 
%training a model with strong predictive performance faster than one would be able to via random (i.i.d.) sampling. 
%; this couples the acquired dataset with the underlying model.
%However, owing to the high cost of annotation and the rapid pace of model development, labeled datasets may remain valuable  long after a particular model is surpassed by new technology.
%
%  Punch: What we do
% bcw -- again watch consistency w/math-mode v not (A v $A$)
%
While AL has shown promise in retrospective evaluations, 
these studies often ignore practical obstacles to its use. % are often ignored. 
%In this paper we consider practical issues in applying AL.
In this paper we show that while AL may provide benefits %functions may afford better than i.i.d. performance 
when used with specific models and for particular domains, 
the benefits of current approaches do not generalize reliably across models and tasks. 
This is problematic because in practice  % (this would require i.i.d. sampled data to begin with).
one does not have the opportunity to explore 
and compare alternative AL strategies.
Moreover, AL couples the training dataset 
with the model used to guide its acquisition.
We find that subsequently training a \emph{successor model} 
with an actively-acquired dataset does not consistently outperform training on i.i.d. sampled data.
% training on a commensurate i.i.d. sample.
%Further, we note that active learning couples the acquired dataset with the model used in its generation.
%the transferability of datasets
%collected with an \emph{acquisition model} $A$
%to a distinct \emph{successor model} $S$.
%We seek to characterize whether the benefits of active learning
%persist when $A$ and $S$ are different models. 
%To this end, we consider two standard NLP tasks 
%and associated datasets: 
%text classification and sequence tagging. 
%We find that training $S$ on a dataset 
%actively acquired with a (different) model $A$ 
%often yields worse performance 
%than when $S$ is trained with ``native'' data 
%(i.e., acquired actively using $S$), 
%and often performs worse than training on i.i.d. sampled data. 
%These findings have implications 
%for the use of active learning in practice,
%suggesting that it is better suited to cases 
%where models are updated no more frequently than labeled data. 
%We find that using such a dataset to train a different model often yields worse performance than that obtained by training on i.i.d. sampled data. 
Our findings raise the question of whether the downsides inherent to AL
% (acquisition of a non-i.i.d. training sample)
are worth the modest and inconsistent performance gains it tends to afford.
%These findings have implications for the use of active learning in practice, 
%suggesting that it may not be generally safe to employ. 

%% file: sections/intro.tex
% bcw -- so the intro is really long... but i don't know that i have a better suggestion
% bcw -- addendum: I have split into two sections. This is... a solution.
% Modern machine learning methods require
% large amounts of labeled data. %to work well. 
Although deep learning now achieves state-of-the-art results 
on a number of supervised learning tasks 
\citep{Johnson:2016:SST:3045390.3045447, C18-1161}, 
realizing these gains requires large annotated datasets \citep{shen2018deep}. 
% bcw: we should cite more nlp SOTA stuff for DL; here i added BERT, but actually kind of counter intuitive, since that is sort of a mechanism for transfer... suggest adding some cites from here to SOTA papers that use DL: https://github.com/sebastianruder/NLP-progress 
This data dependence is problematic because labels are expensive.
Several lines of research seek to reduce the amount of supervision required 
to achieve acceptable predictive performance, 
including semi-supervised \citep{chapelle2009semi}, transfer \citep{pan2010survey}, 
and \emph{active learning} (AL) \citep{cohn1996active, settles2012active}.
% ;here we focus on the latter.

\begin{figure*}
  \centering
  \begin{subfigure}[t]{0.49\linewidth} % width of left subfigure
		\includegraphics[width=\linewidth]{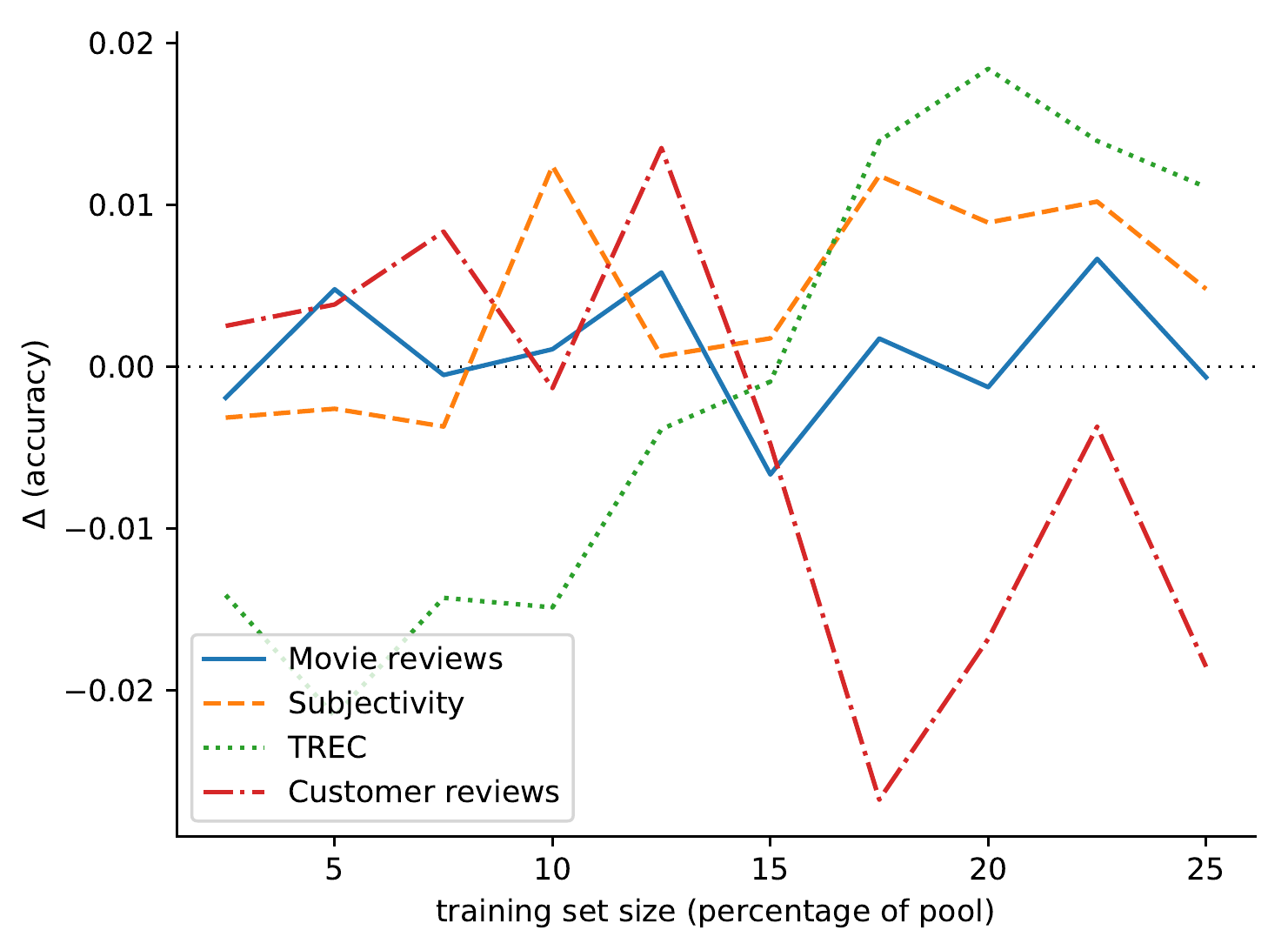}
		\caption{Performance of AL relative to i.i.d. across corpora.} 
		\label{fig:punchline_delta}
  \end{subfigure}
  \begin{subfigure}[t]{0.49\linewidth} % width of left subfigure
		\includegraphics[width=\linewidth]{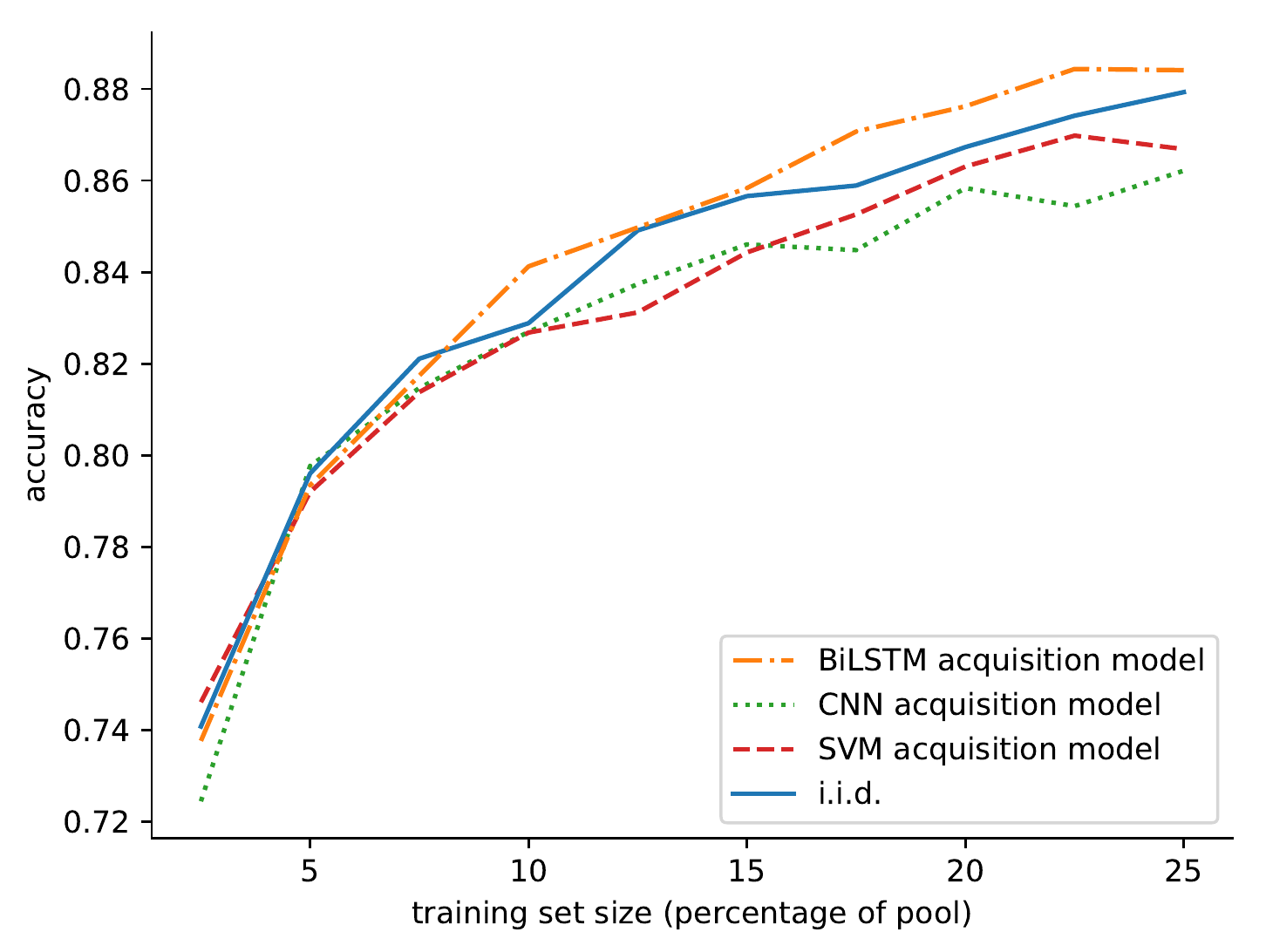} 
		\caption{Transferring actively acquired training sets.} % subcaption
		\label{fig:punchline_transfer}
  \end{subfigure}
  \caption{We highlight practical issues in the use of AL. (a) AL yields inconsistent gains, relative to a baseline of i.i.d. sampling, across corpora. (b) Training a BiLSTM with training sets actively acquired based on the uncertainty of other models tends to result in worse performance than training on i.i.d. samples. }
  %bcw:  this caption is way, way too long
  \begin{comment}
  \caption{An example plot of the delta in accuracy between that achieved using an actively acquired training set and that achieved using commensurate i.i.d. training data.
		This plot shows results for a BiLSTM using maximum entropy sampling.
		Positive delta indicates performance greater than that of i.i.d. data, negative delta indicates performance less than that of i.i.d. data.
		For the subjectivity data set (yellow, dashed line), active learning consistently outperforms i.i.d.
		For the customer reviews data set (red, dashed-dotted line), active learning outperforms i.i.d. in early iterations, but subsequently falls behind.
		In the TREC data set (green, dotted line), the situation is reversed, with active learning initially performing poorly, but overtaking i.i.d. data later.
		The movie reviews data set ( blue, solid line) shows equivocal results. An example learning curve for a BiLSTM model trained to perform text classification using the Subjectivity corpus. 
		Active learning (yellow, dashed-dotted line) outperforms training on i.i.d. data (blue, solid line).
		Training on data actively acquired by a CNN (green, dotted line) or a SVM (red, dashed line) yields performance worse than i.i.d.
		All active learning is performed using maximum entropy sampling.} \end{comment}
  \label{fig:punchline}
\end{figure*}

In AL, rather than training on a set of labeled data 
sampled at i.i.d. random from some larger population, 
the learner engages the annotator in a cycle of learning, 
iteratively selecting training data for annotation and updating its model.
\emph{Pool-based} AL (the variant we consider) proceeds in rounds. 
In each, the learner applies a heuristic to score unlabeled instances,
selecting the highest scoring instances for annotation.\footnote{This may be done either deterministically, by selecting the top-$k$ instances, or stochastically, selecting instances with probabilities proportional to heuristic scores.}
%Because we often have access to far more unlabeled data than we can afford to annotate, it is hoped that by selecting especially informative examples, 
%
% ZL:  The intuition is that
Intuitively,
by selecting training data cleverly, 
an active learner might achieve greater predictive performance than it would by choosing examples at random.

% bcw: i don't think we want to mathmode "i.i.d"
The more informative samples come at the cost of 
violating the standard i.i.d. assumption upon which 
supervised machine learning typically relies.
In other words, the training and test data no longer reflect
the same underlying data distribution. 
Empirically, AL has been found to work well with a variety of tasks and models 
\citep{settles2012active, Ramirez-Loaiza2017, gal2017deep, zhang2017active, shen2018deep}. 
However, academic investigations of AL
typically omit key real-world considerations that 
might overestimate its utility.
For example, once a dataset is actively acquired with one model,
it is seldom investigated whether this training sample will confer benefits
if used to train a second model (vs i.i.d. data).
% as compared to an i.i.d. training dataset.
Given that datasets often outlive learning algorithms,
this is an important practical consideration. 
% bcw: weird to not say at least *something* about this in main intro, since it's one of the main points of the paper?

In contrast to experimental (retrospective) studies, in a real-world setting,
an AL practitioner is not afforded the opportunity 
to retrospectively analyze or alter their scoring function. 
One would instead need to expend significant resources
to validate that a given scoring function performs 
as intended for a particular model and task. 
This would require i.i.d. sampled data to evaluate 
the comparative effectiveness of different AL strategies. 
However, collection of such additional data would defeat the purpose of AL, 
i.e., obviating the need for a large amount of supervision.
To confidently use AL in practice, one must have a reasonable belief 
that a given AL scoring (or \emph{acquisition}) function 
will produce the desired results \emph{before they deploy it} \cite{attenberg2011inactive}.

Most AL research does not explicitly characterize the circumstances under which AL may be expected to perform well. 
Practitioners must therefore make the implicit assumption 
that a given active acquisition strategy 
is likely to perform well under \emph{any} circumstances. 
Our empirical findings suggest that this assumption is not well founded and, 
in fact, common AL algorithms behave inconsistently across model types and datasets, 
often performing no better than random (i.i.d.) sampling (\ref{fig:punchline_delta}). 
Further, while there is typically \emph{some} AL strategy which outperforms i.i.d. random samples for a given dataset, \emph{which} heuristic varies. 

% DL - we have not yet mentioned transfer as an issue. I'm moving this paragraph to the end of section 2

%We note that prior work by Tomanek and Morik \shortcite{tomanek2011inspecting} also explored issues of actively acquired dataset transfer, although they did not share our broader focus on practical issues in AL. 
%Teural models and uses only one acquisition function per model. They conclude that transfer is usually successful, drawing primarily on good performance with the NER task. 
%Our study shows that this assumption is not generally correct.

% \vspace{.25em}
\noindent {\bf Contributions}. 
We highlight important but often overlooked issues in the use of AL in practice. 
We report an extensive set of experimental results on classification and sequence tagging tasks that suggest AL typically affords only marginal performance gains 
at the somewhat high cost of non-i.i.d. training samples, 
which do not consistently transfer well to subsequent models.

\section{The (Potential) Trouble with AL}
% DL - we need to trim some space and this reiterates stuff we just said
%AL aims to mitigate the amount of supervision required to realize acceptable predictive performance.
%However, as mentioned above, we find that particular active acquisition strategies realize this inconsistently across datasets.
%We illustrate this in Figure \ref{fig:punchline_delta}, which plots the relative gains ($\Delta$ to) achieved by a BiLSTM model using a maximum-entropy active sampling strategy, as compared to the same model trained with randomly sampled data.
We illustrate inconsistent comparative performance using AL.
Consider Figure \ref{fig:punchline_delta},
in which we plot the relative gains ($\Delta$) achieved by a BiLSTM model using a maximum-entropy active sampling strategy, 
as compared to the same model trained with randomly sampled data.
Positive values on the $y$-axis correspond to cases
in which AL achieves better performance than random sampling,
$0$ (dotted line) indicates no difference between the two, and negative values correspond to cases in which random sampling performs better than AL.
Across the four datasets shown, results are decidedly mixed.

And yet realizing these equivocal gains using AL brings inherent drawbacks.
For example, acquisition functions generally depend 
on the underlying model being trained \citep{settles.tr09,settles2012active},
which we will refer to as the \emph{acquisition model}. 
Consequently, the collected training data and the acquisition model are \emph{coupled}.
This coupling is problematic because manually labeled data 
tends to have a longer shelf life than models, 
largely because it is expensive to acquire.
However, progress in machine learning is fast. 
Consequently, in many settings, an actively acquired dataset 
may remain in use (much) longer than the source model used to acquire it. 
In these cases, a few natural questions arise:
How does a \emph{successor} model $S$ fare, 
when trained on data collected via an acquisition model $A$? 
How does this compare to training $S$ on natively acquired data? 
How does it compare to training $S$ on i.i.d. data?

For example, if we use uncertainty sampling 
under a support vector machine (SVM) to acquire a training set $\mathcal{D}$, 
and subsequently train a Convolutional Neural Network (CNN) using $\mathcal{D}$, 
will the CNN perform better than it would have 
if trained on a dataset acquired via i.i.d. random sampling? 
And how does it perform compared to using a training corpus 
actively acquired using the CNN? %as the acquisition model? 

Figure \ref{fig:punchline_transfer} shows results 
for a text classification example using the Subjectivity corpus \citep{Pang+Lee:04a}.
We consider three models: a Bidirectional Long Short-Term Memory Network (BiLSTM) \citep{hochreiter1997long}, a Convolutional Neural Network (CNN) \citep{kim2014convolutional,zhang2015sensitivity}, and a Support Vector Machine (SVM) \cite{joachims1998text}.
Training the LSTM  with a dataset 
actively acquired using either of the other models 
yields predictive performance that is \emph{worse} 
than that achieved under i.i.d. sampling. 
Given that datasets tend to outlast models, 
these results raise questions regarding the benefits of using AL in practice. 

We note that in prior work, \citet{tomanek2011inspecting} also explored 
the transferability of actively acquired datasets, 
although their work did not consider modern deep learning models
or share our broader focus on practical issues in AL. 

% bcw: but they sort of have, just not as extensively/explicitly i think?
%As far as we are aware, these questions---which have obvious practical implications---have been raised \cite{attenberg2011inactive} before  not previously been explored empirically.

%% file: sections/experiments.tex
We seek to answer two questions empirically: 
(1) How reliably does AL yield gains over sampling i.i.d.? And,
(2) What happens when we use a dataset actively acquired using one model 
to train a different (successor) model?
To answer these questions, we consider two tasks 
for which AL has previously been shown to confer considerable benefits:
text classification and sequence tagging (specifically NER).\footnote{Recent works have shown that AL is effective 
for these tasks even when using modern, 
% data-hungry 
neural architectures \citep{zhang2017active,shen2018deep}, 
but do not address our primary concerns 
regarding replicability and transferability.
% these did not explore the question of transferability.
}

% ZL:   This use of "those" is a bit weird 
%       we should just directly say deep networks
%       vs the weirdly elided "those representative of..."
%       Also all thse sentences start with "we" ... reworking below
% We consider both linear models and those more representative 
% of the current state-of-the-art for these tasks. 
% We investigate the standard strategy of acquiring data and training using a single model, 
% and also the case of acquiring data using one model 
% and subsequently using it to train a second model. 
% We frame this as considering all possible (acquisition, successor) pairs 
% among the considered models, where the standard case 
% is represented by the acquisition and successor model being the same. 
% For each pair $(A, S)$, we first simulate iterative active data acquisition
% with model $A$ to label a training dataset $\mathcal{D}_A$.
% We then train the successor model $S$ using $\mathcal{D}_A$. 

To build intuition, our experiments address both linear models 
and deep networks more representative 
of the current state-of-the-art for these tasks. 
We investigate the standard strategy of acquiring data
and training using a single model, 
and also the case of acquiring data using one model 
and subsequently using it to train a second model. 
Our experiments consider all possible (acquisition, successor) pairs 
among the considered models, such that the standard AL scheme
corresponds to the setting in which the acquisition and successor models are same. 
For each pair $(A, S)$, we first simulate iterative active data acquisition
with model $A$ to label a training dataset $\mathcal{D}_A$.
We then train the successor model $S$ using $\mathcal{D}_A$. 

In our evaluation, we compare the relative performance 
(accuracy or F1, as appropriate for the task) 
of the successor model trained with corpus $\mathcal{D}_A$ 
to the scores achieved by training on comparable amounts 
of native and i.i.d. sampled data. 
We simulate pool-based AL using labeled benchmark datasets 
by withholding document labels from the models. 
This induces a pool of unlabeled data $\mathcal{U}$. 
In AL, it is common to \emph{warm-start} the acquisition model, 
training on some modest amount of i.i.d. labeled data $\mathcal{D}_{w}$ 
before using the model to score candidates in $\mathcal{U}$ \citep{settles.tr09} 
and commencing the AL process. 
We follow this convention throughout.

Once we have trained the acquisition model on the warm-start data,
we begin the simulated AL loop, 
iteratively selecting instances for labeling 
and adding them to the dataset. 
We denote the dataset acquired by model $A$ 
at iteration $t$ by $\mathcal{D}_A^t$; 
$\mathcal{D}_A^0$ is initialized to $\mathcal{D}_{w}$ 
for all models (i.e., all values of $A$). 
At each iteration, the acquisition model is trained with $\mathcal{D}_A^t$. 
It then scores the remaining unlabeled documents 
in $\mathcal{U}\setminus \mathcal{D}_A^t$ 
according to a standard uncertainty AL heuristic. 
The top $n$ candidates $\mathcal{C}_A^t$ 
are selected for (simulated) annotation. 
Their labels are revealed and they are added to the training set: 
$\mathcal{D}_A^{t+1} \gets \mathcal{D}_A^t \cup \mathcal{C}_A^t$. 
At the experiment's conclusion (time step $T$), 
each acquisition model $A$ will have selected a (typically distinct) 
subset of $\mathcal{U}$ for training.

Once we have acquired datasets from each acquisition model $\mathcal{D}_A$, 
we evaluate the performance of each possible successor model 
when trained on $\mathcal{D}_A$.
Specifically, we train each successor model $S$ 
on the acquired data $\mathcal{D}_A^t$ 
for all $t$ in the range $[0,T]$, 
evaluating its performance on a held-out test set (distinct from $\mathcal{U}$).
We compare the performance achieved in this case
to that obtained using an i.i.d. training set of the same size.

We run this experiment ten times, 
averaging results to create summary learning curves, 
as shown in Figure \ref{fig:punchline}.
All reported results, including i.i.d. baselines, are averages of ten experiments, 
each conducted with a distinct $\mathcal{D}_w$.
These learning curves quantify the comparative performance of a particular model 
achieved using the same amount of supervision, 
but elicited under different acquisition models. 
For each model, we compare the learning curves of each acquisition strategy, 
including active acquisition using a \emph{foreign} model and subsequent transfer, 
active acquisition without changing models (i.e., typical AL), 
and the baseline strategy of i.i.d. sampling.

%We also observe relatively good performance on NER, but demonstrate that difficulties in the classification task have been previously undervalued.

\begin{figure*}
  \centering
  \begin{subfigure}[t]{0.32\linewidth} % width of left subfigure
		\includegraphics[width=\linewidth]{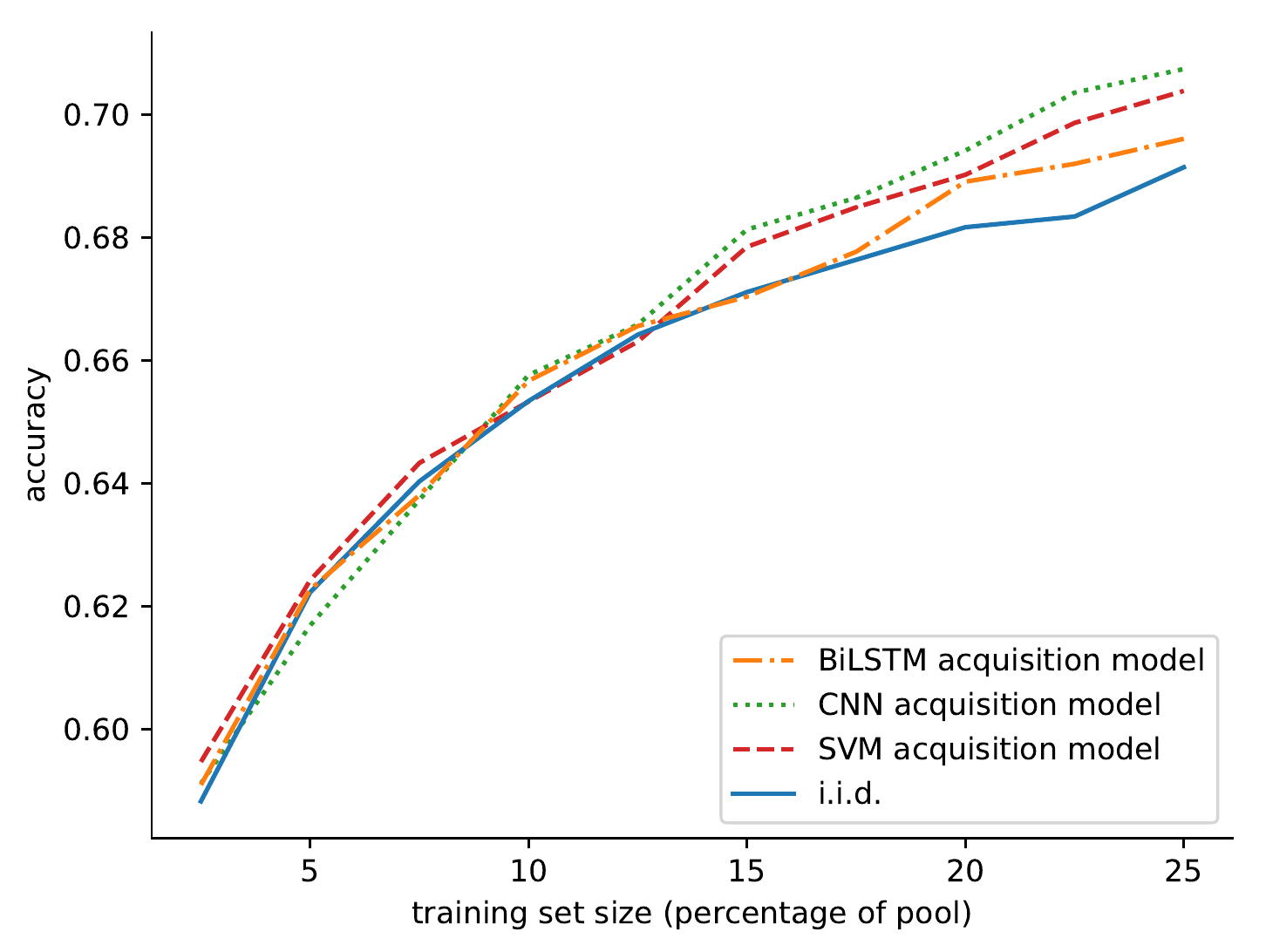}
		\caption{SVM on Movies dataset} % subcaption
  \end{subfigure}
  \begin{subfigure}[t]{0.32\linewidth} % width of left subfigure
		\includegraphics[width=\linewidth]{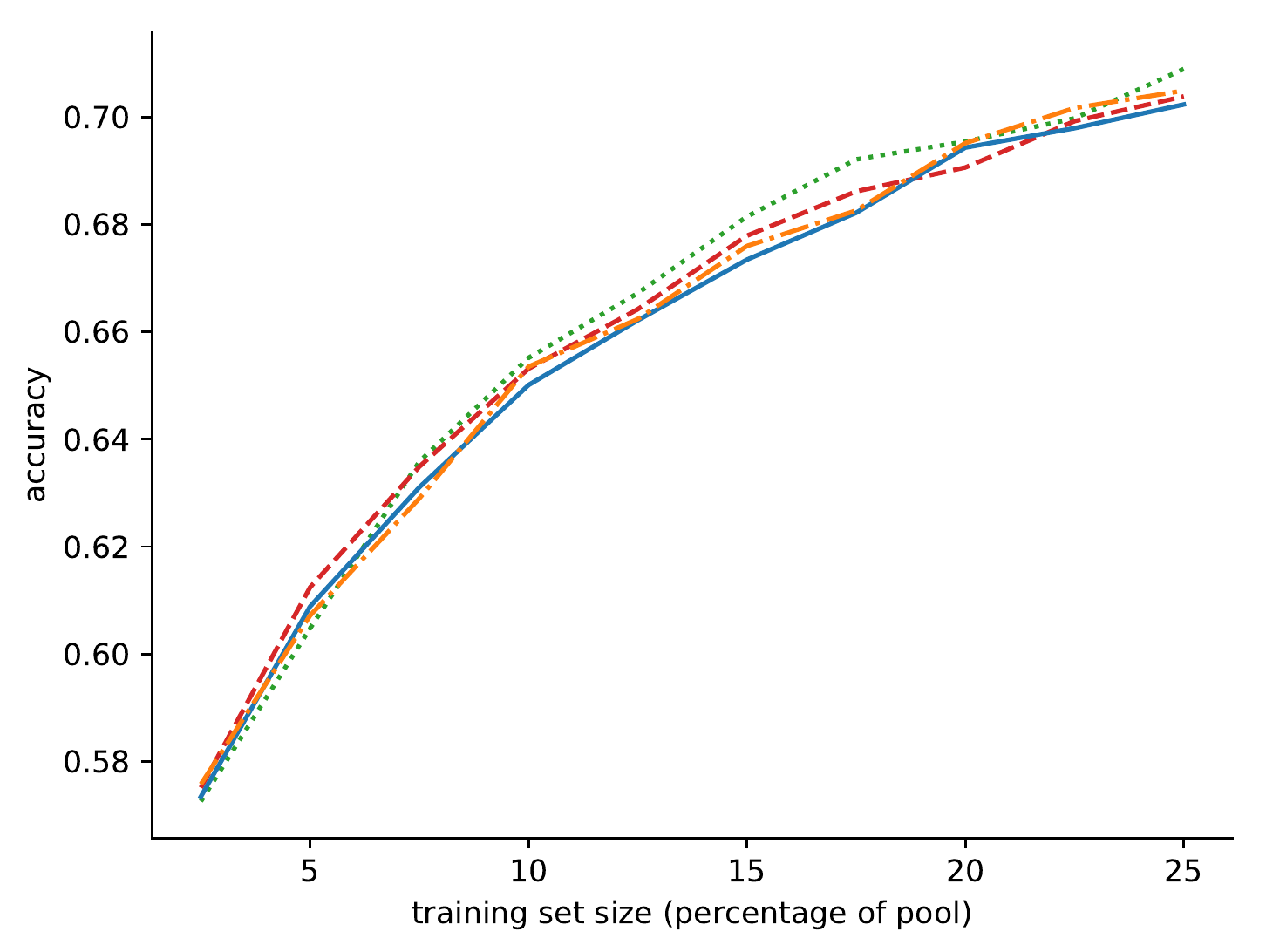}
		\caption{CNN on Movies dataset} % subcaption
  \end{subfigure}
  \begin{subfigure}[t]{0.32\linewidth} % width of left subfigure
		\includegraphics[width=\linewidth]{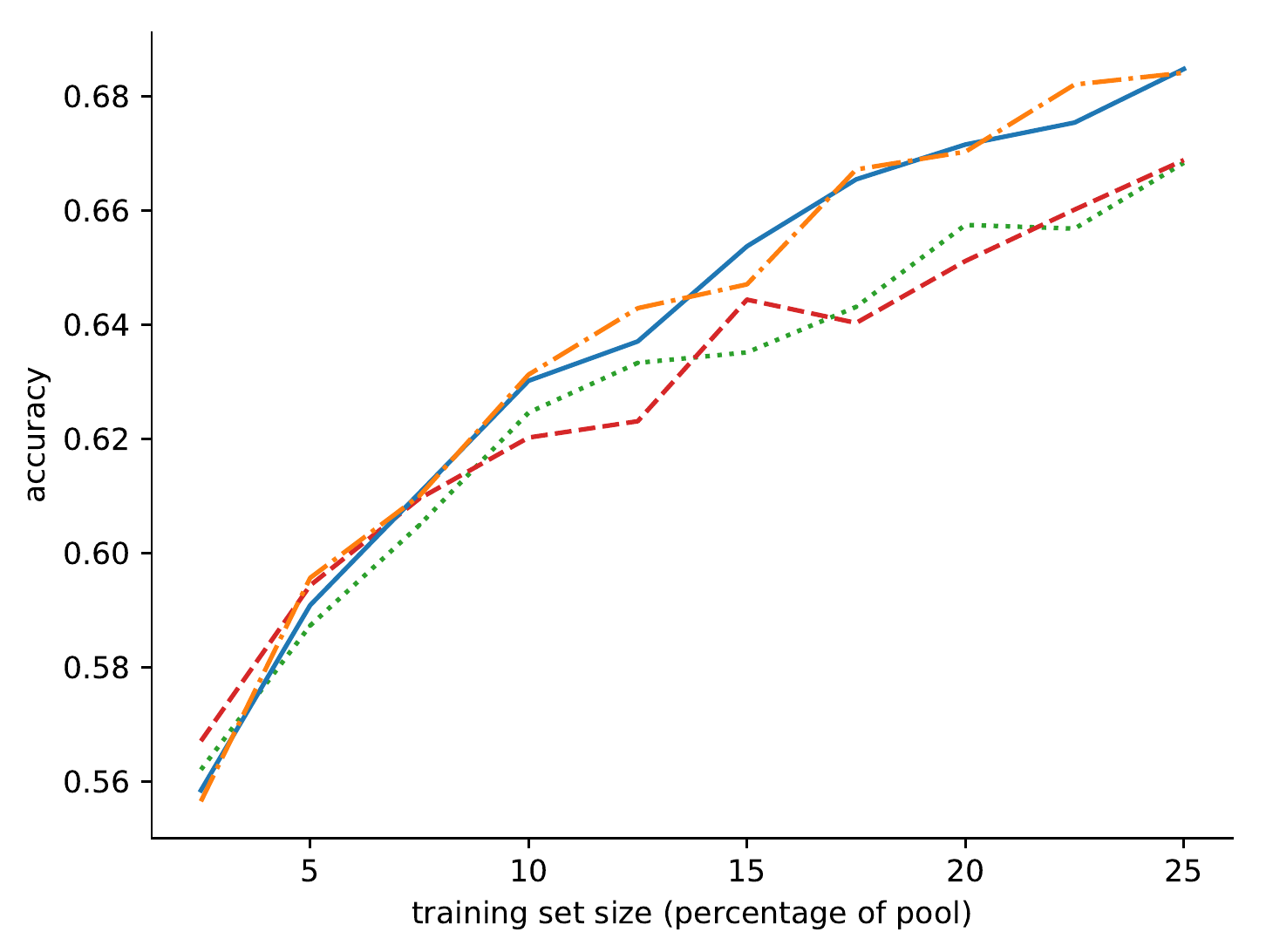}
		\caption{LSTM on Movies dataset} % subcaption
  \end{subfigure} \\
  \vspace{10px}
  \begin{subfigure}[t]{0.49\linewidth} % width of left subfigure
 \centering
\includegraphics[width=.66\linewidth]{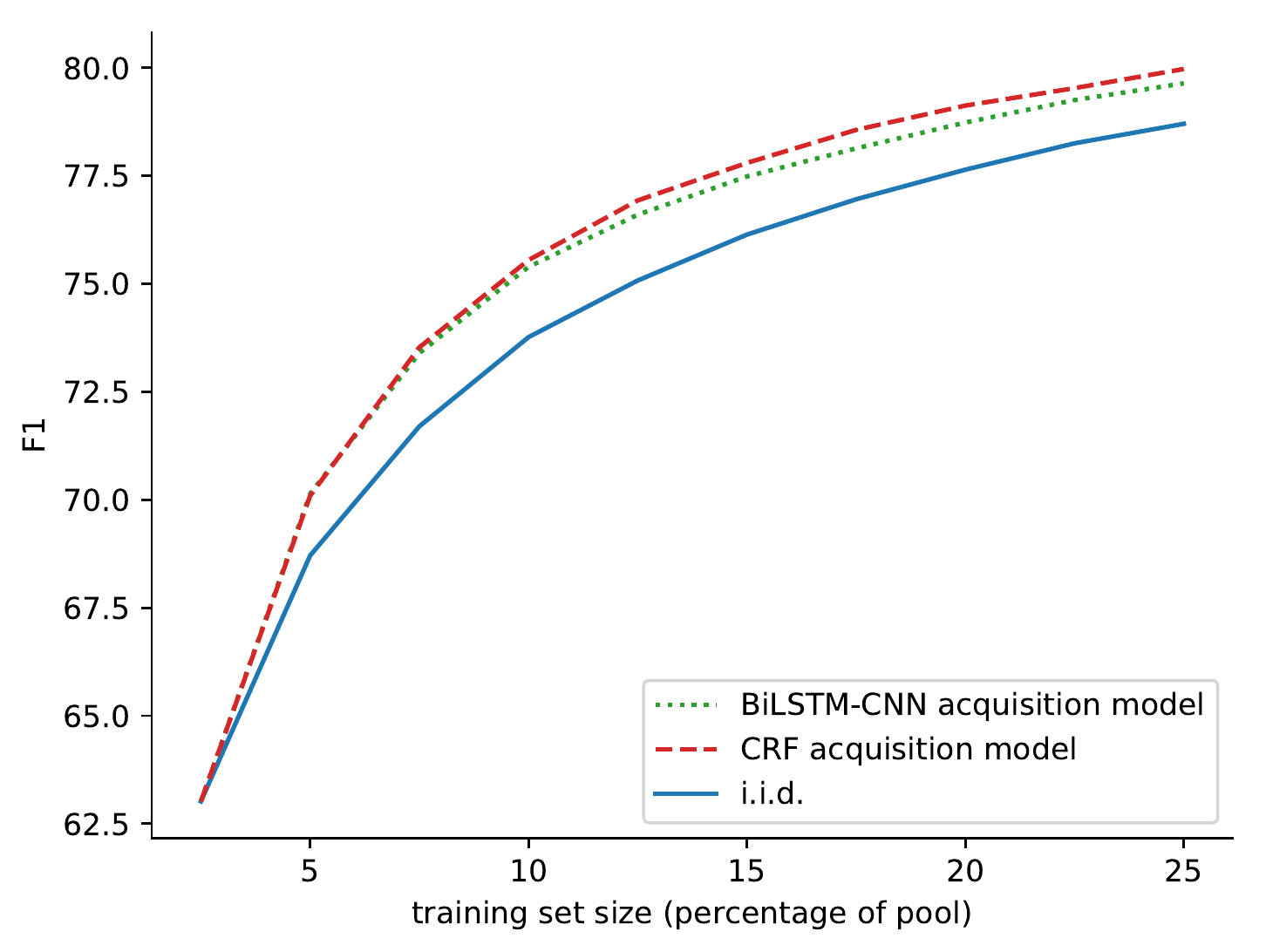}
		\caption{CRF on OntoNotes dataset} % subcaption
  \end{subfigure}
  \begin{subfigure}[t]{0.49\linewidth} % width of left subfigure
\centering
\includegraphics[width=.66\linewidth]{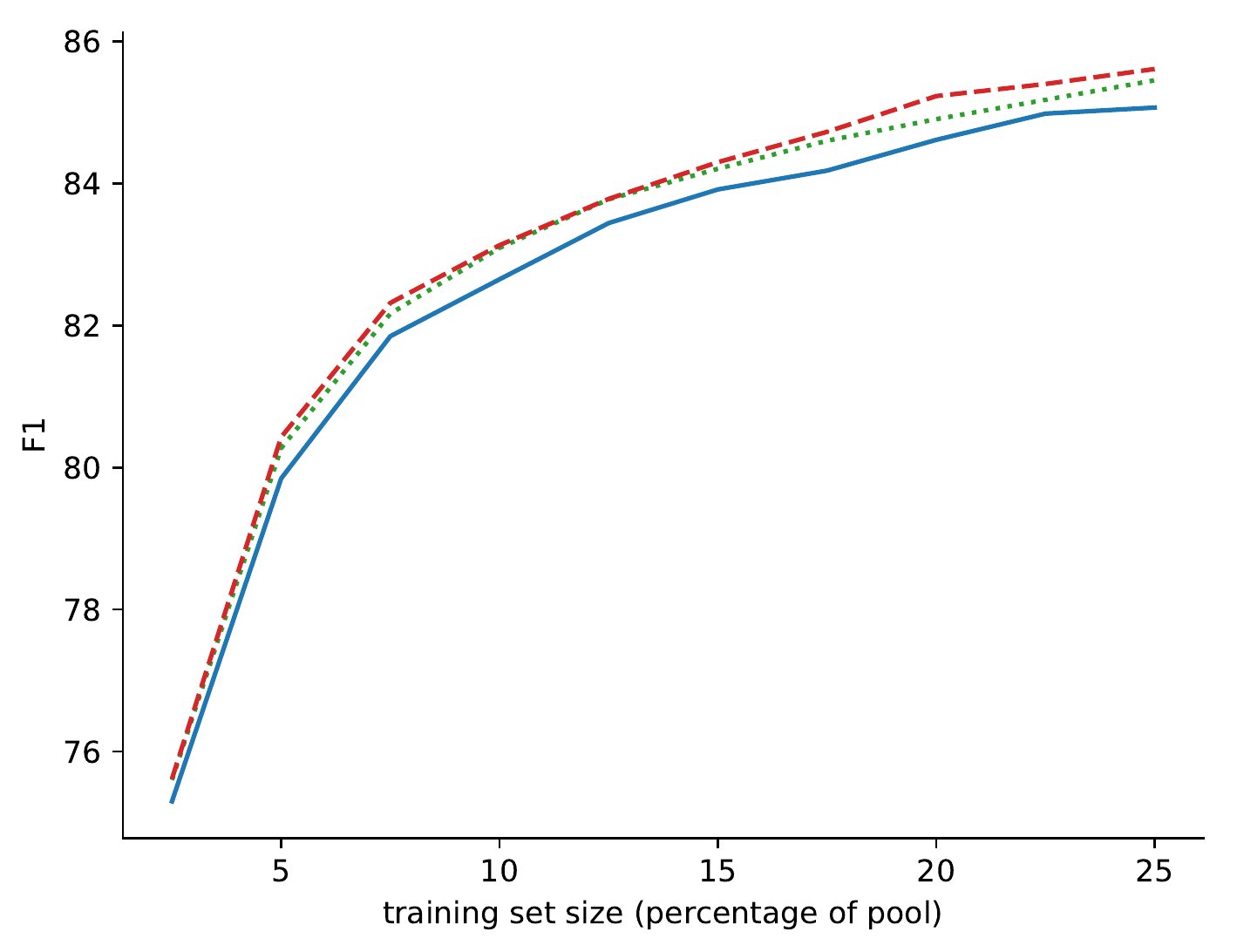}
		\caption{BiLSTM-CNN on OntoNotes dataset} % subcaption
\end{subfigure}
\caption{Sample learning curves for the text classification task on the Movie Reviews dataset and the NER task on the OntoNotes dataset using the maximum entropy acquisition function (we report learning curves for all models and datasets in the Appendix). Individual plots correspond to successor models. 
Each line corresponds to an acquisition model, 
with the blue line representing an i.i.d. baseline. 
}
\label{fig:curves}
\end{figure*}

\begin{table*}[ht!]
\centering
\begin{tabular}{l c c c c c c c c} 
 \multicolumn{9}{c}{\textbf{Text classification}} \\
 \toprule
 &&\multicolumn{6}{c }{{Acquisition model}} \\
& \multicolumn{4}{c}{10\% of pool} & \multicolumn{4}{c}{20\% of pool} \\
Successor & i.i.d. & SVM & CNN & LSTM &  i.i.d. & SVM & CNN & LSTM\\

%
% Movie reviews
%
\midrule
&\multicolumn{8}{c}{Movie reviews} \\
\midrule
SVM & 65.3 & \textbf{65.3} & \textcolor{blue}{65.8} & \textcolor{blue}{65.7} & 68.2 & \textcolor{blue}{\textbf{69.0}} & \textcolor{blue}{69.4} & \textcolor{blue}{68.9}\\
CNN & 65.0 & \textcolor{blue}{65.3} & \textcolor{blue}{\textbf{65.5}} & \textcolor{blue}{65.4} & 69.4 & \textcolor{red}{69.1} & \textcolor{blue}{\textbf{69.5}} & \textcolor{blue}{69.5}\\
LSTM & 63.0 & \textcolor{red}{62.0} & \textcolor{red}{62.5} & \textcolor{blue}{\textbf{63.1}} & 67.2 & \textcolor{red}{65.1} & \textcolor{red}{65.8} & \textcolor{red}{\textbf{67.0}}\\
%
% Subjectivity
%
\midrule
&\multicolumn{8}{c}{Subjectivity} \\
\midrule
SVM & 85.2 & \textcolor{blue}{\textbf{85.6}} & \textcolor{blue}{85.3} & \textcolor{blue}{85.5} & 87.5 & \textcolor{blue}{\textbf{87.6}} & \textcolor{red}{87.4} & \textcolor{blue}{87.6}\\
CNN & 85.3 & \textcolor{red}{85.2} & \textcolor{blue}{\textbf{86.3}} & \textcolor{blue}{86.0} & 87.9 & \textcolor{red}{87.6} & \textcolor{blue}{\textbf{88.4}} & \textcolor{blue}{88.6}\\
LSTM & 82.9 & \textcolor{red}{82.7} & \textcolor{red}{82.7} & \textcolor{blue}{\textbf{84.1}} & 86.7 & \textcolor{red}{86.3} & \textcolor{red}{85.8} & \textcolor{blue}{\textbf{87.6}}\\
%
% TREC
%
\midrule
&\multicolumn{8}{c}{TREC} \\
\midrule
SVM & 68.5 & \textcolor{red}{\textbf{68.3}} & \textcolor{red}{66.8} & 68.5 & 74.1 & \textcolor{blue}{\textbf{74.7}} & \textcolor{red}{73.2} & \textcolor{blue}{74.3}\\
CNN & 70.9 & \textcolor{red}{70.5} & \textcolor{red}{\textbf{69.0}} & \textcolor{red}{70.0} & 76.1 & \textcolor{blue}{77.7} & \textcolor{blue}{\textbf{77.3}} & \textcolor{blue}{78.0}\\
LSTM & 65.2 & \textcolor{red}{64.5} & \textcolor{red}{63.6} & \textcolor{red}{\textbf{63.8}} & 71.5 & \textcolor{blue}{72.7} & \textcolor{red}{71.0} & \textcolor{blue}{\textbf{73.3}}\\
%
% Customer reviews
%
\midrule
&\multicolumn{8}{c}{Customer reviews} \\
\midrule
SVM & 68.8 & \textcolor{blue}{\textbf{70.5}} & \textcolor{blue}{70.3} & \textcolor{red}{68.5} & 73.6 & \textcolor{blue}{\textbf{74.2}} & \textcolor{red}{72.9} & \textcolor{red}{71.1}\\
CNN & 70.6 & \textcolor{blue}{70.9} & \textcolor{blue}{\textbf{71.7}} & \textcolor{red}{68.2}  & 74.1 & \textcolor{blue}{74.5} & \textcolor{blue}{\textbf{74.8}} & \textcolor{red}{71.5}\\
LSTM & 66.1 & \textcolor{blue}{67.2} & \textcolor{red}{65.1} & \textcolor{red}{\textbf{65.9}} & 68.0 & \textcolor{red}{66.6} & \textcolor{red}{66.5} & \textcolor{red}{\textbf{66.3}}\\
\bottomrule
\end{tabular}
\caption{Text classification accuracy, 
evaluated for each combination 
of acquisition and successor models using uncertainty sampling.
Accuracies are reported for training sets composed of 10\% and 20\% of the document pool. 
Colors indicate performance relative to i.i.d. baselines: 
Blue indicates that a model fared better, 
red that it performed worse,
and black that it performed the same.}
\label{tab:classification}
\end{table*}

\begin{table*}[ht!]
\centering
\vspace{7px}
\begin{tabular}{ l c c c c c c} 
 \multicolumn{7}{c}{\textbf{Named Entity Recognition}} \\
 \toprule
 &\multicolumn{6}{c}{Acquisition Model}\\
 & \multicolumn{3}{c}{10\% of pool} & \multicolumn{3}{c}{20\% of pool} \\
Successor & i.i.d. & CRF & BiLSTM-CNN & i.i.d. & CRF & BiLSTM-CNN\\
%
% CoNLL
%
\midrule
&\multicolumn{6}{c}{CoNLL} \\
\midrule
CRF & 69.2 & \textcolor{blue}{\textbf{70.5}} & \textcolor{blue}{70.2} & 73.6 & \textcolor{blue}{\textbf{74.4}} & \textcolor{blue}{74.0}\\
BiLSTM-CNN & 87.4 & 87.4 & \textcolor{blue}{\textbf{87.8}} & 89.1 & \textcolor{blue}{89.6} & \textcolor{blue}{\textbf{89.6}}\\
%
% OntoNotes
%
\midrule
&\multicolumn{6}{c}{OntoNotes} \\
\midrule
CRF& 73.8 & \textcolor{blue}{\textbf{75.5}} & \textcolor{blue}{75.4} & 77.6 & \textcolor{blue}{\textbf{79.1}} & \textcolor{blue}{78.7}\\
BiLSTM-CNN & 82.6 & \textcolor{blue}{83.1} & \textcolor{blue}{\textbf{83.1}} & 84.6 & \textcolor{blue}{85.2} & \textcolor{blue}{\textbf{84.9}} \\
 \bottomrule
\end{tabular}
\caption{F1 measurements for the NER task, with training sets comprising 10\% and 20\% of the training pool.}
\label{tab:ner}
\end{table*}

%% file: sections/tasks.tex
We now briefly describe the models, datasets, acquisition functions, 
and implementation details for the experiments we conduct with active learners 
for text classification (\ref{sec:text}) and NER (\ref{sec:ner}).

\subsection{Text Classification}
\label{sec:text}

\paragraph{Models}
We consider three standard models for text classification: 
Support Vector Machines (SVMs),
Convolutional Neural Networks (CNNs) \citep{kim2014convolutional,zhang2015sensitivity},
and Bidirectional Long Short-Term Memory (BiLSTM) networks \citep{hochreiter1997long}.
For SVM, we represent texts via sparse,
TF-IDF bag-of-words (BoW) vectors. 
For neural models (CNN and BiLSTM), 
we represent each document as a sequence of word embeddings,
stacked into an $l \times d$ matrix 
where $l$ is the length of the sentence and $d$ 
is the dimensionality of the word embeddings.
We initialize all word embeddings 
with pre-trained GloVe vectors \citep{pennington2014glove}. 

% bcw -- "uniformly at random" -- i presume there is a range [-1, 1] or [0, 1] here???
We initialize vector representations for all words for which we do not have pre-trained embeddings uniformly at random.
For the CNN, we impose a maximum sentence length of $120$ words, 
truncating sentences exceeding this length and padding shorter sentences. 
We used filter sizes of $3$, $4$, and $5$, with $128$ filters per size. 
For BiLSTMs, we selected the maximum sentence length 
such that $90\%$ of sentences in $\mathcal{D}^t$ would be of equal or lesser length.\footnote{Passing longer sentences to the BiLSTM degraded performance 
in preliminary experiments.}
% bcw -- I *swear* I just saw a paper that came out talking about how LSTMs forget negation after about 40 tokens, but cannot for the life of me find it, oh well. that might be a nice explanation...  
We trained all neural models using the Adam optimizer \citep{kingma2014adam}, 
with a learning rate of $0.001$, $\beta_1=0.9$, $\beta_1=0.999$, and $\epsilon = 10^{-8}$.
%Embedding is performed using TF-IDF weighted bag-of-words vectors in LR %and SVM models 
%and pre-trained word2vec vectors  in CNN and LSTM models.

% \begin{itemize}
% \item document logistic regression
% \item document convolutional neural network (cite Kim)
% \end{itemize}

\paragraph{Datasets}
%Enumerate datasets, sources, and any special handling (e.g. under-sampling to even class sizes).

We perform text classification experiments using four benchmark datasets. 
We reserve $20\%$ of each dataset (sampled at i.i.d. random) 
as test data, and use the remaining $80\%$ as the pool of unlabeled data $\mathcal{U}$. 
We sample $2.5\%$ of the remaining documents randomly 
from $\mathcal{U}$ for each $\mathcal{D}_{w}$. 
All models receive the same $\mathcal{D}_{w}$ for any given experiment.

\begin{table*}[ht!]
\centering
\begin{tabular}{l c c c} 
 Dataset & \# Classes & \# Documents & Examples per Class\\ 
 \toprule
 Movie Reviews & 2 & 10662 & 5331, 5331 \\ 
 Subjectivity & 2 & 10000 & 5000, 5000 \\
 TREC & 6 & 5952 & 1300, 916, 95, 1288, 1344, 1009 \\
 Customer Reviews & 2 & 3775 & 1368, 2407 \\ 
\bottomrule
\end{tabular}
\caption{Text classification dataset statistics.}
\label{table:datasets}
\end{table*}

\begin{itemize}[leftmargin=*]
\item \textbf{Movie Reviews}: This corpus consists of sentences drawn from movie reviews. 
The task is to classify sentences as expressing positive or negative sentiment \citep{Pang+Lee:05a}.

\item \textbf{Subjectivity}: 
This dataset consists of statements labeled as either objective or subjective \citep{Pang+Lee:04a}.

\item \textbf{TREC}: 
This task entails categorizing questions into $1$ of $6$ categories based on the subject of the question (e.g., questions about people, locations, and so on) \citep{li2002learning}. 
The TREC dataset defines standard train/test splits, but we generate our own for consistency in train/validation/test proportions across corpora.
%because the default splits
%do not follow the train/validation/test proportions that we desire \citep{li2002learning}.

% bcw -- before we wrote "review sentences" -- are these really sentences, or ar they full reviews?
\item \textbf{Customer Reviews}: 
This dataset is composed of product reviews. The task is to categorize them as positive or negative \citep{hu2004mining}.

\end{itemize}

\subsection{Named Entity Recognition}
\label{sec:ner}

\paragraph{Models}
We consider transfer between two NER models: Conditional Random Fields (CRF) \citep{lafferty2001conditional} and Bidirectional LSTM-CNNs (BiLSTM-CNNs) \citep{chiu2015named}. 

% bcw -- am assuming we tune the word vectors and just use GloVe for initialization. Also, a bit strange - don't we say we use Word2Vec above (for classification)? why the difference?
% bcw -- small thing but a peeve of mine: suggest avoiding "employ" unless you really have a reason; otherwise just say "use"
% bcw -- "To train the CRF, " not really -- these are features, and so independent of training as such
%To train the CRF, 
For the CRF model we use a set of features including word-level and character-based embeddings,
word suffix, capitalization, digit contents, and part-of-speech tags. 
The BiLSTM-CNN model\footnote{Implementation of BiLSTM-CNN is based on \url{https://github.com/asiddhant/Active-NLP}.} 
initializes word vectors to pre-trained GloVe vector embeddings \citep{pennington2014glove}. 
We learn all word and character level features from scratch, initializing with random embeddings. 

\paragraph{Datasets}
%Enumerate datasets, sources, and any special handling.
We perform NER experiments on the CoNLL-2003 and OntoNotes-5.0 English datasets. 
We used the standard test sets for both corpora, 
but merged training and validation sets to form $\mathcal{U}$. 
We initialize each $\mathcal{D}_{w}$ to $2.5\%$ of $\mathcal{U}$.

\begin{itemize}[leftmargin=*]
\item \textbf{CoNLL-2003}: Sentences from Reuters news with words tagged as person, location, organization, 
or miscellaneous entities using an IOB scheme \citep{tjongkimsang2003conll}. 
The corpus contains 301,418 words.
\item \textbf{OntoNotes-5.0}: 
A corpus of sentences drawn from a variety of sources 
including newswire, broadcast news, broadcast conversation, and web data. 
Words are categorized using eighteen entity categories annotated using the IOB scheme \citep{weischedel2013ontonotes}. 
The corpus contains 2,053,446 words.
\end{itemize}

\subsection{Acquisition Functions}
\label{sec:acquisition}
We evaluate these models using three common active learning acquisition functions: classical uncertainty sampling, query by committee (QBC), and Bayesian active learning by disagreement (BALD).

\paragraph{Uncertainty Sampling}

For text classification we use the entropy variant of uncertainty sampling, 
which is perhaps the most widely used AL heuristic \citep{settles.tr09}.
Documents are selected for annotation according to the function 
$$ \underset{\mathbf{x} \in \mathcal{U}}{\operatorname{argmax}}  -\sum_j P(y_j|\mathbf{x})\log P(y_j|\mathbf{x}),$$ 

where $\mathbf{x}$ are instances in the pool $\mathcal{U}$, $j$ indexes potential labels of these 
(we have elided the instance index here) 
and $P(y_j|\mathbf{x})$ is the predicted probability 
that $\mathbf{x}$ belongs to class $y_j$ 
(this estimate is implicitly conditioned 
on a model that can provide such estimates).
For SVM, the equivalent form of this is to choose 
documents closest to the decision boundary.

For the NER task we use maximized normalized log-probability (MNLP) \citep{shen2018deep} 
as our AL heuristic, which adapts the least confidence heuristics 
to sequences by normalizing the log probabilities 
of predicted tag sequence by the sequence length. 
This avoids favoring selecting longer sentences 
(owing to the lower probability of getting the entire tag sequence right).

Documents are sorted in ascending order according to the function 
%Question for Byron and Zack: should we include the explicit definition of our active learning heuristics like this? Is that too much background information? Question for Zack: should the sum of logs be over (y_i|y_1,...y_i-1) rather than (y_i|y_1,...,y_n-1)?
% @ZACK
%%% bcw: I vote yes, we should be explicit, especially because for seq. to you other Q, no -- because it's a bi-directional model, so the conditioning is on both the preceding and following labels. However, y_i should itself be excluded from the r.h.s. (obviously), i think we should make this explicit. 
% bcw STILL A TODO
$$\max_{y_1,...,y_n} \frac{1}{n}\sum_{i=1}^n \log P(y_i|y_1,...,y_{n-1}, \mathbf{x})$$
%% @ZACK (or Dave)
% bcw: - what is j? We do not define this anywwhere as far as I can tell
% - why is it n-1 if we're not using 0-indexing? Shouldn't it just be n?
Where the max over $y$ assignments denotes the most likely set of tags for instance $\mathbf{x}$ and $n$ is the sequence length.
Because explicitly calculating the most likely tag sequence is computationally expensive, we follow \cite{shen2018deep} in using a greedy decoding (i.e., beam search with width $1$) to determine the model's prediction.

\paragraph{Query by Committee}

For our QBC experiments, we use the bagging variant of QBC \citep{mamitsuka1998query}, 
in which a committee of $n$ models is assembled 
by sampling with replacement $n$ sets of $m$ documents 
from the training data ($\mathcal{D}^t$ at each $t$). 
Each model is then trained using a distinct resulting set, 
and the pool documents that maximize their disagreement are selected.
We use $10$ as our committee size, and set $m$ as equal to the number of documents in $\mathcal{D}^t$.

For the text classification task, we compute disagreement using Kullback-Leibler divergence \citep{mccallumzy1998employing}, selecting documents for annotation
according to the function 
$$ \underset{\mathbf{x} \in \mathcal{U}}{\operatorname{argmax}}  \frac{1}{C}\sum_{c=1}^C\sum_j P_c(y_j|\mathbf{x}) \log \frac{P_c(y_j|\mathbf{x})}{P_C(y_j|\mathbf{x})}$$

where $\mathbf{x}$ are instances in the pool $\mathcal{U}$, $j$ indexes potential labels of these instances, and $C$ is the committee size. $P_c(y_j|\mathbf{x})$ is the probability 
that $\mathbf{x}$ belongs to class $y_j$ as predicted by committee member $c$. $P_C(y_j|\mathbf{x})$ represents the consensus probability that $\mathbf{x}$ belongs to class $y_j$, $\frac{1}{C}\sum_{c=1}^{C}P_c(y_j|\mathbf{x})$.

For NER, we compute disagreement using the average per word vote-entropy \citep{dagan1995committee}, selecting sequences for annotation which maximize the function
$$ -\frac{1}{n}\sum_{i=1}^n\sum_m \frac{V(y_i, m)}{C} \log \frac{V(y_i, m)}{C}$$

where $n$ is the sequence length, $C$ is the committee size, and $V(y_i, m)$ is the number of committee members who assign tag $m$ to word $i$ in their most likely tag sequence.
We do not apply the QBC acquisition function to the OntoNotes dataset, 
as training the committee for this larger dataset becomes impractical.

\paragraph{Bayesian AL by Disagreement}

We use the Monte Carlo variant of BALD, 
which exploits an interpretation of dropout regularization as a Bayesian approximation to a Gaussian process \citep{DBLP:journals/corr/GalIG17,siddhant2018deep}. 
This technique entails applying dropout at test time, and then 
estimating uncertainty as the disagreement between outputs realized via multiple passes through the model.
We use the acquisition function proposed in \cite{siddhant2018deep},
which selects for annotation those instances that maximize the number of passes through the model that disagree with the most popular choice:

$$ \underset{\mathbf{x} \in \mathcal{U}}{\operatorname{argmax}}  (1 - \frac{\operatorname{count}(\operatorname{mode}(y_\mathbf{x}^1,...,y_\mathbf{x}^T))}{T})$$

where $\mathbf{x}$ are instances in the pool $\mathcal{U}$, $y_\mathbf{x}^i$ is the class prediction of the $i$th model pass on instance $x$, and $T$ is the number of passes taken through the model. Any ties are resolved using uncertainty sampling over the mean predicted probabilities of all $T$ passes.

In the NER task, agreement is measured across the entire sequence.
Because this acquisition function relies on dropout, 
we do not consider it for non-neural models (SVM and CRF).

%% file: sections/results.tex
We compare transfer between all possible (acquisition, successor) model pairs for each task.
% DL - sentence fragment
%achieved using that training set to the learning curve achieved using an i.i.d. training set. 
We report the performance of each model 
under all acquisition functions both in tables compiling results (Table \ref{tab:classification} and Table \ref{tab:ner} for classification and NER, respectively) and graphically via learning curves that plot predictive performance as a function of train set size (Figure \ref{fig:curves}). 

We report additional results, including all learning curves (for all model pairs and for all tasks), and tabular results (for all acquisition functions) in the Appendix.
We also provide in the Appendix plots resembling \ref{fig:punchline_delta} for all (model, acquisition function) pairs that report the difference between performance under standard AL (in which acquisition and successor model are the same) and that under commensurate i.i.d. data, which affords further analysis of the gains offered by standard AL.
For text classification tasks, we report accuracies; for NER tasks, we report F1.

% bcw 7/2 -- figure (e) is still off; caption runs on to next line...  also, "data set" -> "dataset"

To compare the learning curves, 
we select incremental points along the $x$-axis 
and report the performance at these points. Specifically, we report results with training sets containing 10\% and 20\% of the training pool.
%This allows direct comparison of results.

%% file: sections/discussion.tex
\begin{table*}
\centering
\vspace{7px}
\begin{tabular}{ l c c c c c c c c} 
 &\multicolumn{8}{c}{Successor}\\
 &\multicolumn{2}{c}{Movie Reiews} & \multicolumn{2}{c}{Subjectivity} & \multicolumn{2}{c}{TREC} & \multicolumn{2}{c}{Customer Reviews} \\
Acquisition Model & CNN & LSTM & CNN & LSTM & CNN & LSTM & CNN & LSTM\\
\toprule
CNN & -- & 0.961 & -- & 0.968 & -- & 0.988 & -- & 0.973 \\
LSTM & 0.989 & -- & 0.996 & -- & 0.992 & -- & 0.980 & -- \\
SVM & 0.991 & 0.961 & 0.997 & 0.970 & 0.990 & 0.987 & 0.991 & 0.974 \\
 \bottomrule
\end{tabular}
\caption{Average Spearman's rank correlation coefficients (over five runs) of cosine distances between test set representations learned with native active learning and distances between those learned with transferred actively acquired datasets, at the end of the AL process. Uncertainty is used as the acquisition function in all cases.}
\label{tab:spearman} %\vspace{-.5em}
\end{table*}

%\vspace{-.5em}
Results in Tables \ref{tab:classification} and \ref{tab:ner} 
demonstrate that standard AL --- 
where the acquisition and successor models 
are one and the same --- performs inconsistently across text classification datasets.
In $75\%$ of all combinations of model, dataset, and training set size, there exists some acquisition function that outperforms i.i.d. data. 
This is consistent with the prior literature indicating the effectiveness of AL.
However, when implementing AL in a real, live setting, a practitioner would choose a \emph{single} acquisition function ahead of time.
To accurately reflect this scenario, we must consider the performance of individual acquisition functions across multiple datasets. 
Results for individual AL strategies are more equivocal.
In our reported classification datapoints, standard AL outperforms i.i.d. sampling in only a slight majority ($60.9\%$) of cases.

AL thus seems to yield modest (though inconsistent) improvements over i.i.d. random sampling, but our results further suggest that this comes at an additional cost: the acquired dataset may not generalize well to new learners.
Specifically, models trained on \emph{foreign} actively acquired datasets tend to underperform those trained on i.i.d. datasets. 
We observe this most clearly in the classification task, where only a handful of (acquisition, successor, acquisition function) combinations lead to performance greater than that achieved using i.i.d. data. 
%We observe performance greater than the i.i.d. baseline in only 37.5\% of the tabulated data points representing dataset transfer (in which acquisition and successor models differ).
Specifically, only 37.5\% of the tabulated data points representing dataset transfer (in which acquisition and successor models differ) outperform the i.i.d. baseline.

Results for NER are more favorable for AL.
For this task we observe consistent improved performance versus the i.i.d. baseline in both standard AL data points and transfer data points.
These results are consistent with previous findings on transferring actively acquired datasets for NER \cite{tomanek2011inspecting}. % @Dave TODO double check, and in general integrate discussion of htis paper somewhere?

In standard AL for text classification, the only (model, acquisition function) pairs that we observe to produce better than i.i.d. results with any regularity are uncertainty with SVM or CNN, and BALD with CNN.
When transferring actively acquired datasets, we do not observe consistently better than i.i.d. results with \emph{any} combination of acquisition model, successor model, and acquisition function. The success of AL appears to depend very much on the dataset. For example, AL methods -- both in the standard and acquisition/successor settings -- perform much more reliably on the Subjectivity dataset than any other. In contrast, AL performs consistently poorly on the TREC dataset.

 Our findings suggest that AL is brittle. During experimentation, we also found that performance often depends on factors that one may think are minor design decisions.
 For example, our setup largely resembles that of \citet{siddhant2018deep}, yet initially we observed large discrepancies in results. 
 Digging into this revealed that much of the difference was due to our use of word2vec \cite{mikolov2013efficient} rather than GloVe \cite{pennington2014glove} for word embedding initializations.
 %An additional difference concerned the decision to devote a portion of the budget to collection of an i.i.d. validation set during AL.
 That small decisions like this can result in relatively pronounced performance differences for AL strategies is disconcerting. 
 %We note that other research has found more consistently positive results using various active learning strategies.
 %Our models closely mirror those used in \citet{siddhant2018deep}, yet we do not produce similarly consistent results.
% In contrast to this work, we do not devote a portion of our acquisition budget to building a validation set, as active learning is likely to be employed under circumstances where the budget is too low for this to be feasible.
 %We further evaluate on two additional datasets.
 %These changes have outsized impact on the final performance of active learning.
% We also observe that changes to word embedding initialization (e.g. using word2vec \citep{mikolov2013efficient} instead of glove) can have surprising downstream impact on active learning performance, sometimes making the difference between results that are better than i.i.d. and results that are not.

%A hypothetical cause for this poor performance when using transferred actively acquired data is that datasets actively acquired using foreign models induce different representation spaces than those attained using natively actively acquired data. 
A key advantage afforded by neural models is representation learning. A natural question here is therefore whether the representations induced by the neural models differs as a function of the acquisition strategy. 
To investigate this, we measure pairwise distances between instances in the learned feature space after training. 
Specifically, for each test instance we calculate its cosine similarity to all other test instances, inducing a ranking. 
We do this in the three different feature spaces learned by the CNN and LSTM models, respectively, after sampling under the three acquisition models.% shown in Table \ref{tab:spearman}. 

We quantify dissimilarities between the rankings induced under different representations via Spearman's rank correlation coefficients. 
We repeat this for all instances in the test set, and average over these coefficients to derive
an overall similarity measure, which may be viewed as quantifying the similarity between learned feature spaces via average pairwise similarities within them.
As reported in Table \ref{tab:spearman}, despite the aforementioned differences in predictive performance, the learned representations seem to be similar. In other words, sampling under foreign acquisition models does not lead to notably different representations.  % bcw -- although this is only in absolute terms and we don't really have a "baseline" to which we can make a relative comparison.

%compare the cosine distances all documents in the test set in given learned representation space.
%We then calculate the average Spearman's rank correlation coefficients of the distances obtained using representations learned with native data and those learned  with transferred data. We report the results for uncertainty sampling in Table \ref{tab:spearman}, which demonstrate that the learned representations, in fact, encode largely the same relationships, indicating that the point of failure is further downstream. Average Spearman's rank correlation coefficients for other acquisition strategies are reported in the appendix.

%% file: sections/conclusions.tex
%\vspace{-.5em}
We extensively evaluated standard AL methods under varying model, domain, and acquisition function combinations for two standard NLP tasks (text classification and sequence tagging). 
We also assessed performance achieved when transferring an actively sampled training dataset from an acquisition model to a distinct successor model. 
Given the longevity and value of training sets and the frequency at which new ML models advance the state-of-the-art, this should be an anticipated scenario: Annotated data often outlives models. 

%We have analyzed these via an extensive empirical study including two standard natural language processing tasks (text classification and sequence tagging), coupled with multiple datasets, acquisition functions, and acquisition, successor model pairs with each. 

Our findings indicate that AL performs unreliably. 
While a specific acquisition function and model applied to a particular task and domain may be quite effective, it is not clear that this can be predicted ahead of time.
Indeed, there is no way to retrospectively determine the relative success of AL without collecting a relatively large quantity of i.i.d. sampled data, and this would undermine the purpose of AL in the first place. 
Further, even if such an i.i.d. sample were taken as a diagnostic tool early in the active learning cycle, relative success early in the AL cycle is not necessarily indicative of relative success later in the cycle, as illustrated by Figure \ref{fig:punchline_delta}.

Problematically, even in successful cases, an actively sampled training set is linked to the model used to acquire it. 
We have found that training successor models with this set will often result in performance worse than that attained using an equivalently sized i.i.d. sample. Results are more favorable to AL for NER, as compared to text classification, which is consistent with prior work \cite{tomanek2011inspecting}.

%The apparent importance of the domain in which AL is applied, the degree to which acquisition function performance varies over domains, and the inconsistent performance of subsequent (distinct) models trained with actively acquired data collectively raise questions about the practicality of employing AL. There is no obvious way to predict whether a given dataset will produce positive results with a given AL selection function prior to the acquisition of that dataset.

%It is difficult to predict beforehand whether any specific application of active learning will provide positive results. The only factor that seems to provide significant predictive power is the machine learning task, and even this is not completely reliable. 
% DL - is this the line we want to end on? We've spent the whole paper pointing out how often it DOESN'T outperform i.i.d.
In short, the relative performance of individual active acquisition functions varies considerably over datasets and domains.
While AL often does yield gains over i.i.d. sampling, 
these tend to be marginal and inconsistent. 
Moreover, this comes at a relatively steep cost: 
The acquired dataset may be disadvantageous for training subsequent models. 
Together these findings raise serious concerns 
regarding the efficacy of active learning in practice.

%% file: sections/appendix_b.tex
Below, we present full results for all our experiments in the form of tabular results and learning curves. Tables \ref{tab:classification_appendix} and \ref{tab:ner_appendix} enumerate performance metrics for all source, successor, acquisition function combinations after acquiring $10\%$ and $20\%$ of the pool. Figure \ref{fig:appendix_pg1} shows the learning curves for all combinations. We report all average Spearman's rank correlation coefficients in Table \ref{tab:spearman_appendix}.\\

\renewcommand*{\thesubfigure}{\arabic{subfigure}}

\begin{table*}
%\footnotesize
% \centering
\begin{minipage}[c]{\textwidth}
\begin{tabular}{l c c c c c c c c c c} 
 \multicolumn{11}{c}{\textbf{Text classification}} \\
 \toprule
 &&\multicolumn{8}{c }{{Acquisition model}} \\
&&& \multicolumn{3}{c}{Uncertainty} & \multicolumn{3}{c}{QBC} & \multicolumn{2}{c}{BALD} \\
%  &\multicolumn{4}{c}{\small Source Model}\\
Successor & pool \% & i.i.d. & SVM & CNN & LSTM & 
 SVM & CNN & LSTM & CNN & LSTM\\

%
% Movie reviews
%
\midrule
&&\multicolumn{9}{c}{Movie reviews} \\
\midrule
\multirow{2}{*}{SVM} & 10 & 65.3 & \textbf{65.3} & \textcolor{blue}{65.8} & \textcolor{blue}{65.7} & \textcolor{red}{\textbf{64.9}} & \textcolor{red}{64.9} & \textcolor{red}{65.1} & \textcolor{red}{64.9} & \textcolor{red}{65.2}\\ & 20 & 68.2 & \textcolor{blue}{\textbf{69.0}} & \textcolor{blue}{69.4} & \textcolor{blue}{68.9} & \textcolor{red}{\textbf{68.1}} & \textcolor{blue}{68.4} & \textcolor{blue}{68.7} & \textcolor{blue}{68.5} & \textcolor{blue}{69.0}\\
\multirow{2}{*}{CNN} & 10 & 65.0 & \textcolor{blue}{65.3} & \textcolor{blue}{\textbf{65.5}} & \textcolor{blue}{65.4} & \textcolor{red}{64.8} & \textcolor{blue}{\textbf{65.1}} & \textcolor{red}{64.7} & \textcolor{blue}{\textbf{65.1}} & \textcolor{red}{64.9}\\ & 20 & 69.4 & \textcolor{red}{69.1} & \textcolor{blue}{\textbf{69.5}} & \textcolor{blue}{69.5} & \textcolor{red}{68.5} & \textcolor{red}{\textbf{69.1}} & \textcolor{red}{69.1} & \textcolor{red}{\textbf{68.3}} & \textcolor{red}{69.1}\\
\multirow{2}{*}{LSTM} & 10 & 63.0 & \textcolor{red}{62.0} & \textcolor{red}{62.5} & \textcolor{blue}{\textbf{63.1}} & \textcolor{red}{61.9} & \textcolor{red}{61.9} & \textcolor{red}{\textbf{62.6}} & \textcolor{red}{61.7} & \textcolor{red}{\textbf{62.2}}\\ & 20 & 67.2 & \textcolor{red}{65.1} & \textcolor{red}{65.8} & \textcolor{red}{\textbf{67.0}} & \textcolor{red}{65.4} & \textcolor{red}{65.7} & \textcolor{red}{\textbf{66.8}} & \textcolor{red}{65.6} & \textcolor{red}{\textbf{67.1}}\\
%
% Subjectivity
%
\midrule
&&\multicolumn{9}{c}{Subjectivity} \\
\midrule
\multirow{2}{*}{SVM} & 10 & 85.2 & \textcolor{blue}{\textbf{85.6}} & \textcolor{blue}{85.3} & \textcolor{blue}{85.5} & \textcolor{blue}{\textbf{85.4}} & \textcolor{red}{85.0} & \textcolor{blue}{85.4} & \textcolor{blue}{85.8} & \textcolor{blue}{85.4}\\ & 20 & 87.5 & \textcolor{blue}{\textbf{87.6}} & \textcolor{red}{87.4} & \textcolor{blue}{87.6} & \textcolor{blue}{\textbf{87.7}} & \textcolor{red}{87.0} & 87.5 & \textcolor{red}{87.0} & \textcolor{blue}{87.6}\\
\multirow{2}{*}{CNN} & 10 & 85.3 & \textcolor{red}{85.2} & \textcolor{blue}{\textbf{86.3}} & \textcolor{blue}{86.0} & 85.3 & \textcolor{blue}{\textbf{86.0}} & \textcolor{blue}{85.7} & \textcolor{blue}{\textbf{86.2}} & \textcolor{blue}{85.7}\\ & 20 & 87.9 & \textcolor{red}{87.6} & \textcolor{blue}{\textbf{88.4}} & \textcolor{blue}{88.6} & \textcolor{blue}{88.4} & \textcolor{blue}{\textbf{88.5}} & \textcolor{blue}{88.6} & \textcolor{blue}{\textbf{88.6}} & \textcolor{blue}{88.3}\\
\multirow{2}{*}{LSTM} & 10 & 82.9 & \textcolor{red}{82.7} & \textcolor{red}{82.7} & \textcolor{blue}{\textbf{84.1}} & \textcolor{blue}{83.3} & \textcolor{blue}{83.7} & \textcolor{blue}{\textbf{84.8}} & \textcolor{blue}{83.1} & \textcolor{blue}{\textbf{84.2}}\\ & 20 & 86.7 & \textcolor{red}{86.3} & \textcolor{red}{85.8} & \textcolor{blue}{\textbf{87.6}} & \textcolor{blue}{86.9} & \textcolor{blue}{87.0} & \textcolor{blue}{\textbf{87.7}} & \textcolor{red}{84.7} & \textcolor{blue}{\textbf{87.0}}\\
%
% TREC
%
\midrule
&&\multicolumn{9}{c}{TREC} \\
\midrule
\multirow{2}{*}{SVM} & 10 & 68.5 & \textcolor{red}{\textbf{68.3}} & \textcolor{red}{66.8} & 68.5 & \textcolor{red}{\textbf{68.1}} & \textcolor{red}{63.1} & \textcolor{red}{64.9} & \textcolor{red}{68.2} & \textcolor{red}{68.3}\\ & 20 & 74.1 & \textcolor{blue}{\textbf{74.7}} & \textcolor{red}{73.2} & \textcolor{blue}{74.3} & \textcolor{red}{\textbf{73.7}} & \textcolor{red}{71.6} & \textcolor{red}{71.2} & 74.1 & 74.1\\
\multirow{2}{*}{CNN} & 10 & 70.9 & \textcolor{red}{70.5} & \textcolor{red}{\textbf{69.0}} & \textcolor{red}{70.0} & \textcolor{red}{67.4} & \textcolor{red}{\textbf{62.8}} & \textcolor{red}{69.5} & \textcolor{blue}{\textbf{71.0}} & \textcolor{red}{70.5}\\ & 20 & 76.1 & \textcolor{blue}{77.7} & \textcolor{blue}{\textbf{77.3}} & \textcolor{blue}{78.0} & \textcolor{blue}{76.5} & \textcolor{red}{\textbf{73.7}} & \textcolor{blue}{76.3} & \textcolor{blue}{\textbf{79.8}} & \textcolor{blue}{77.7}\\
\multirow{2}{*}{LSTM} & 10 & 65.2 & \textcolor{red}{64.5} & \textcolor{red}{63.6} & \textcolor{red}{\textbf{63.8}} & \textcolor{red}{61.7} & \textcolor{red}{60.1} & \textcolor{red}{\textbf{64.6}} & \textcolor{red}{64.1} & \textcolor{red}{\textbf{64.5}}\\ & 20 & 71.5 & \textcolor{blue}{72.7} & \textcolor{red}{71.0} & \textcolor{blue}{\textbf{73.3}} & \textcolor{red}{71.4} & \textcolor{red}{69.9} & \textcolor{blue}{\textbf{71.8}} & \textcolor{blue}{72.9} & \textcolor{blue}{\textbf{72.6}}\\
%
% Customer reviews
%
\midrule
&&\multicolumn{9}{c}{Customer reviews} \\
\midrule
\multirow{2}{*}{SVM} & 10 & 68.8 & \textcolor{blue}{\textbf{70.5}} & \textcolor{blue}{70.3} & \textcolor{red}{68.5} & \textcolor{blue}{\textbf{70.5}} & \textcolor{blue}{69.5} & \textcolor{red}{64.6} & \textcolor{blue}{70.0} & \textcolor{blue}{69.2}\\ & 20 & 73.6 & \textcolor{blue}{\textbf{74.2}} & \textcolor{red}{72.9} & \textcolor{red}{71.1} & \textcolor{blue}{\textbf{73.8}} & \textcolor{red}{72.6} & \textcolor{red}{65.7} & \textcolor{red}{73.5} & \textcolor{red}{71.7}\\
\multirow{2}{*}{CNN} & 10 & 70.6 & \textcolor{blue}{70.9} & \textcolor{blue}{\textbf{71.7}} & \textcolor{red}{68.2} & \textcolor{blue}{71.5} & \textcolor{blue}{\textbf{71.4}} & \textcolor{red}{63.8} & \textcolor{blue}{\textbf{72.2}} & \textcolor{red}{68.4}\\ & 20 & 74.1 & \textcolor{blue}{74.5} & \textcolor{blue}{\textbf{74.8}} & \textcolor{red}{71.5} & \textcolor{blue}{74.9} & \textcolor{blue}{\textbf{74.9}} & \textcolor{red}{65.2} & \textcolor{blue}{\textbf{75.3}} & \textcolor{red}{71.3}\\
\multirow{2}{*}{LSTM} & 10 & 66.1 & \textcolor{blue}{67.2} & \textcolor{red}{65.1} & \textcolor{red}{\textbf{65.9}} & \textcolor{red}{65.0} & \textcolor{red}{64.8} & \textcolor{red}{\textbf{64.0}} & \textcolor{red}{65.2} & \textcolor{red}{\textbf{65.4}}\\ & 20 & 68.0 & \textcolor{red}{66.6} & \textcolor{red}{66.5} & \textcolor{red}{\textbf{66.3}} & \textcolor{red}{66.3} & \textcolor{red}{66.4} & \textcolor{red}{\textbf{65.4}} & \textcolor{blue}{68.3} & \textbf{68.0}\\
\bottomrule
\end{tabular}
\caption{Text classification accuracy, 
evaluated for each combination 
of acquisition and successor models using uncertainty sampling, QBC, and BALD.
Accuracies are reported for training sets composed of 10\% and 20\% of the document pool. 
Colors indicate performance relative to i.i.d. baselines: 
Blue implies that a model fared better, 
red that it performed worse,
and black that it performed the same.}
\label{tab:classification_appendix}
\end{minipage}
\end{table*}

\begin{table*}
\centering
\vspace{7px}
\begin{tabular}{ l c c c c c c c} 
 \multicolumn{8}{c}{\textbf{Named Entity Recognition}} \\
 \toprule
 &&\multicolumn{6}{c}{Acquisition Model}\\
 &&& \multicolumn{2}{c}{Uncertainty} & BALD & \multicolumn{2}{c}{QBC} \\
Successor & pool \% & i.i.d. & CRF & BiLSTM-CNN &  BiLSTM-CNN & CRF & BiLSTM-CNN\\
%
% CoNLL
%
\midrule
&&\multicolumn{6}{c}{CoNLL} \\
\midrule
\multirow{2}{*}{CRF} & 10 & 69.2 & \textcolor{blue}{\textbf{70.5}} & \textcolor{blue}{70.2} & \textcolor{blue}{70.3} & \textcolor{blue}{\textbf{70.3}} & \textcolor{blue}{70.0}\\ & 20 & 73.6 & \textcolor{blue}{\textbf{74.4}} & \textcolor{blue}{74.0} & \textcolor{blue}{74.1} & \textcolor{blue}{\textbf{74.5}} & \textcolor{blue}{74.1}\\
\multirow{2}{*}{BiLSTM-CNN} & 10 & 87.4 & 87.4 & \textcolor{blue}{\textbf{87.8}} & \textcolor{blue}{\textbf{88.0}} & \textcolor{blue}{87.5} & \textcolor{blue}{\textbf{87.7}}\\ & 20 & 89.1 & \textcolor{blue}{89.6} & \textcolor{blue}{\textbf{89.6}} & \textcolor{blue}{\textbf{89.8}} & \textcolor{blue}{89.2} & \textcolor{blue}{\textbf{89.5}}\\
 \midrule
\end{tabular}

\begin{tabular}{ l c c c c c} 
 &&\multicolumn{4}{c}{Acquisition Model}\\
 &&& \multicolumn{2}{c}{Uncertainty} & BALD \\
Successor & pool \% & i.i.d. & CRF & BiLSTM-CNN &  BiLSTM-CNN\\
%
% OntoNotes
%
\midrule
&&\multicolumn{4}{c}{OntoNotes} \\
\midrule
\multirow{2}{*}{CRF} & 10 & 73.8 & \textcolor{blue}{\textbf{75.5}} & \textcolor{blue}{75.4} & \textcolor{blue}{75.3} \\ & 20 & 77.6 & \textcolor{blue}{\textbf{79.1}} & \textcolor{blue}{78.7} & \textcolor{blue}{78.7} \\
\multirow{2}{*}{BiLSTM-CNN} & 10 & 82.6 & \textcolor{blue}{83.1} & \textcolor{blue}{\textbf{83.1}} & \textcolor{blue}{\textbf{83.2}} \\ & 20 & 84.6 & \textcolor{blue}{85.2} & \textcolor{blue}{\textbf{84.9}} & \textcolor{blue}{\textbf{85.1}} \\
 \bottomrule
\end{tabular}
\caption{F1 measurements for the NER task, with training sets comprising 10\% and 20\% of the training pool.}
\label{tab:ner_appendix}
\end{table*}

\begin{figure*}[ht]
  \centering
  \caption{This appendix contains the full set of collected learning curves for the text classification and NER. Error bars represent one standard deviation.
  }
	\begin{subfigure}{0.44\linewidth} % width of left subfigure
		\includegraphics[width=\linewidth]{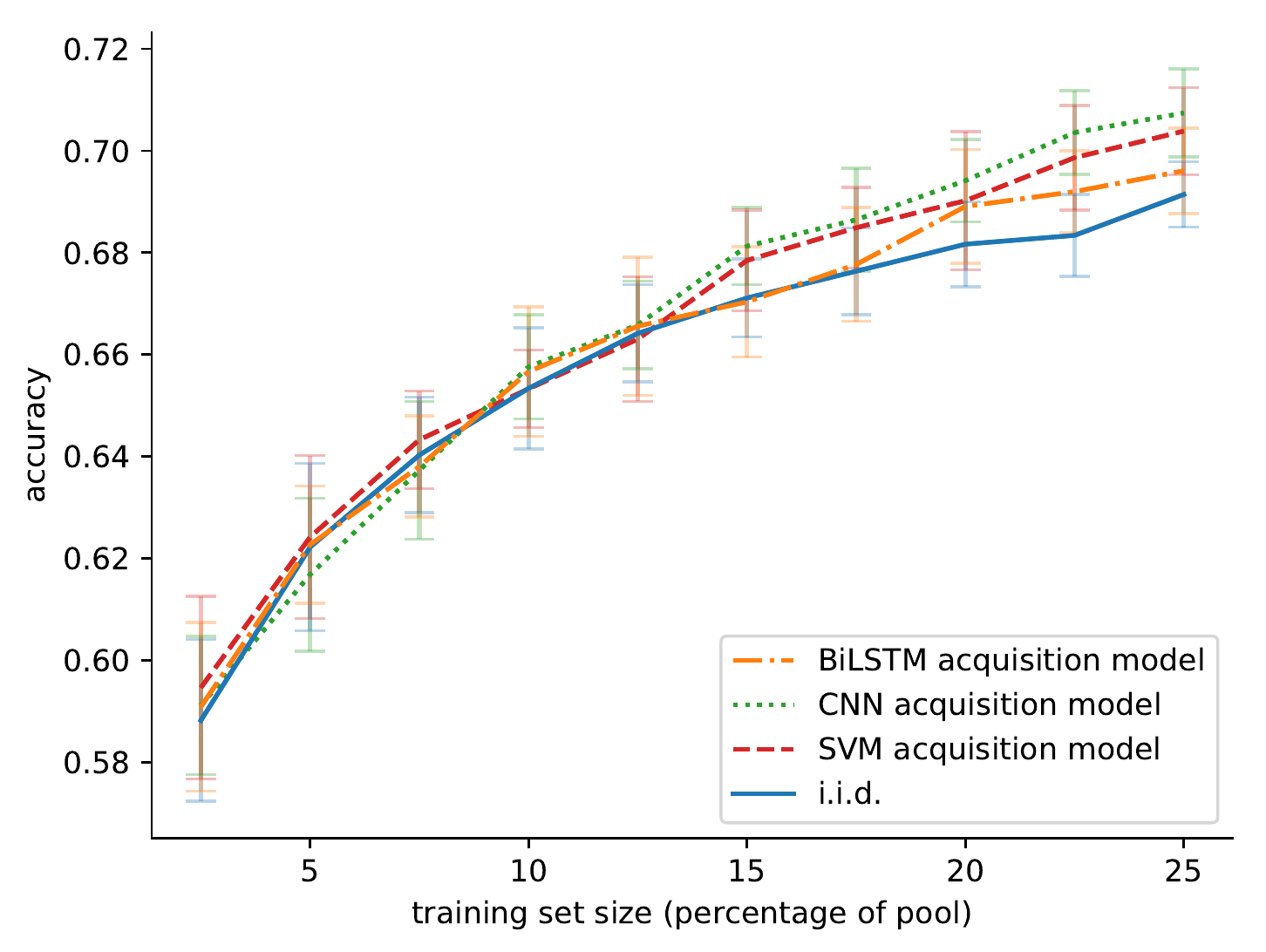}
		\caption{SVM on Movie Reviews dataset using max entropy } % subcaption
  \end{subfigure}
  \begin{subfigure}{0.44\linewidth} % width of left subfigure
		\includegraphics[width=\linewidth]{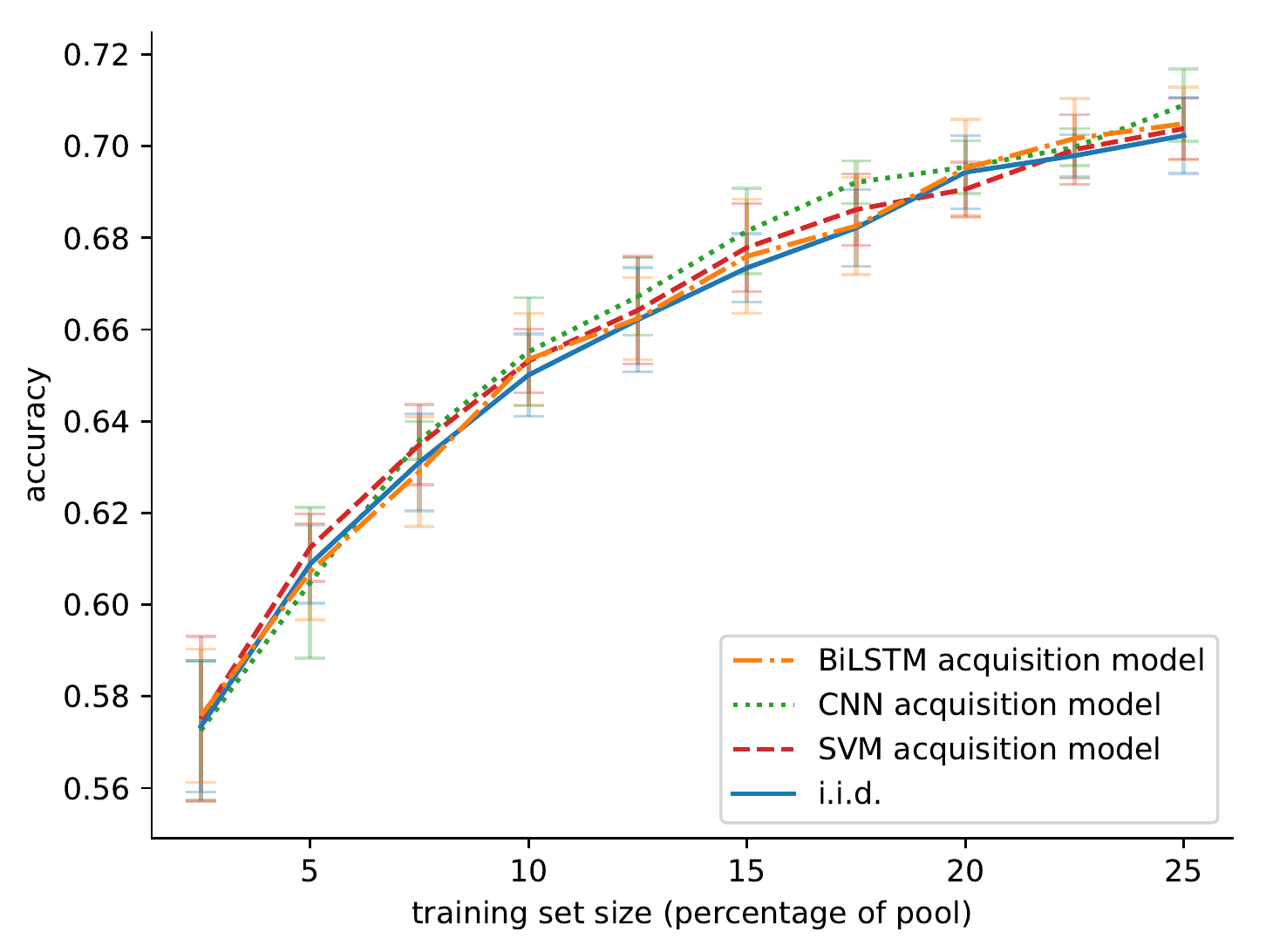}
		\caption{CNN on Movie Reviews dataset using max entropy } % subcaption
  \end{subfigure}\\
 
  \begin{subfigure}{0.44\linewidth} % width of left subfigure
		\includegraphics[width=\linewidth]{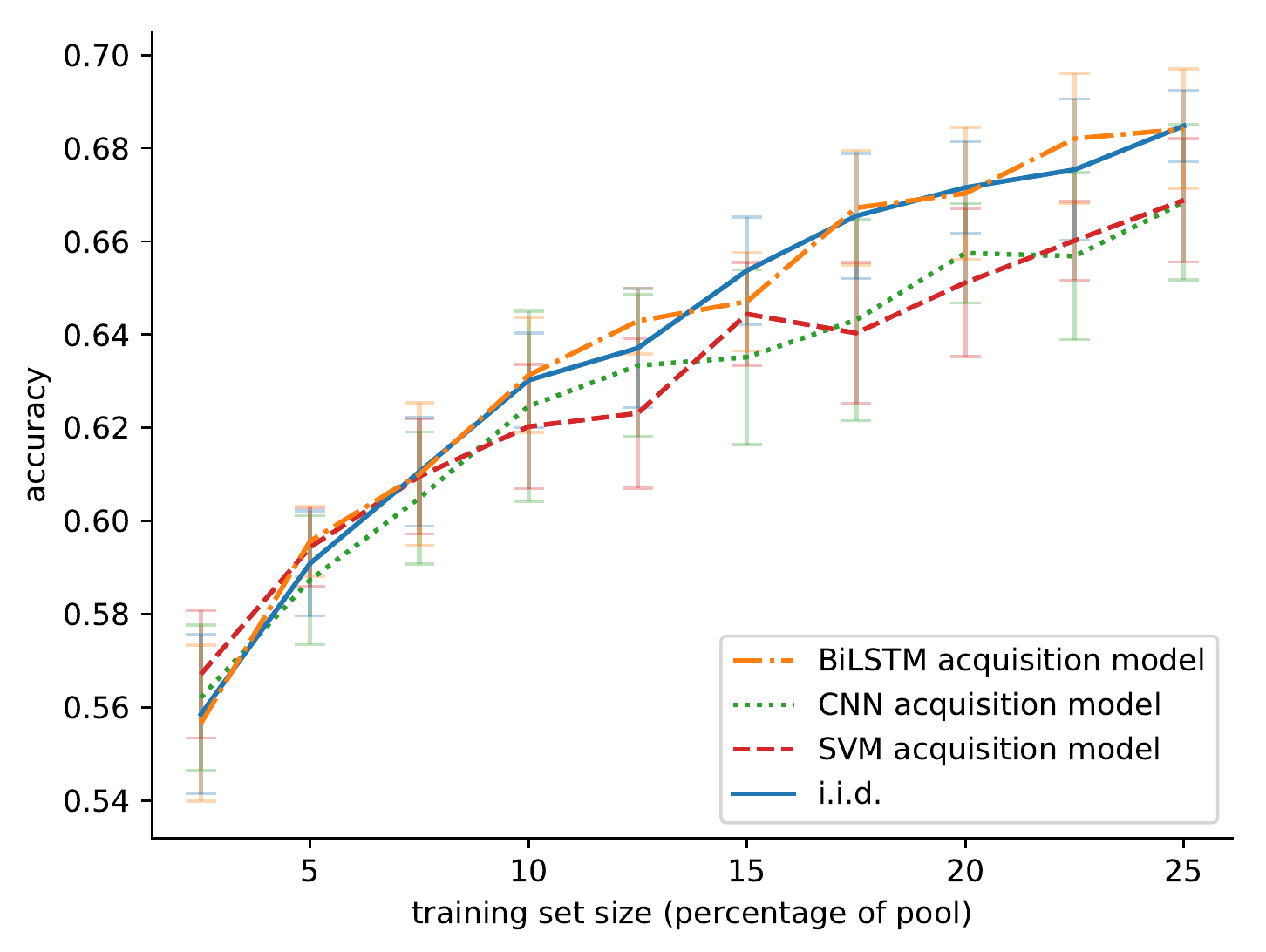}
		\caption{BiLSTM on Movie Reviews dataset using max entropy } % subcaption
  \end{subfigure}
  \begin{subfigure}{0.44\linewidth} % width of left subfigure
		\includegraphics[width=\linewidth]{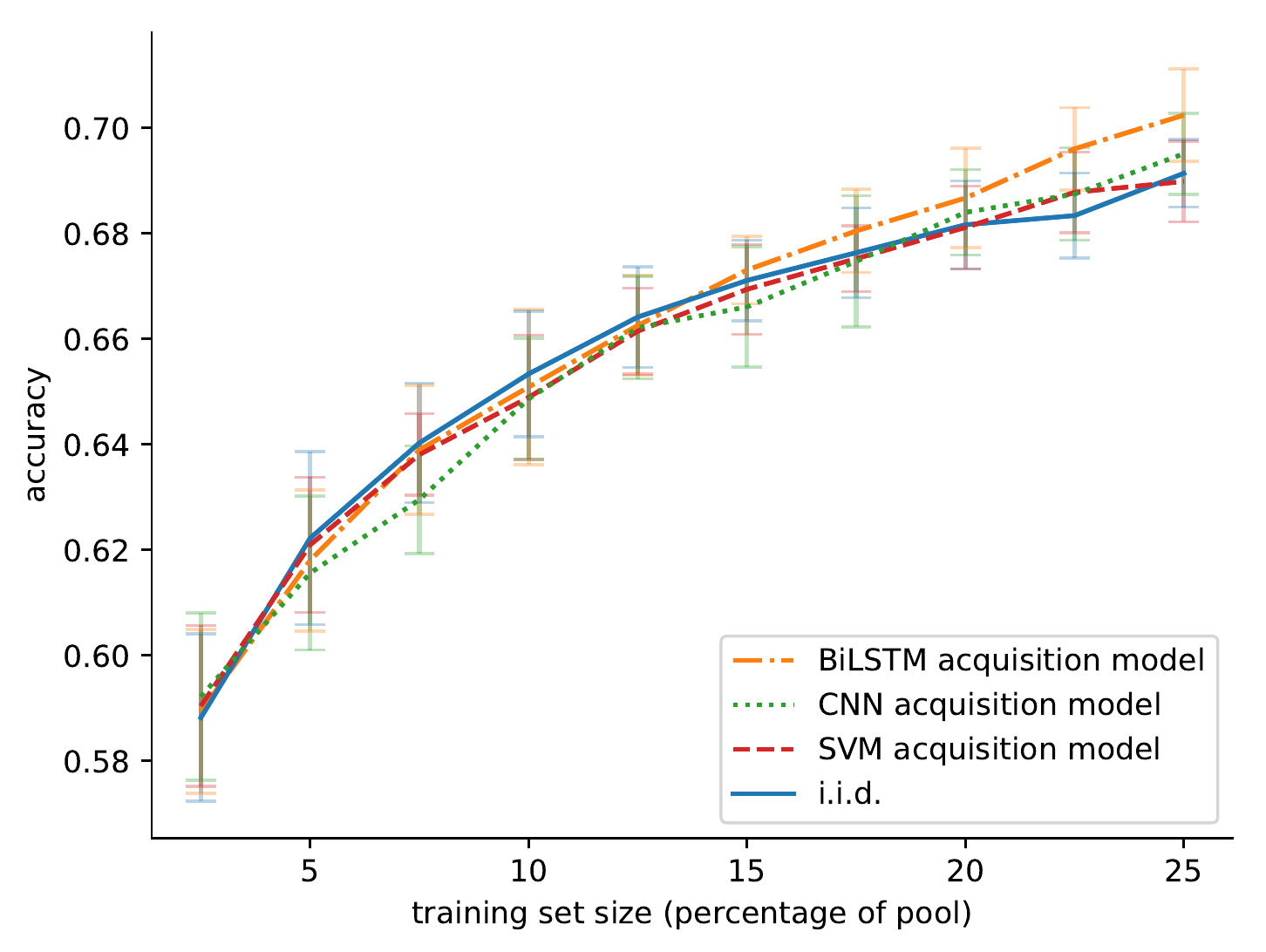}
		\caption{SVM on Movie Reviews dataset using QBC } % subcaption
  \end{subfigure}\\
  
  \begin{subfigure}{0.44\linewidth} % width of left subfigure
		\includegraphics[width=\linewidth]{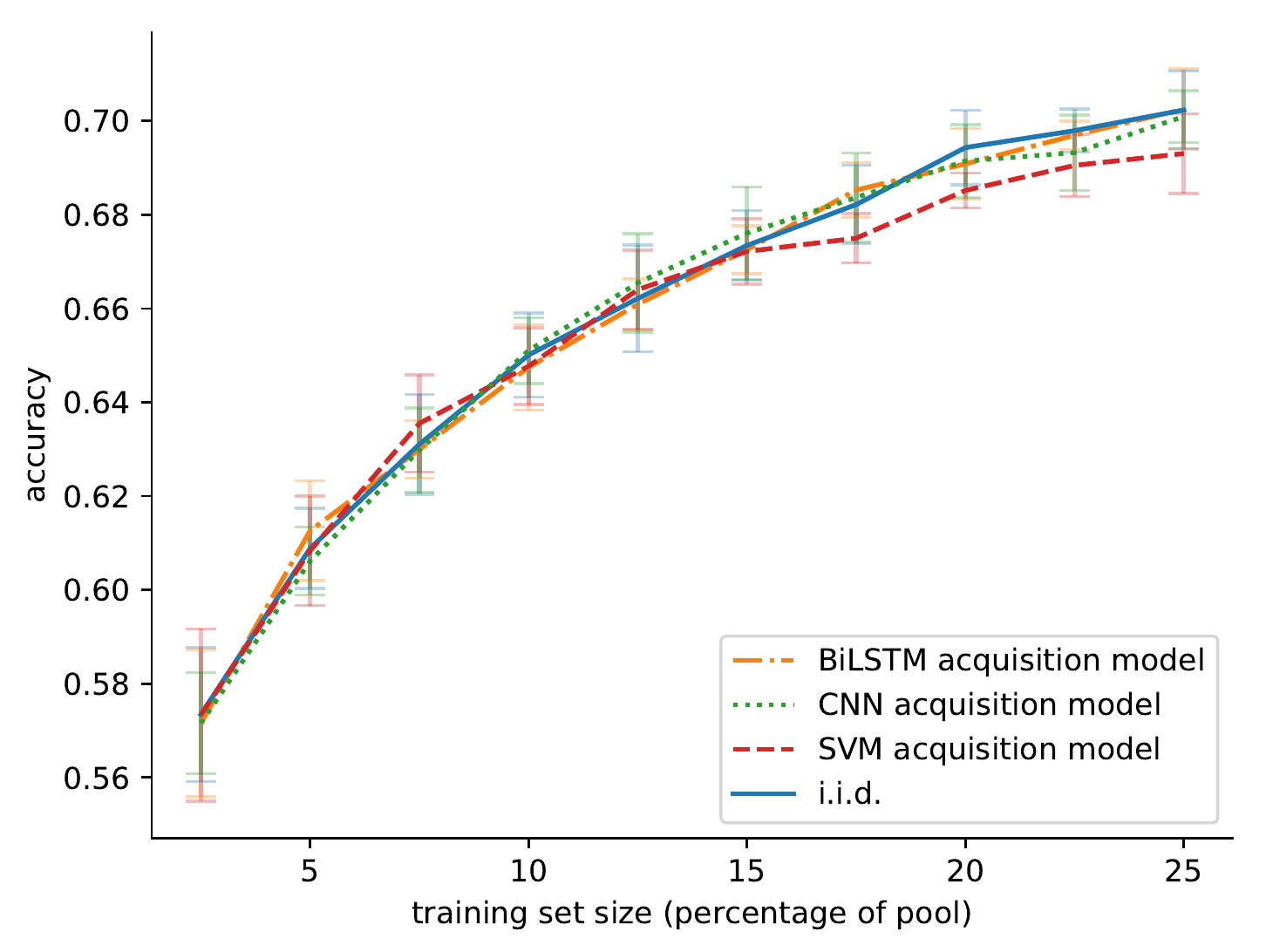}
		\caption{CNN on Movie Reviews dataset using QBC } % subcaption
  \end{subfigure}
    \begin{subfigure}{0.44\linewidth} % width of left subfigure
		\includegraphics[width=\linewidth]{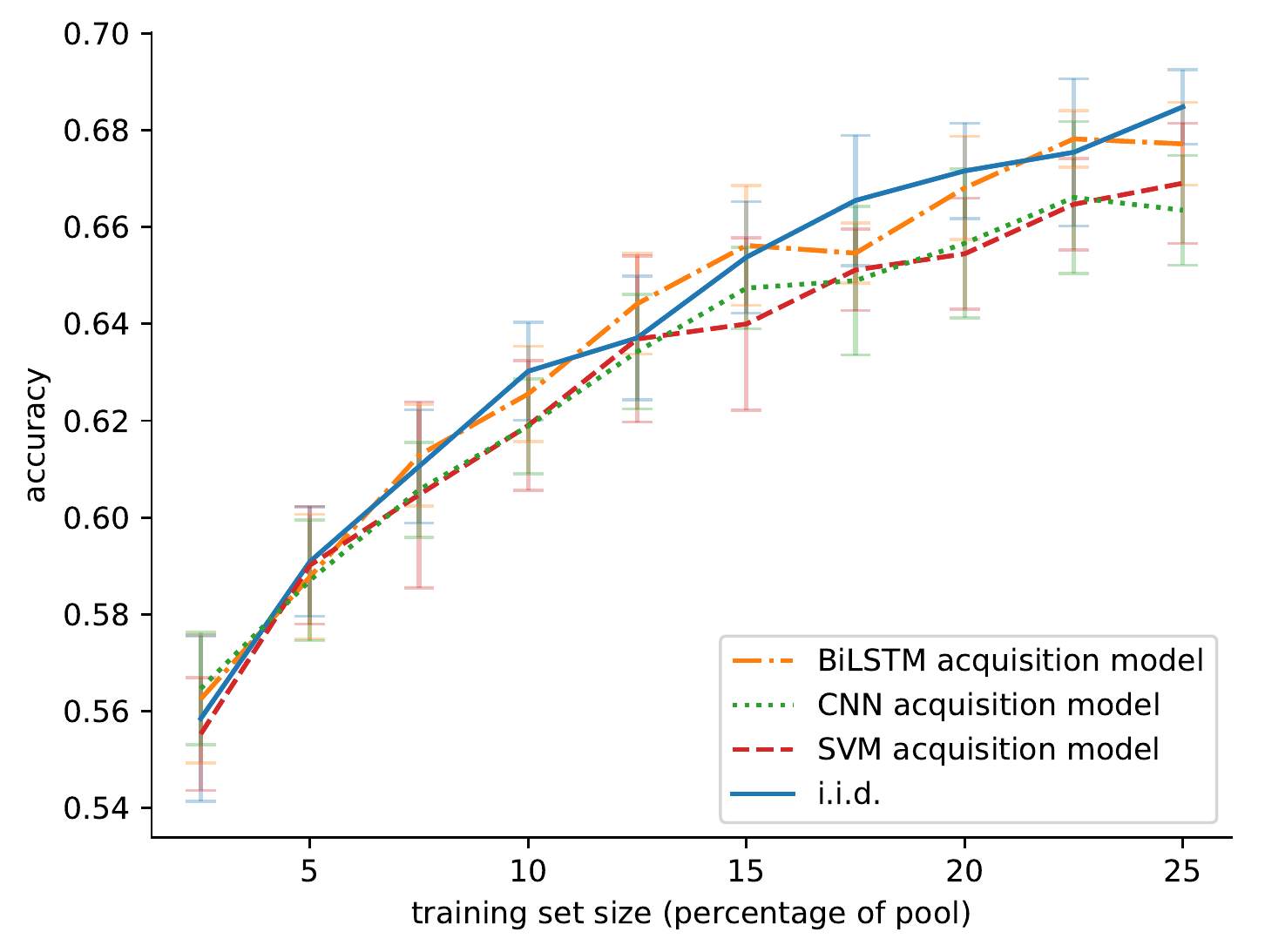}
		\caption{BiLSTM on Movie Reviews dataset using QBC } % subcaption
  \end{subfigure}\\
  
  \label{fig:appendix_pg1}
\end{figure*}

\begin{figure*}\ContinuedFloat
  \centering
  \begin{subfigure}[t]{0.44\linewidth} % width of left subfigure
		\includegraphics[width=\linewidth]{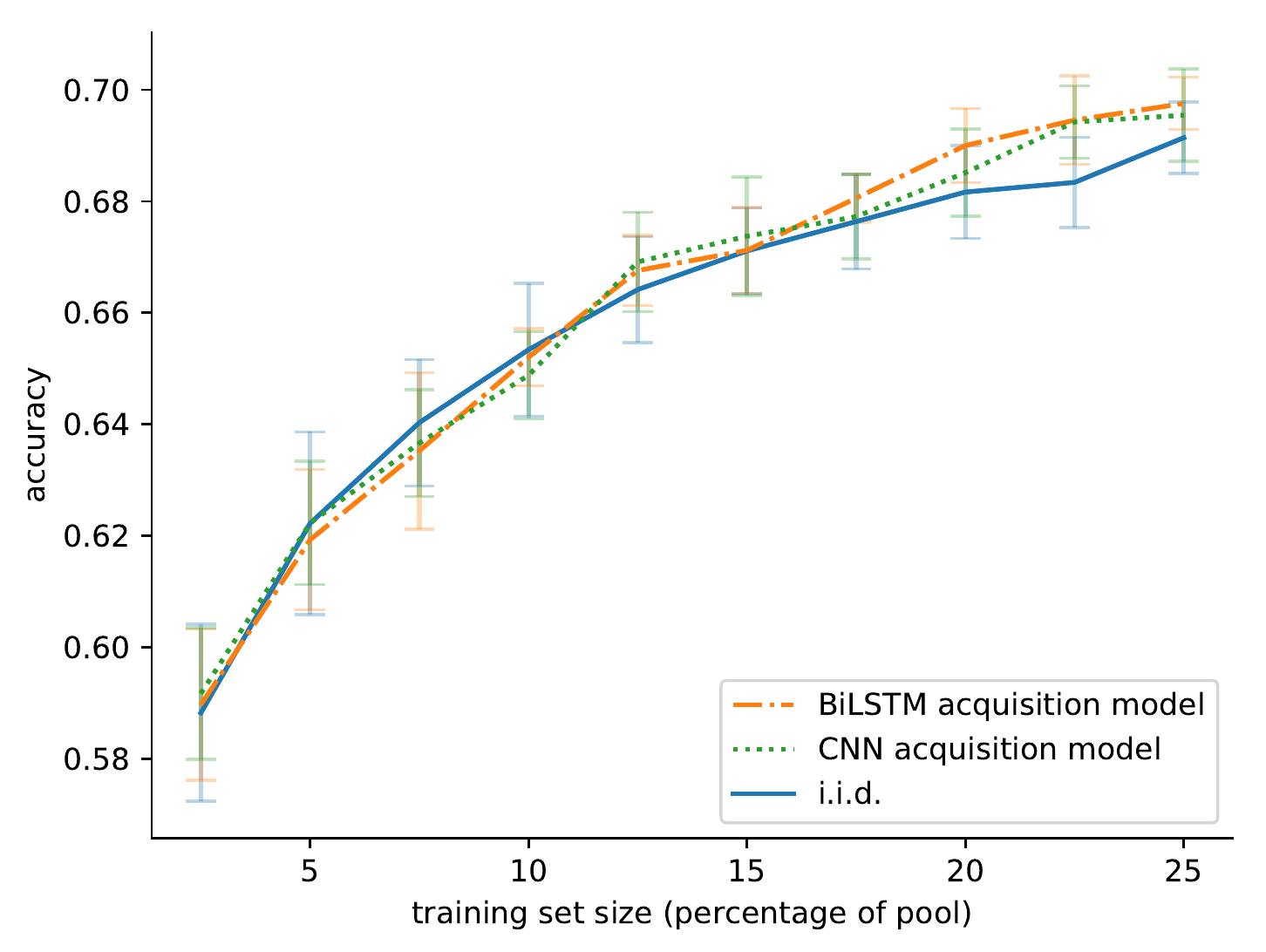}
		\caption{SVM on Movie Reviews dataset using BALD } % subcaption
  \end{subfigure}
  \begin{subfigure}[t]{0.44\linewidth} % width of left subfigure
		\includegraphics[width=\linewidth]{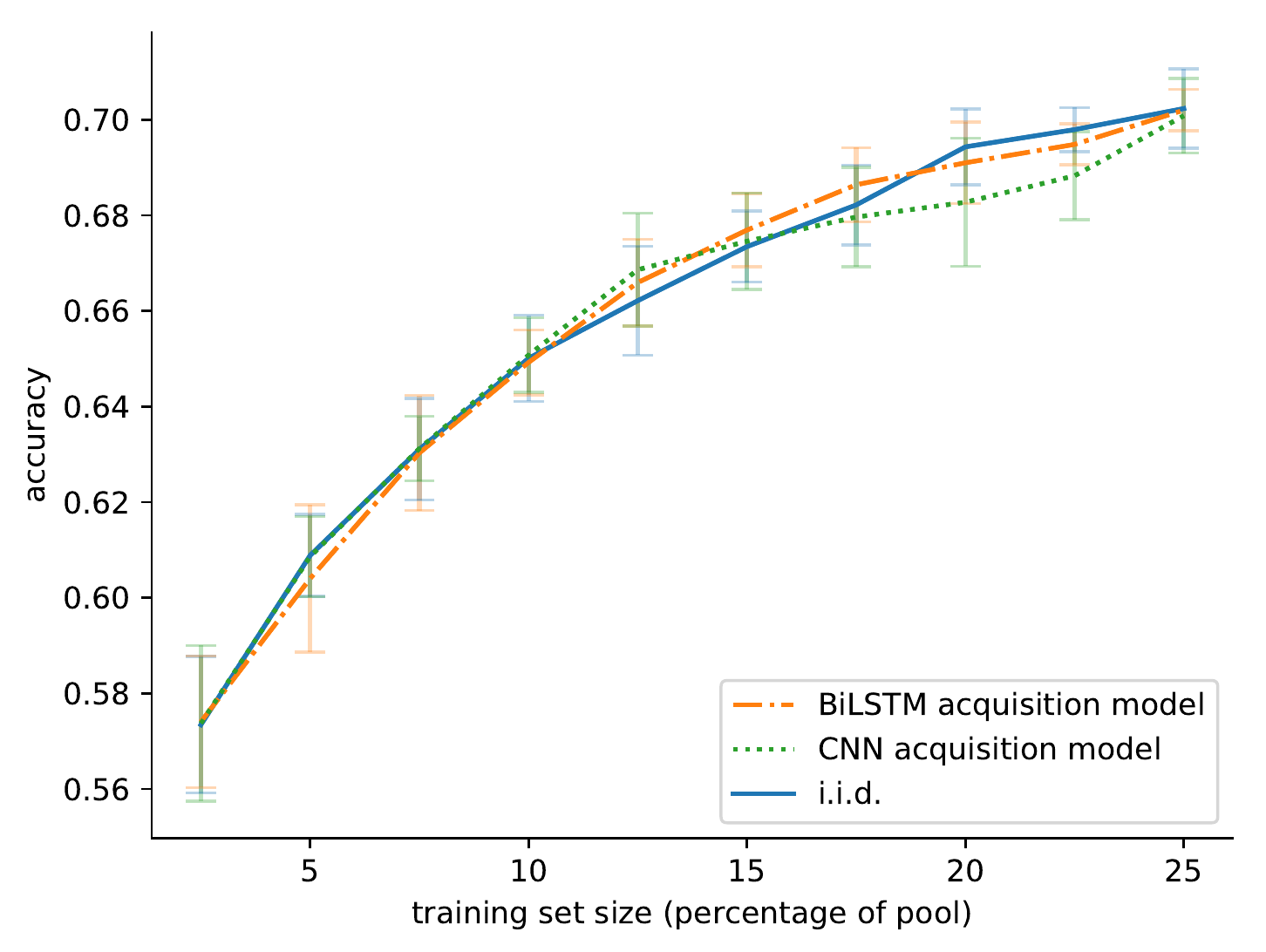}
		\caption{CNN on Movie Reviews dataset using BALD } % subcaption
  \end{subfigure}\\
  
    \begin{subfigure}[t]{0.44\linewidth} % width of left subfigure
		\includegraphics[width=\linewidth]{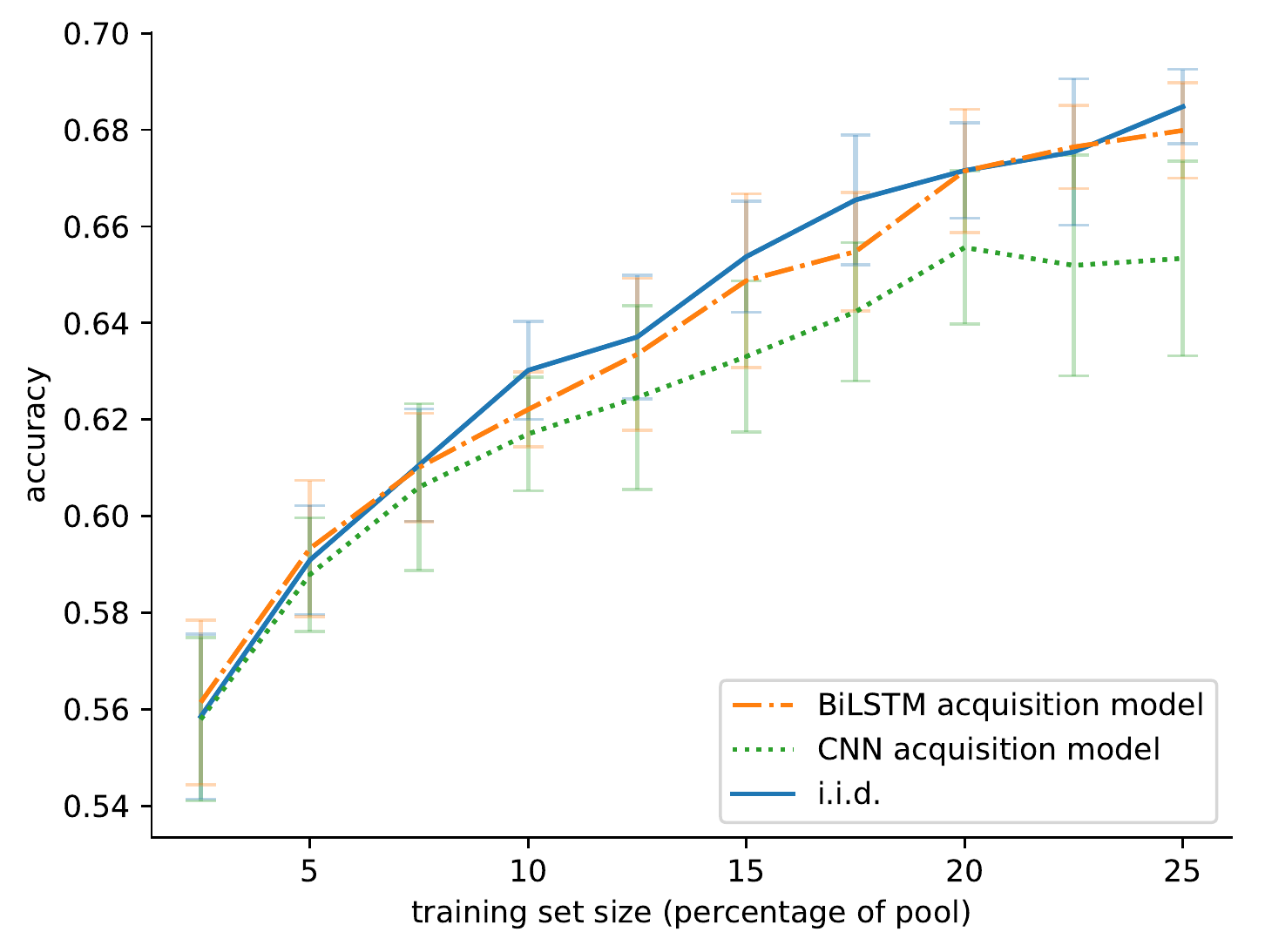}
		\caption{BiLSTM on Movie Reviews dataset using BALD } % subcaption
  \end{subfigure}
  \begin{subfigure}[t]{0.44\linewidth} % width of left subfigure
		\includegraphics[width=\linewidth]{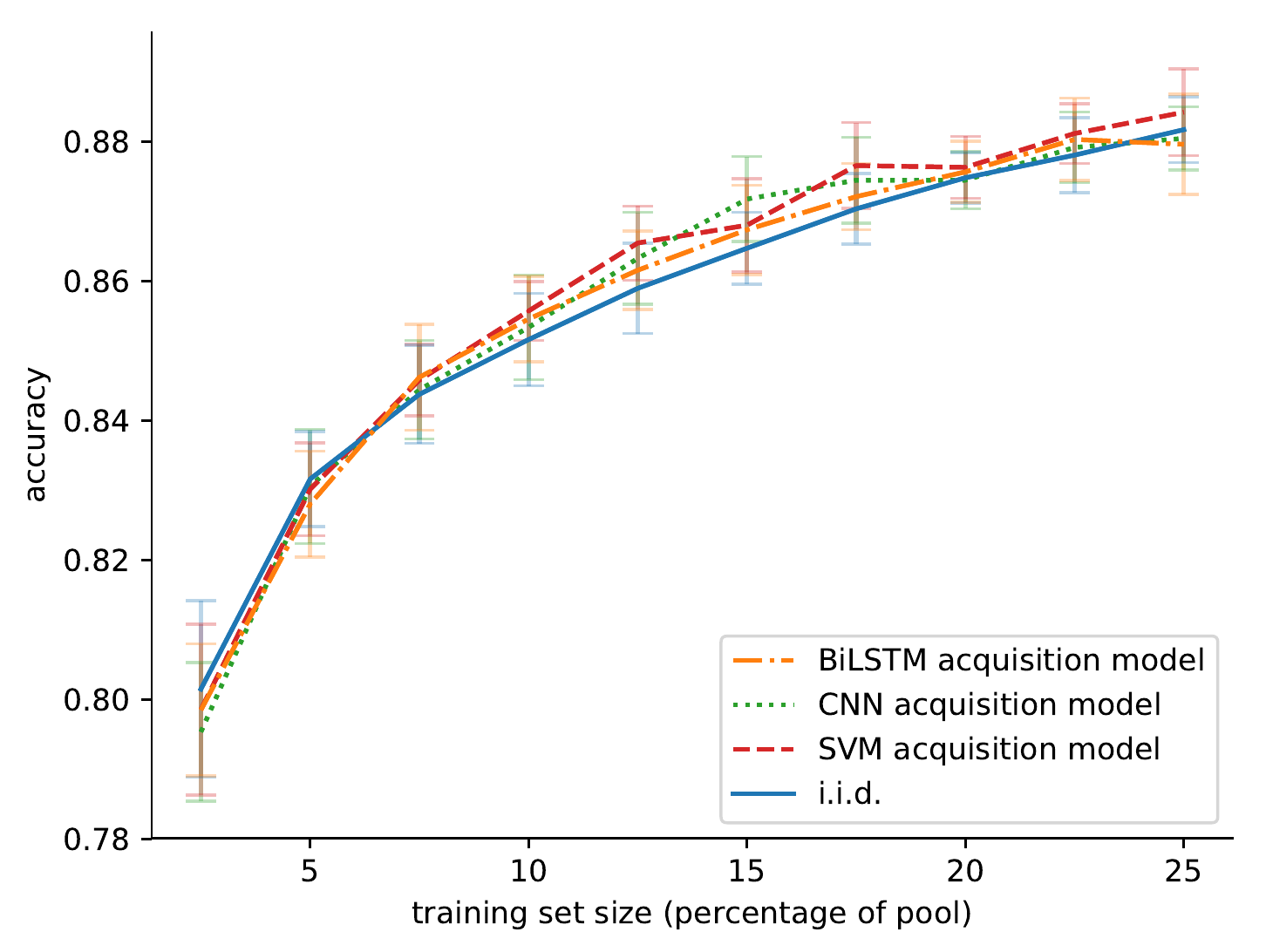}
		\caption{SVM on Subjectivity dataset using max entropy } % subcaption
  \end{subfigure}\\
  
  \begin{subfigure}[t]{0.44\linewidth} % width of left subfigure
		\includegraphics[width=\linewidth]{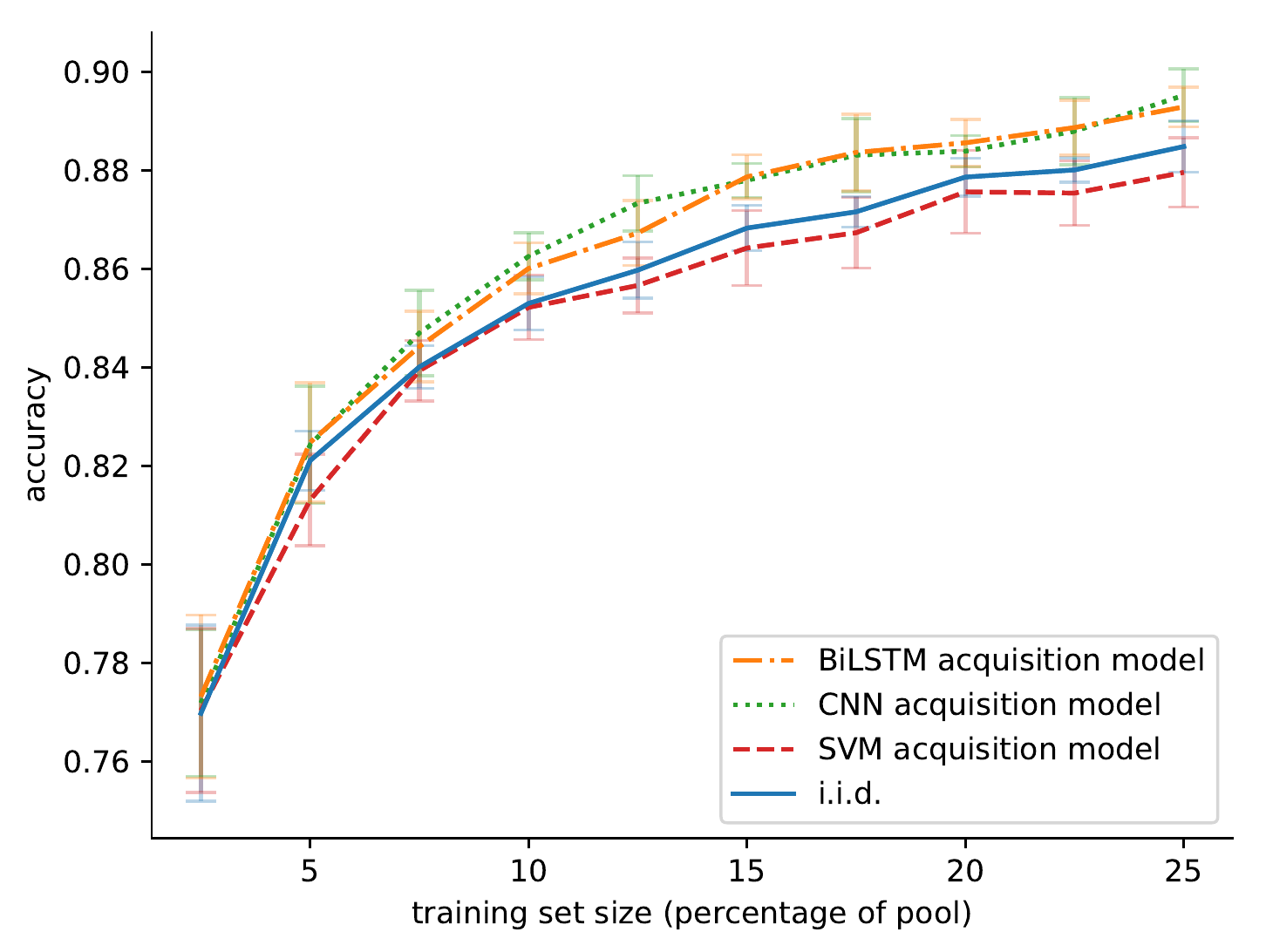}
		\caption{CNN on Subjectivity dataset using max entropy } % subcaption
  \end{subfigure}
    \begin{subfigure}[t]{0.44\linewidth} % width of left subfigure
		\includegraphics[width=\linewidth]{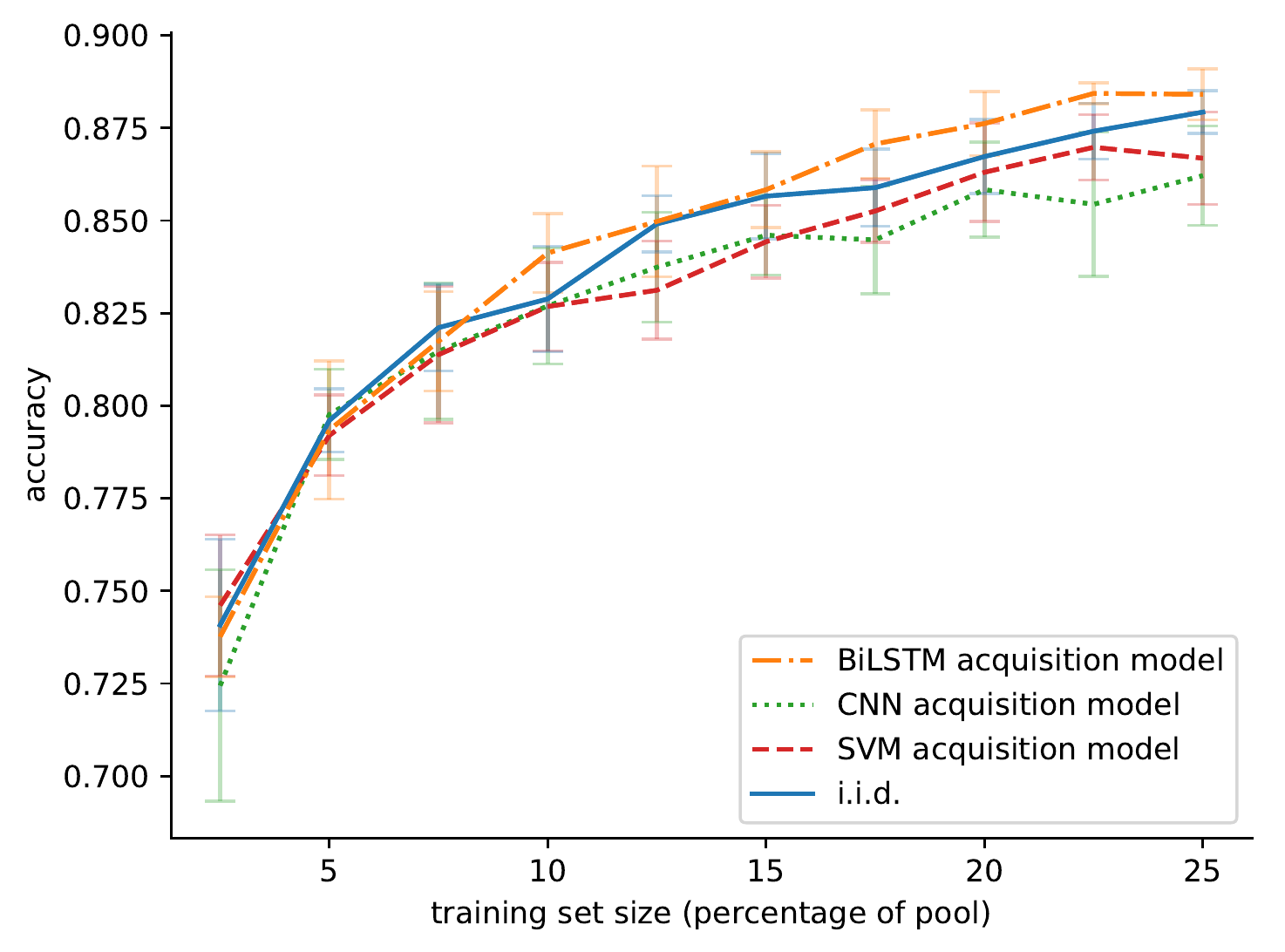}
		\caption{BiLSTM on Subjectivity dataset using max entropy } % subcaption
  \end{subfigure}\\
  
  \begin{subfigure}[t]{0.44\linewidth} % width of left subfigure
		\includegraphics[width=\linewidth]{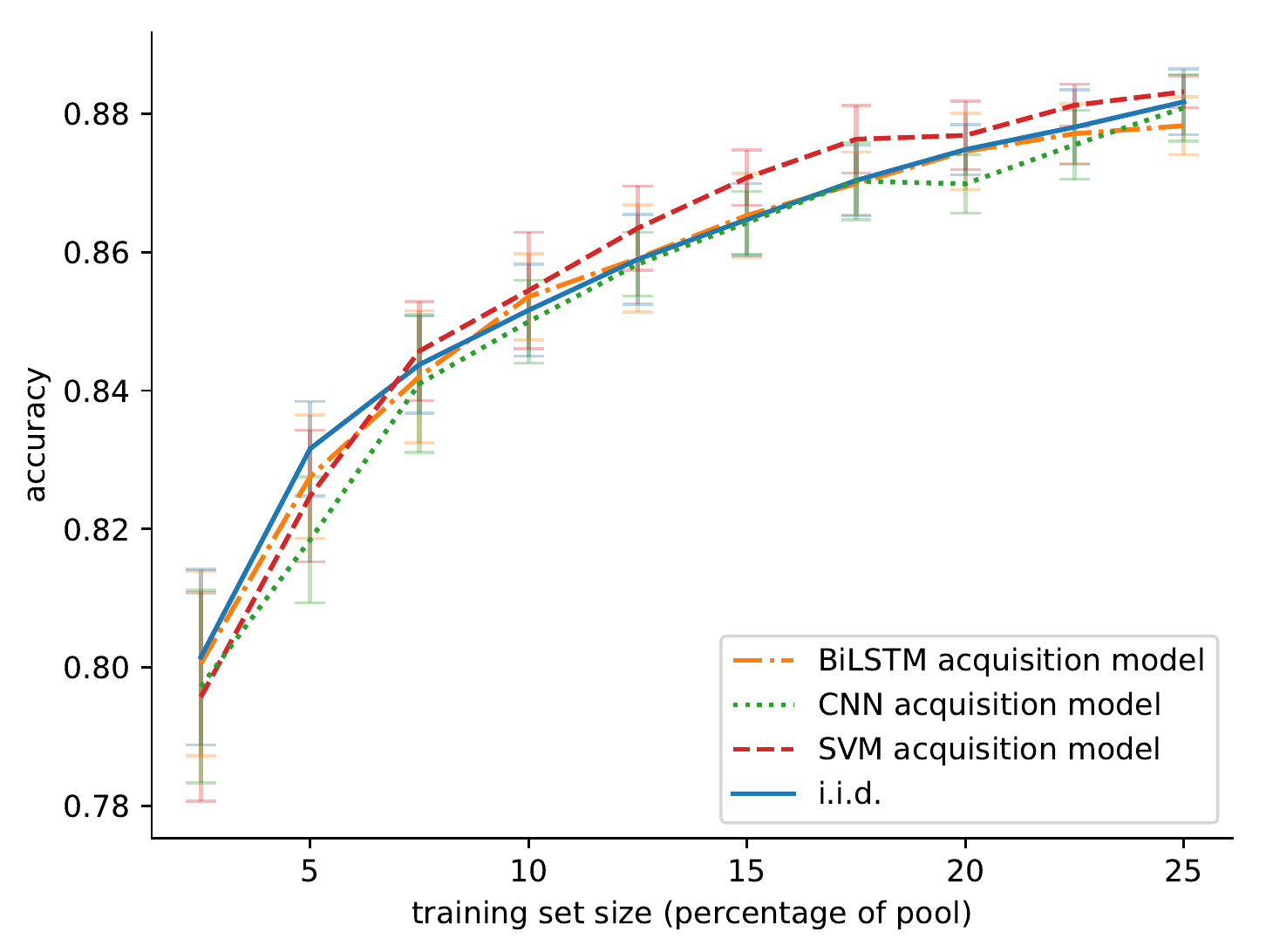}
		\caption{SVM on Subjectivity dataset using QBC } % subcaption
  \end{subfigure}
  \begin{subfigure}[t]{0.44\linewidth} % width of left subfigure
		\includegraphics[width=\linewidth]{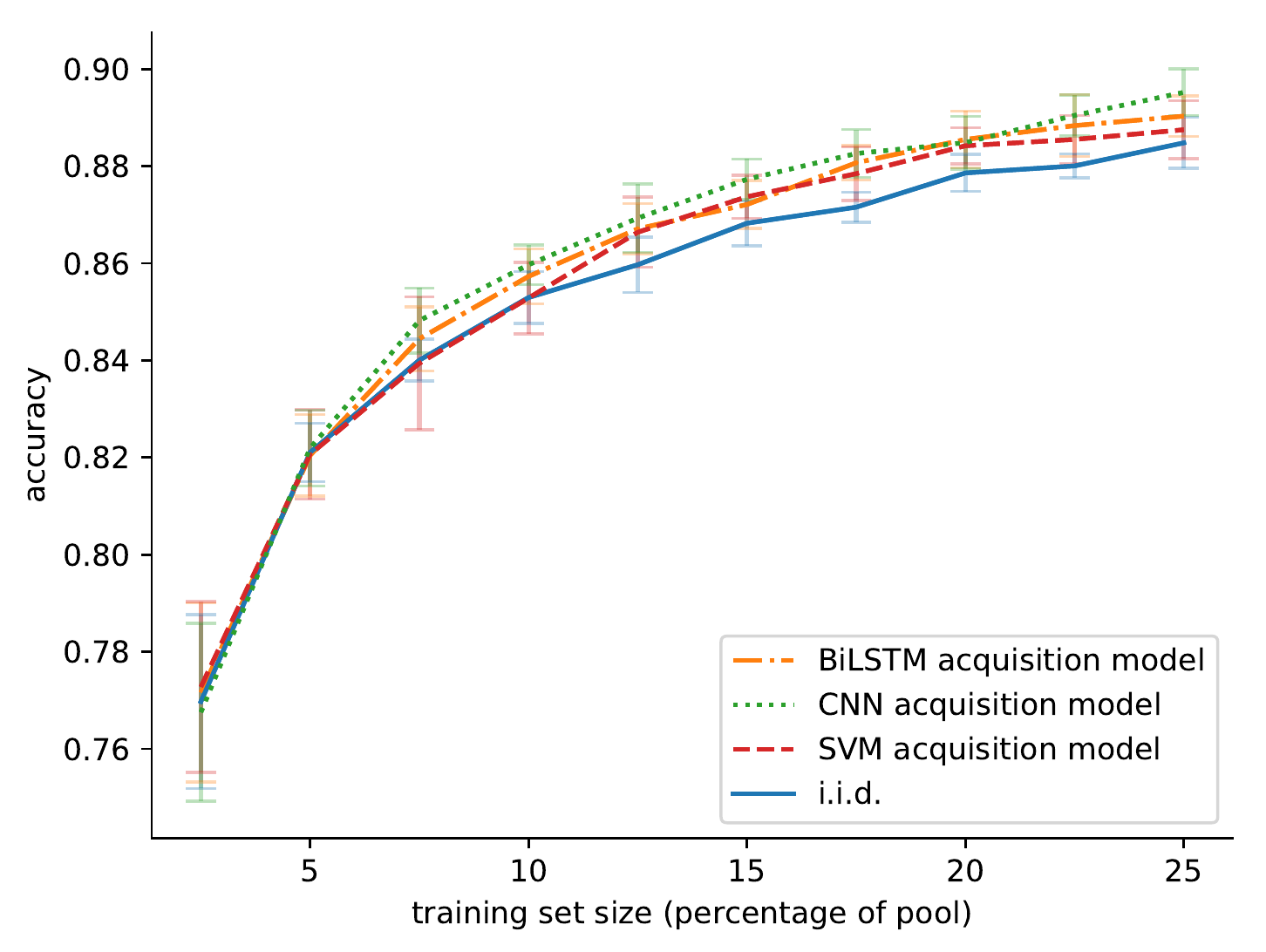}
		\caption{CNN on Subjectivity dataset using QBC } % subcaption
  \end{subfigure}\\
  
  \label{fig:appendix_pg2}
\end{figure*}
  
\begin{figure*}\ContinuedFloat
  \centering
    \begin{subfigure}[t]{0.44\linewidth} % width of left subfigure
		\includegraphics[width=\linewidth]{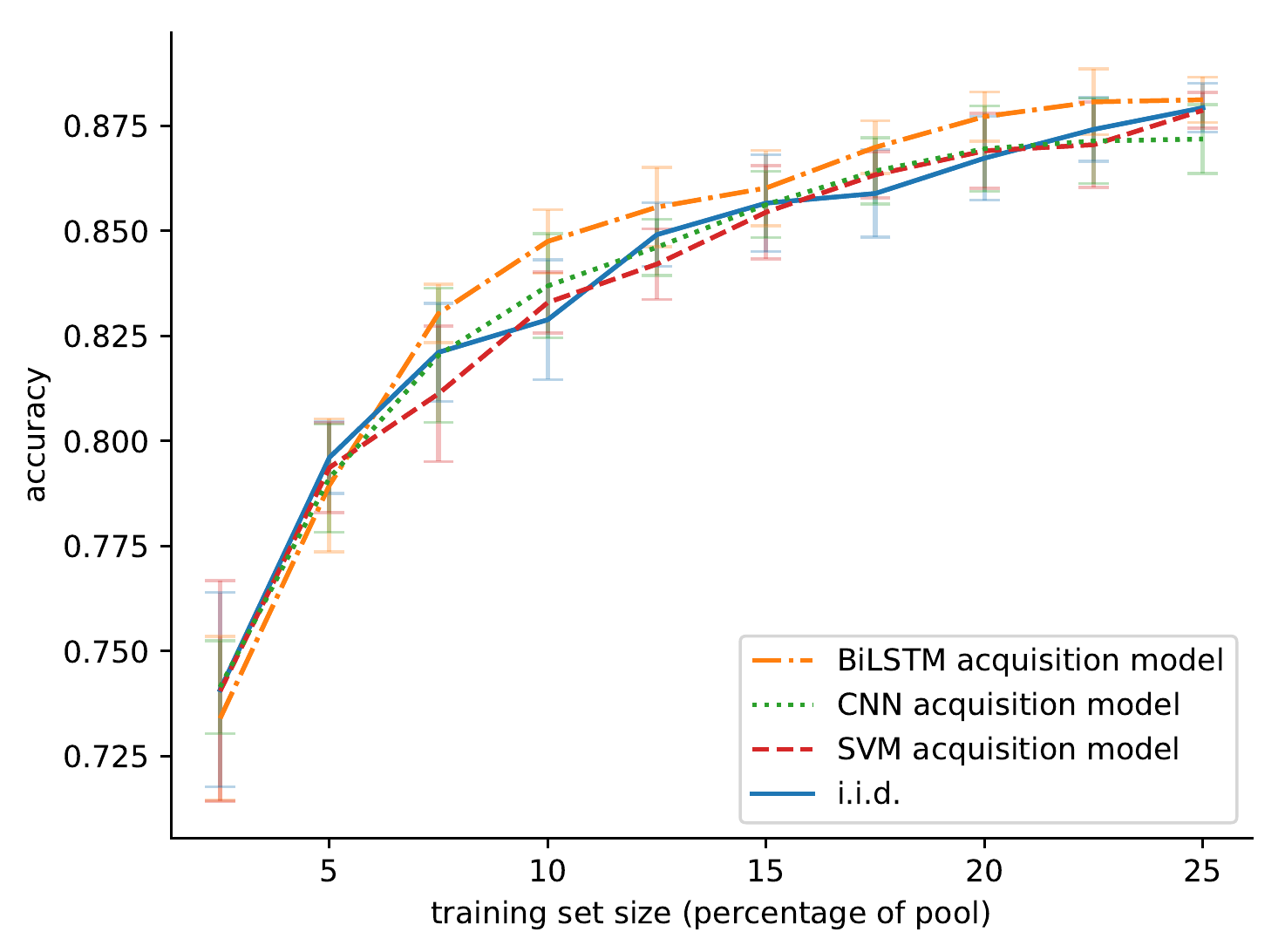}
		\caption{BiLSTM on Subjectivity dataset using QBC } % subcaption
  \end{subfigure}
  \centering
  \begin{subfigure}[t]{0.44\linewidth} % width of left subfigure
		\includegraphics[width=\linewidth]{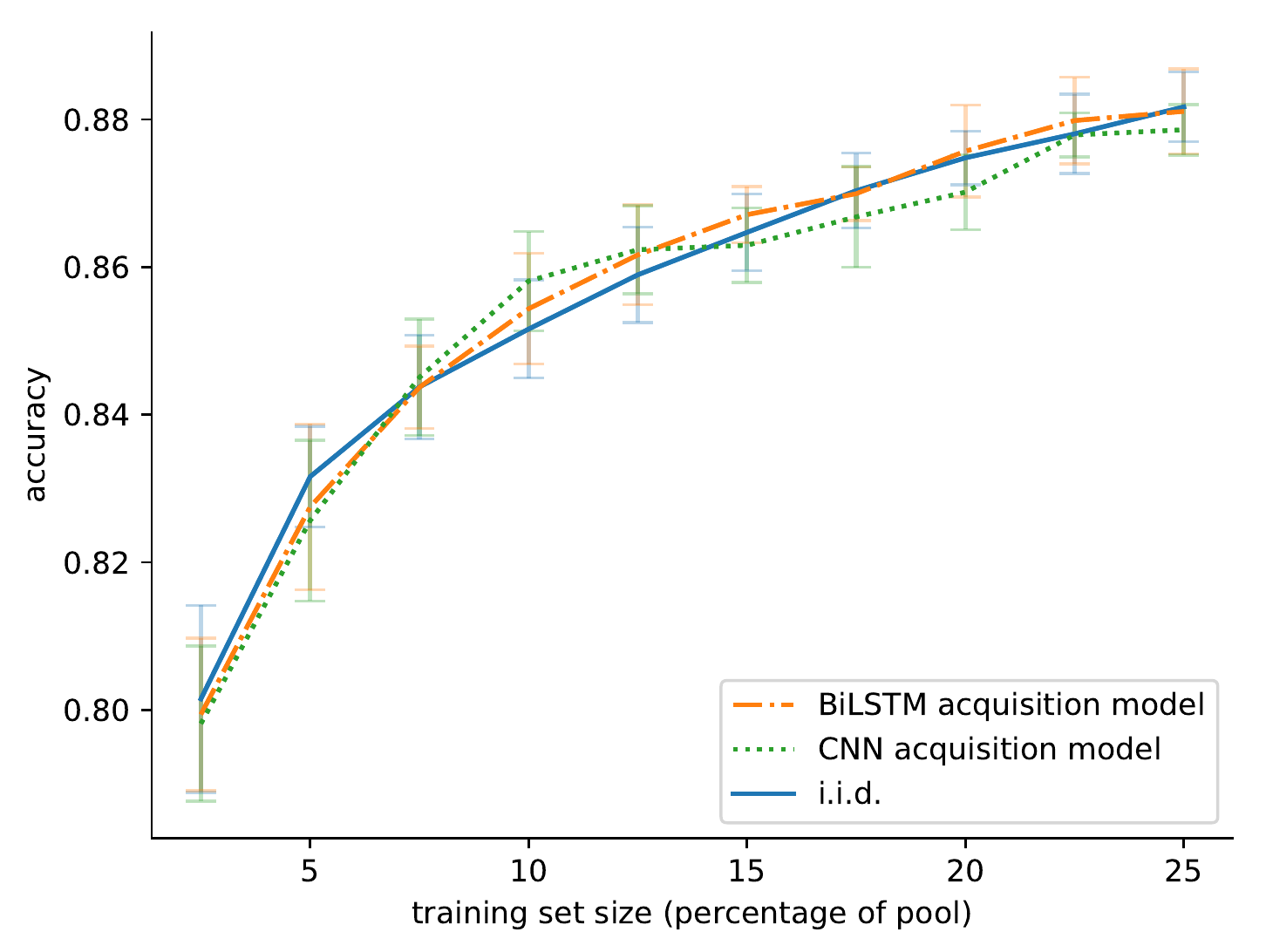}
		\caption{SVM on Subjectivity dataset using BALD } % subcaption
  \end{subfigure}\\
  
  \begin{subfigure}[t]{0.44\linewidth} % width of left subfigure
		\includegraphics[width=\linewidth]{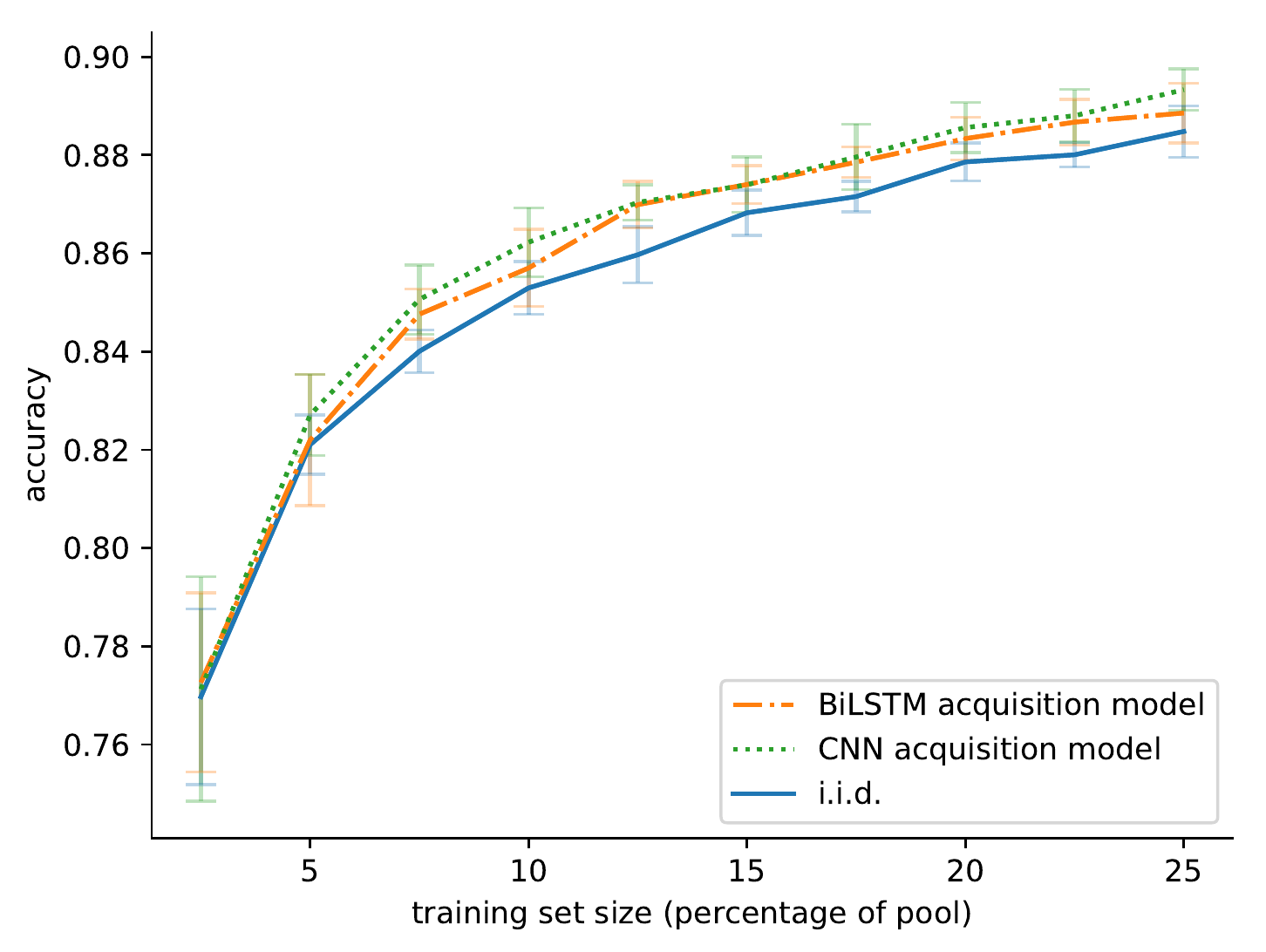}
		\caption{CNN on Subjectivity dataset using BALD } % subcaption
  \end{subfigure}
    \begin{subfigure}[t]{0.44\linewidth} % width of left subfigure
		\includegraphics[width=\linewidth]{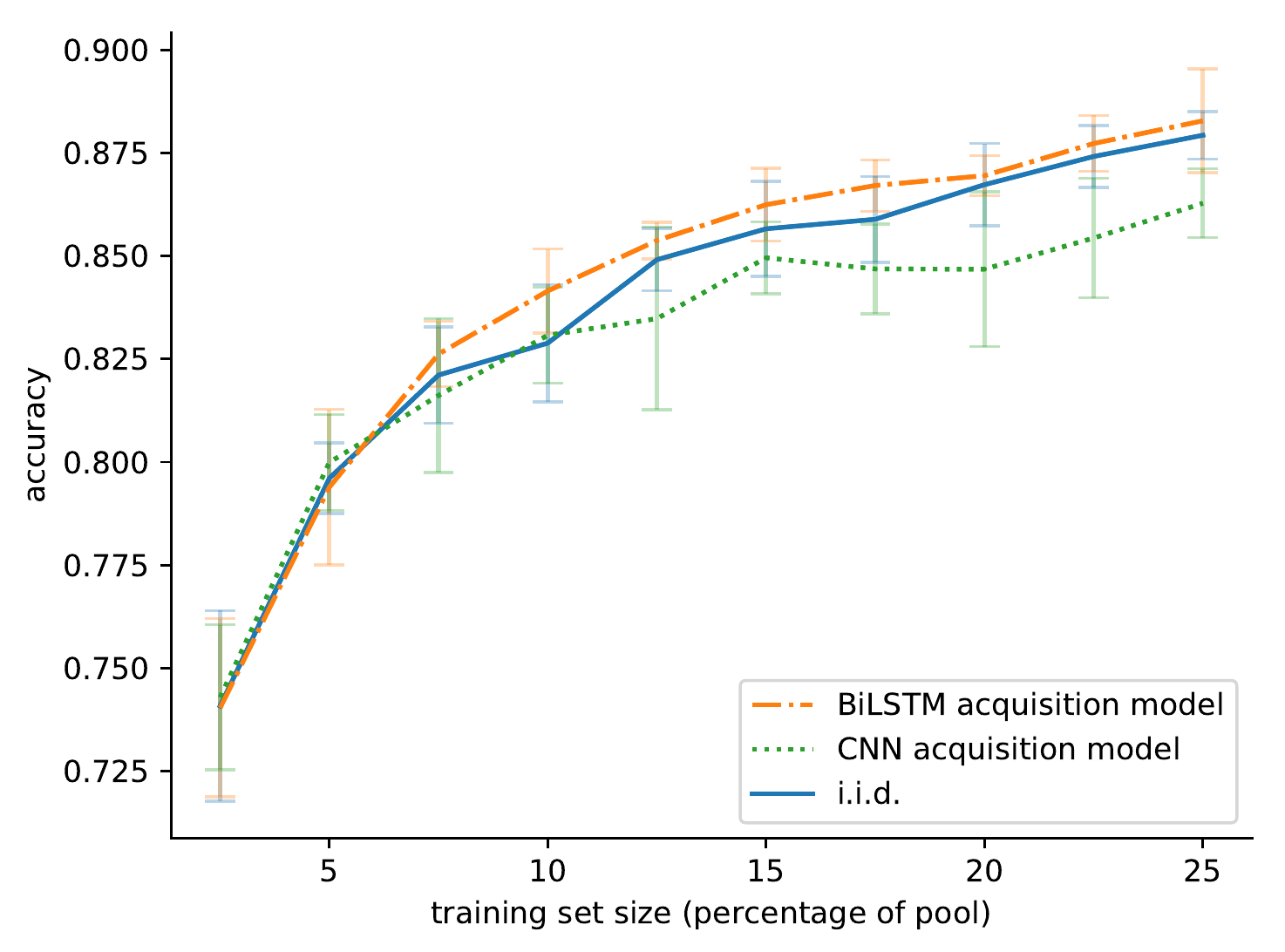}
		\caption{BiLSTM on Subjectivity dataset using BALD } % subcaption
  \end{subfigure}\\
 
   \begin{subfigure}[t]{0.44\linewidth} % width of left subfigure
		\includegraphics[width=\linewidth]{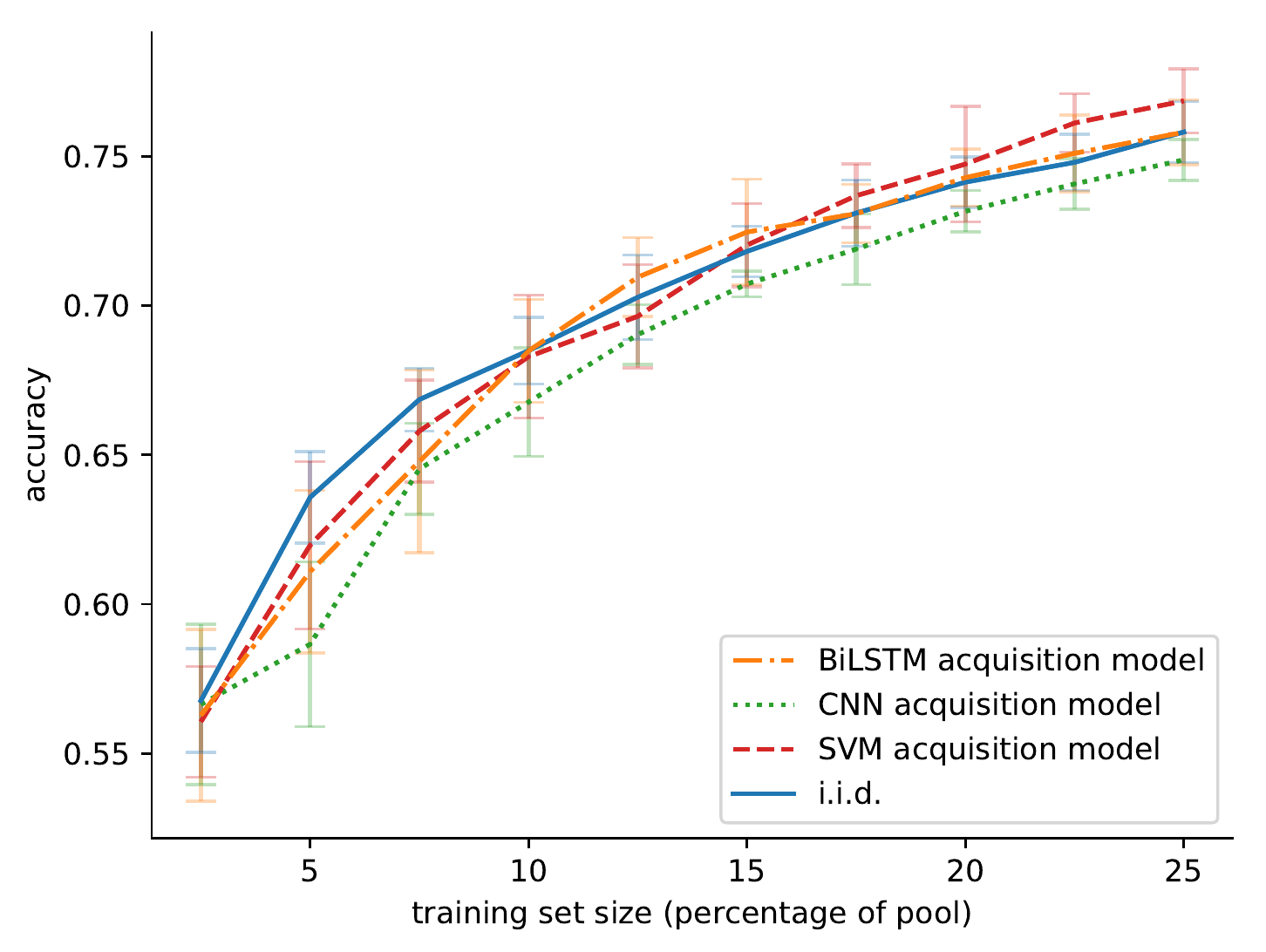}
		\caption{SVM on TREC dataset using max entropy } % subcaption
  \end{subfigure}
  \begin{subfigure}[t]{0.44\linewidth} % width of left subfigure
		\includegraphics[width=\linewidth]{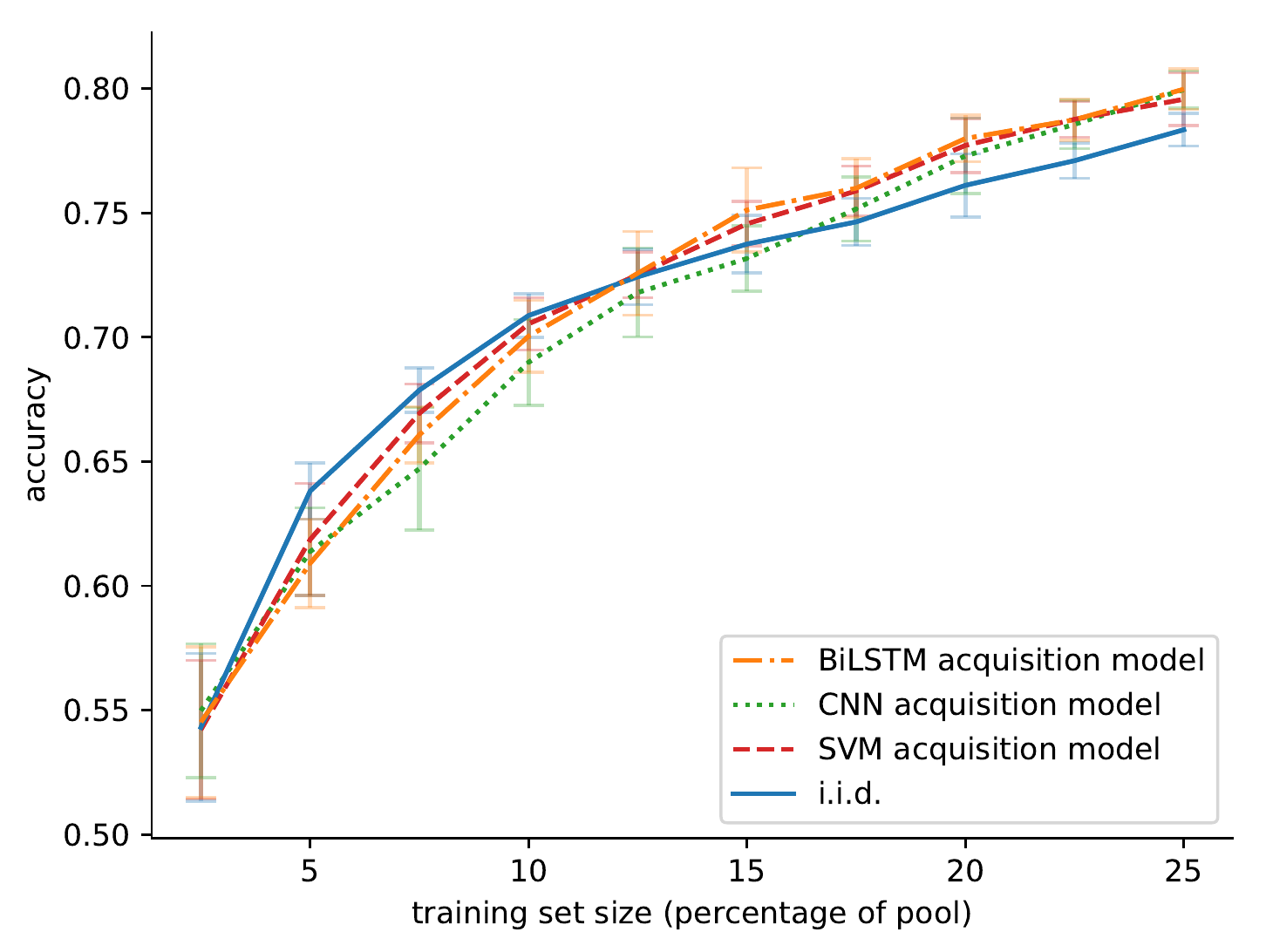}
		\caption{CNN on TREC dataset using max entropy } % subcaption
  \end{subfigure}\\
  
    \begin{subfigure}[t]{0.44\linewidth} % width of left subfigure
		\includegraphics[width=\linewidth]{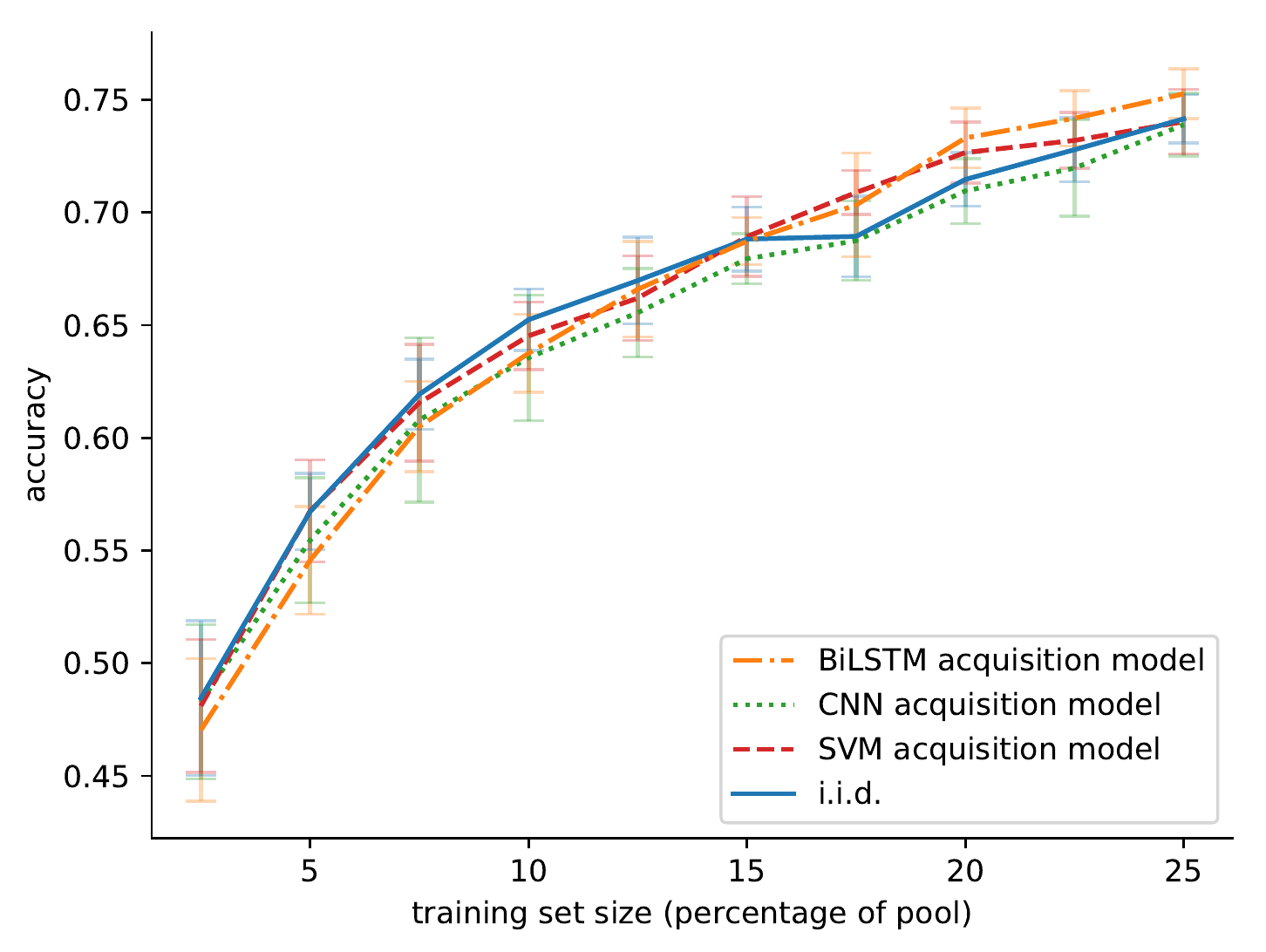}
		\caption{BiLSTM on TREC dataset using max entropy } % subcaption
  \end{subfigure}
  \begin{subfigure}[t]{0.44\linewidth} % width of left subfigure
		\includegraphics[width=\linewidth]{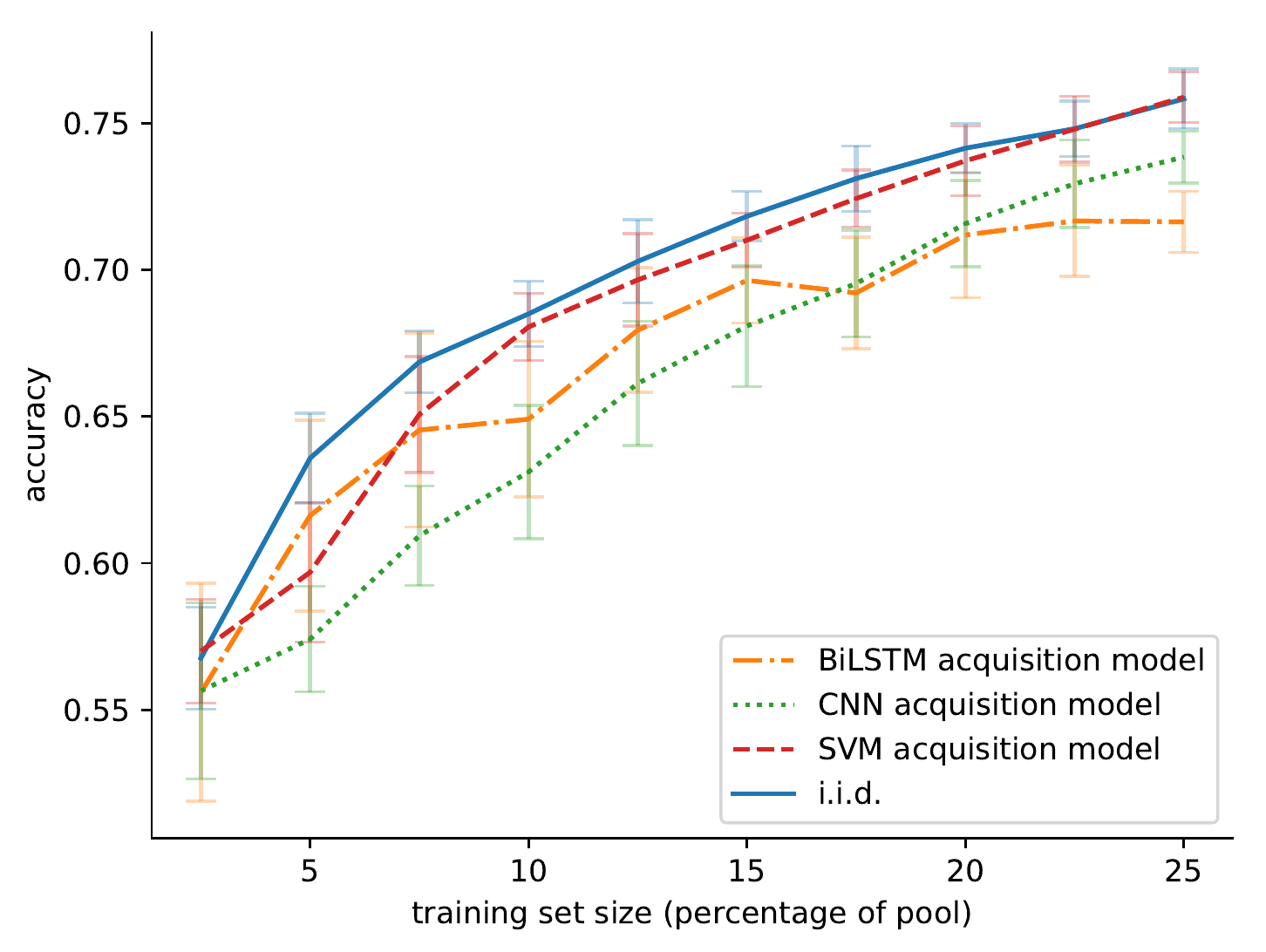}
		\caption{SVM on TREC dataset using QBC } % subcaption
  \end{subfigure}\\
  
  \label{fig:appendix_pg3}
\end{figure*}
  
\begin{figure*}\ContinuedFloat
  \centering
  
  \begin{subfigure}[t]{0.44\linewidth} % width of left subfigure
		\includegraphics[width=\linewidth]{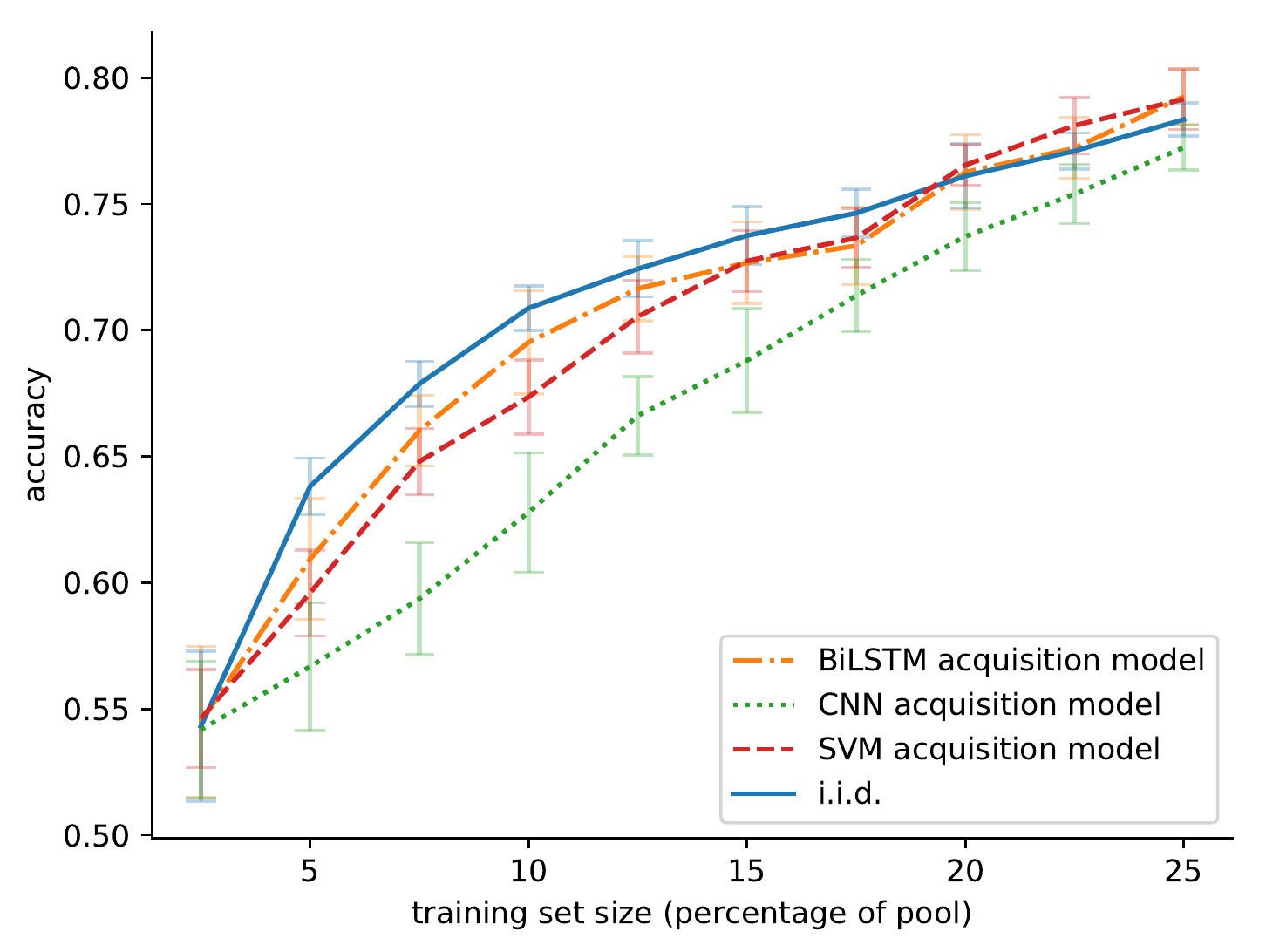}
		\caption{CNN on TREC dataset using QBC } % subcaption
  \end{subfigure}
  \begin{subfigure}[t]{0.44\linewidth} % width of left subfigure
		\includegraphics[width=\linewidth]{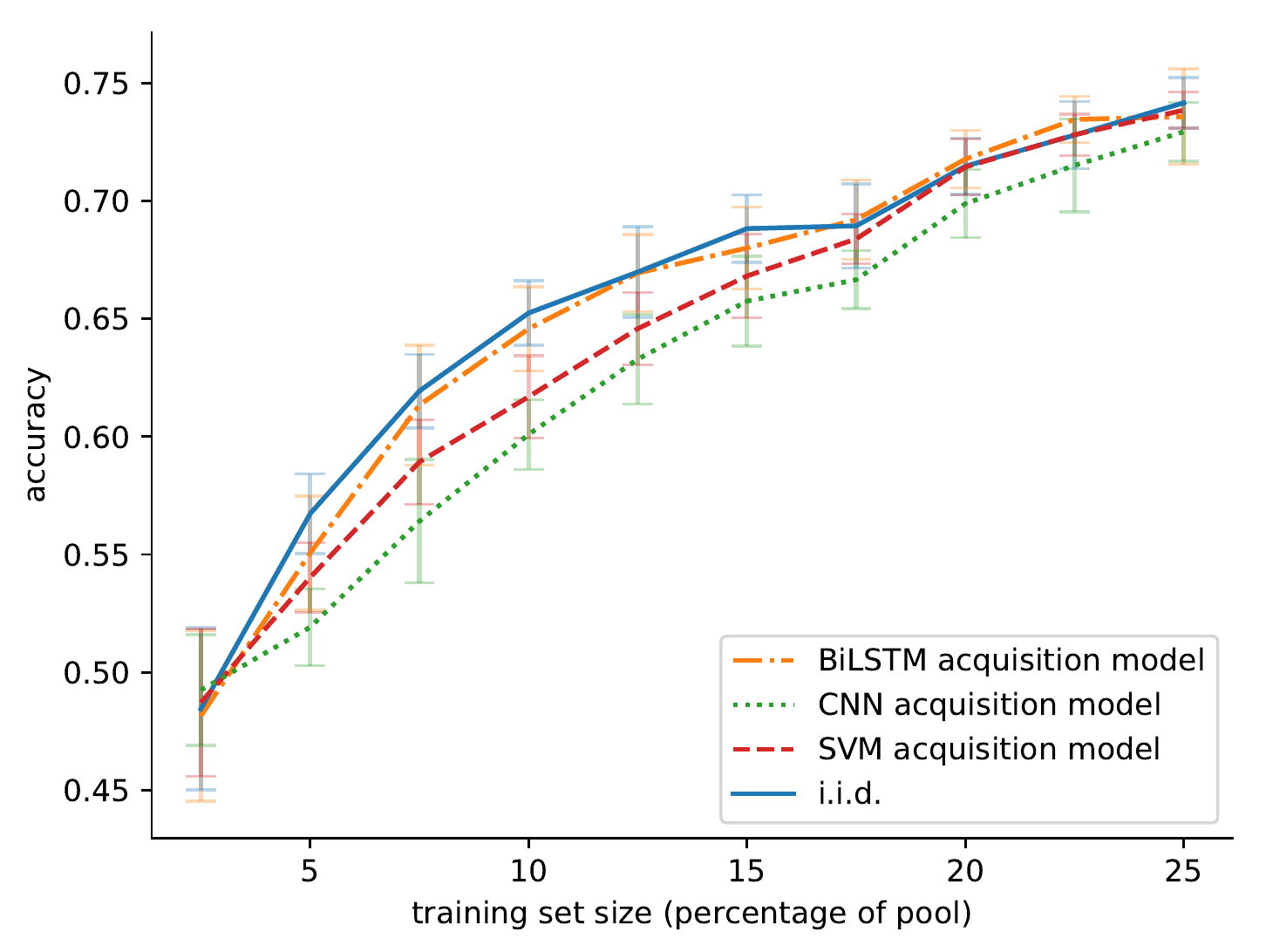}
		\caption{BiLSTM on TREC dataset using QBC } % subcaption
  \end{subfigure}\\
  
  \begin{subfigure}[t]{0.44\linewidth} % width of left subfigure
		\includegraphics[width=\linewidth]{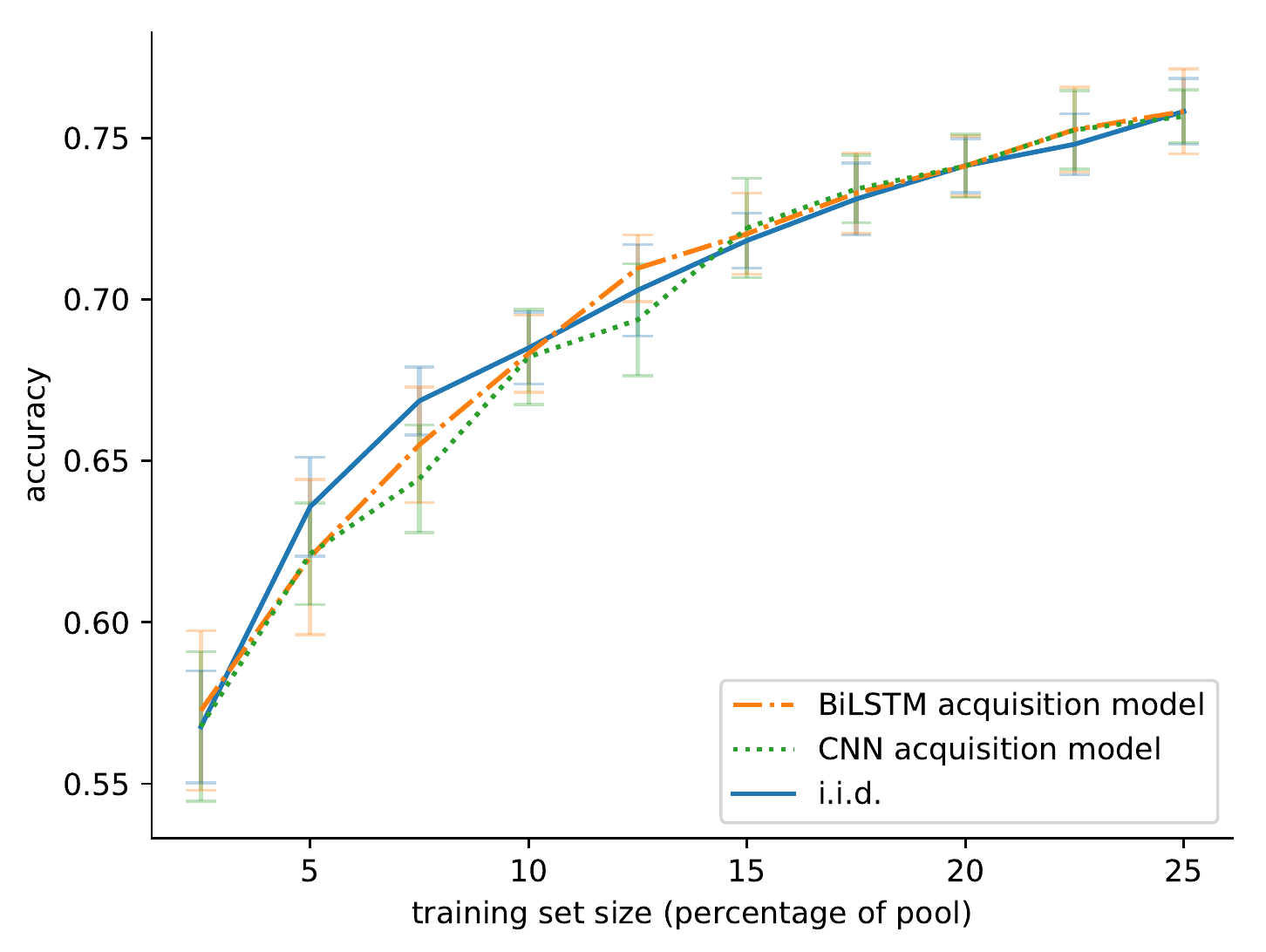}
		\caption{SVM on TREC dataset using BALD } % subcaption
  \end{subfigure}
  \begin{subfigure}[t]{0.44\linewidth} % width of left subfigure
		\includegraphics[width=\linewidth]{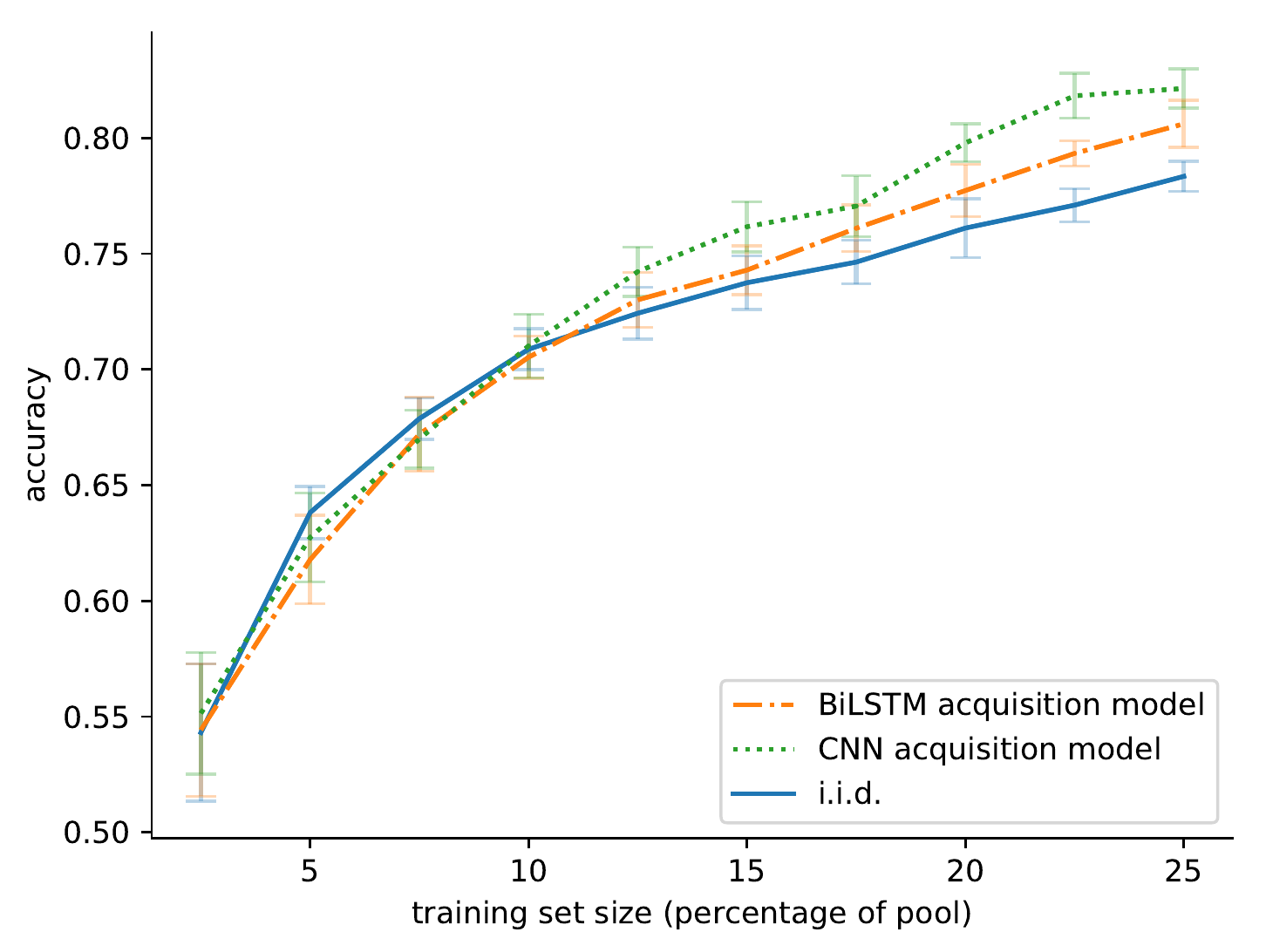}
		\caption{CNN on TREC dataset using BALD } % subcaption
  \end{subfigure}\\
  
    \begin{subfigure}[t]{0.44\linewidth} % width of left subfigure
		\includegraphics[width=\linewidth]{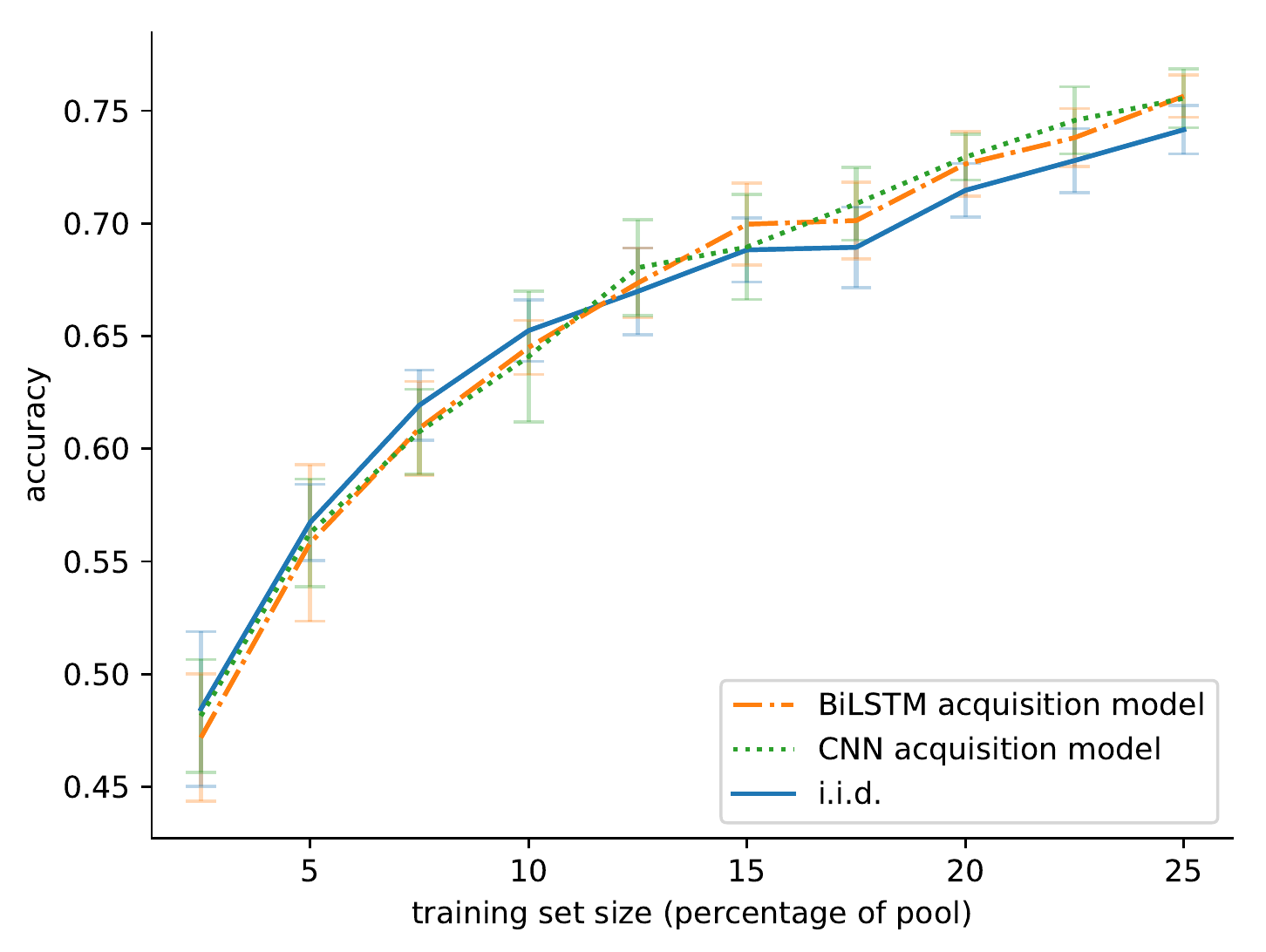}
		\caption{BiLSTM on TREC dataset using BALD } % subcaption
  \end{subfigure}
 \begin{subfigure}[t]{0.44\linewidth} % width of left subfigure
		\includegraphics[width=\linewidth]{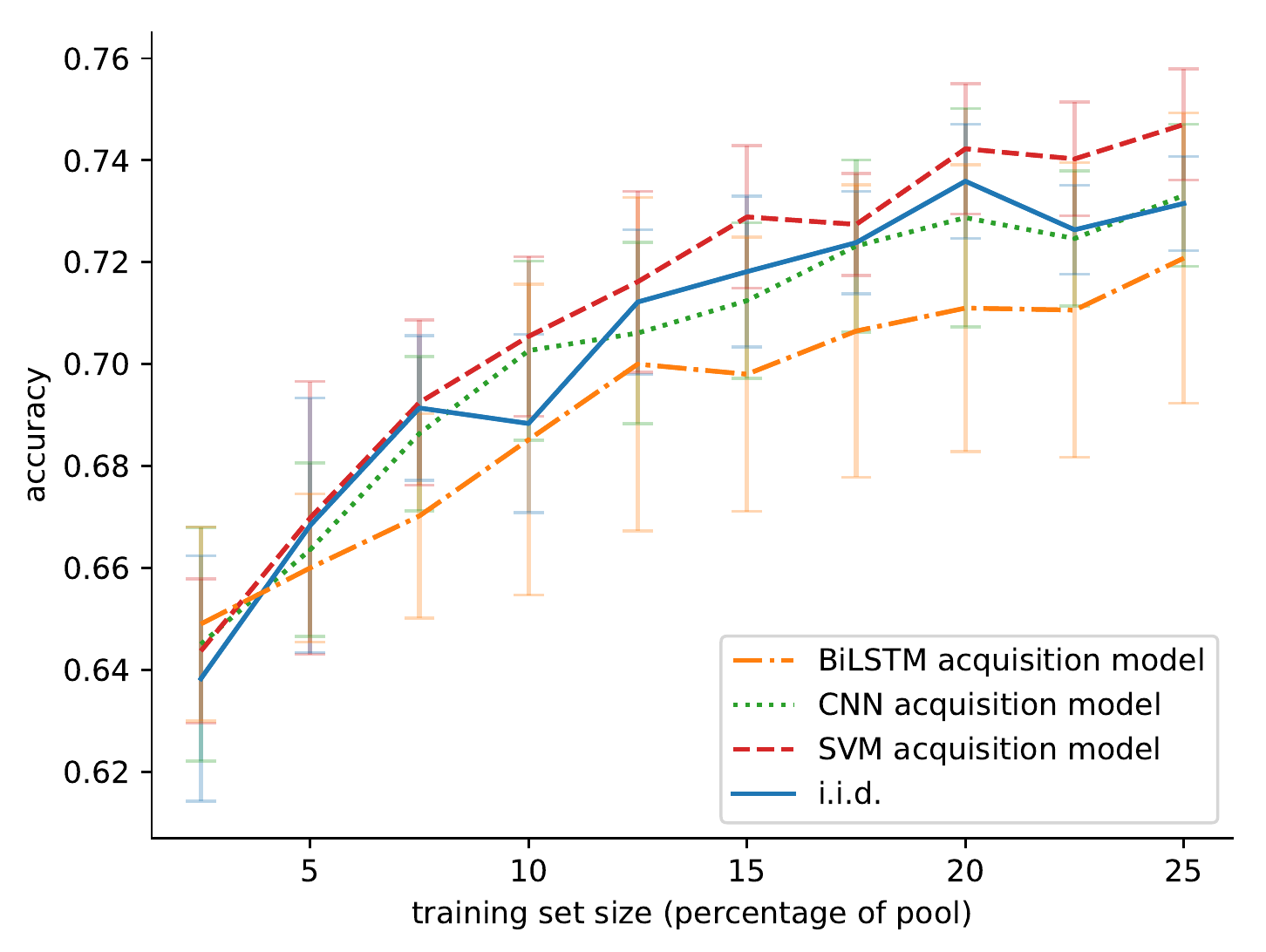}
		\caption{SVM on Customer Review dataset using max entropy } % subcaption
  \end{subfigure}\\
  
  \begin{subfigure}[t]{0.44\linewidth} % width of left subfigure
		\includegraphics[width=\linewidth]{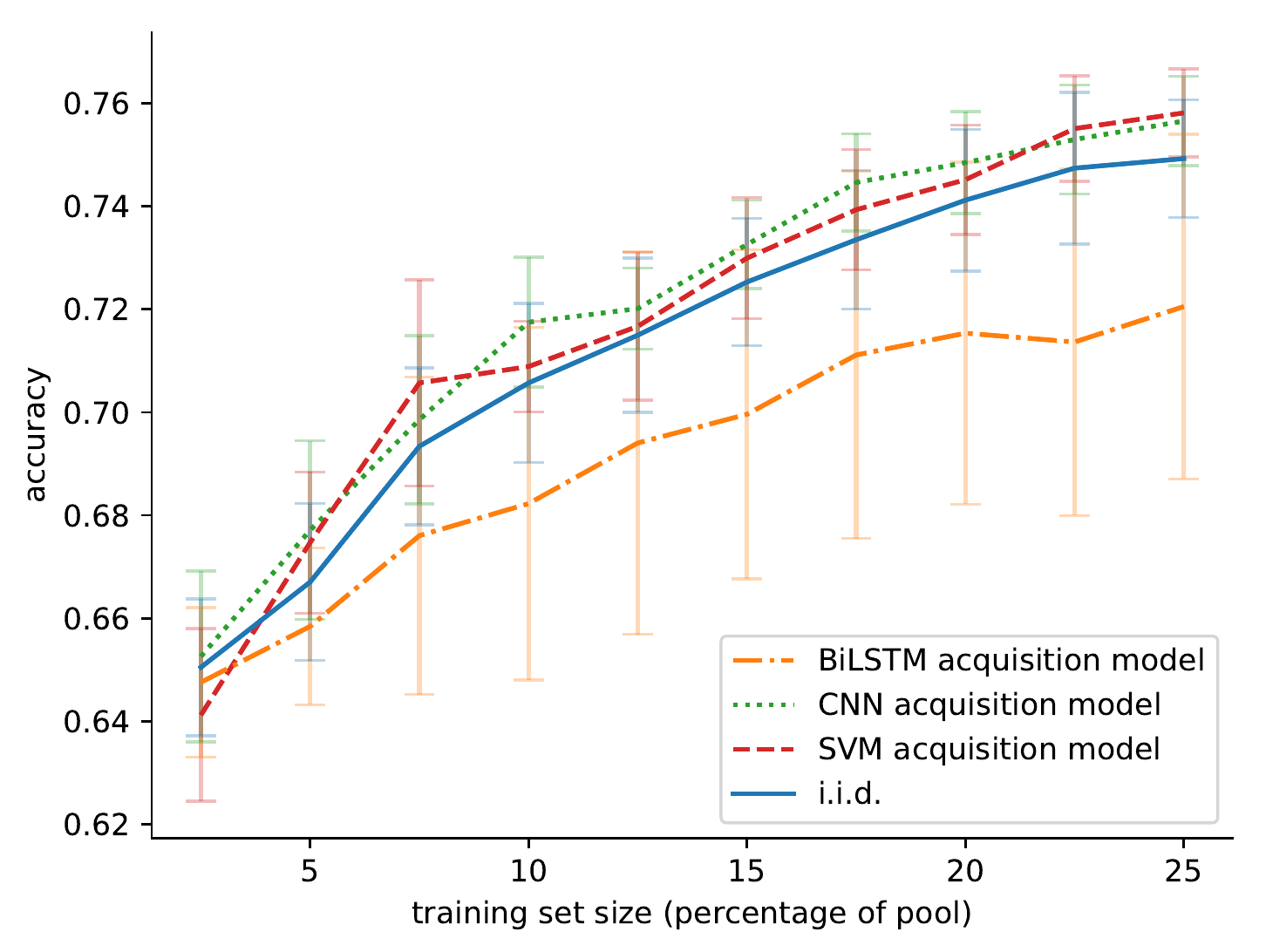}
		\caption{CNN on Customer Review dataset using max entropy } % subcaption
  \end{subfigure}
    \begin{subfigure}[t]{0.44\linewidth} % width of left subfigure
		\includegraphics[width=\linewidth]{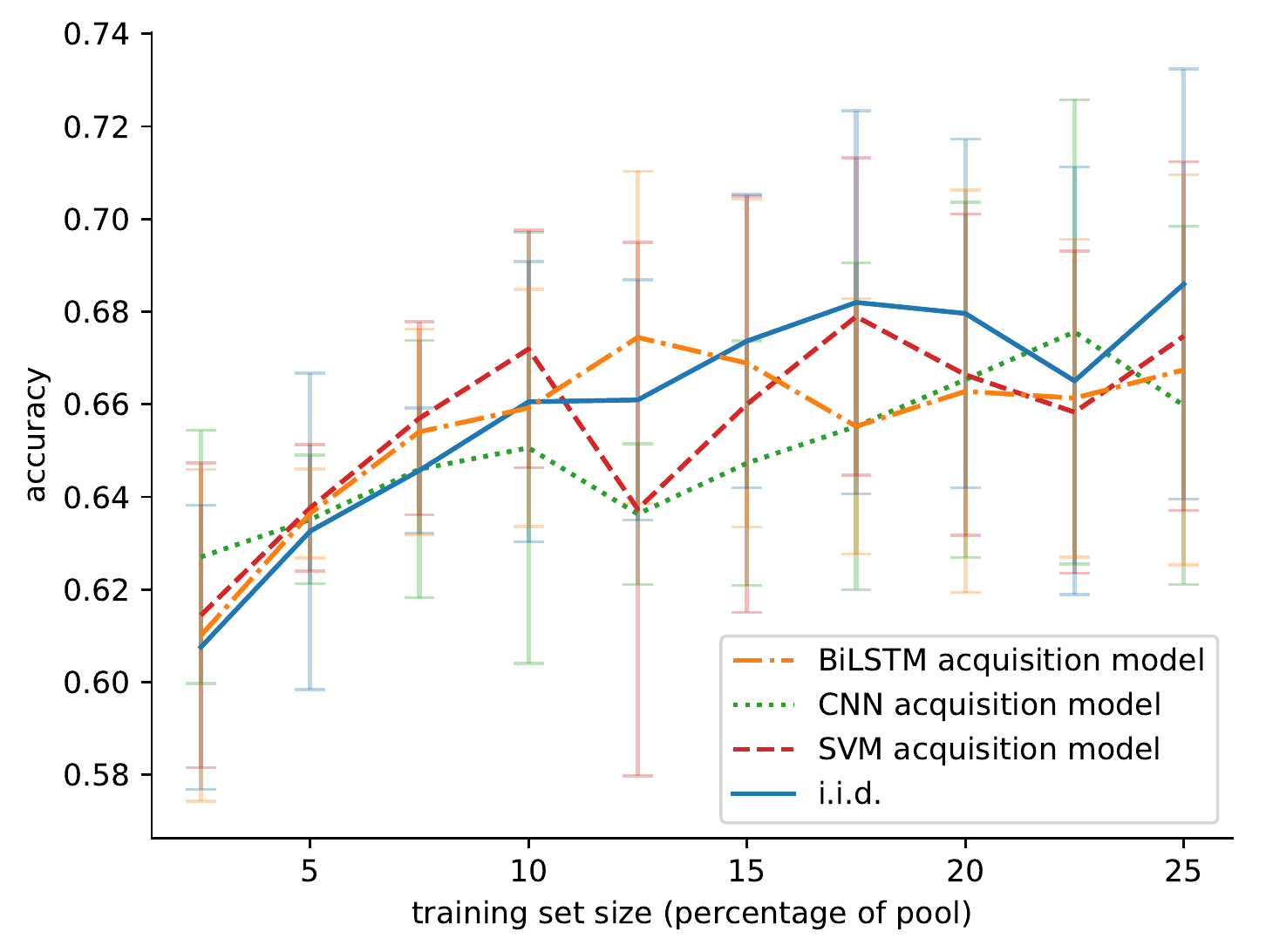}
		\caption{BiLSTM on Customer Review dataset using max entropy } % subcaption
  \end{subfigure}\\
  
  \label{fig:appendix_pg4}
\end{figure*} 
  
  \begin{figure*}\ContinuedFloat
  \centering
  \begin{subfigure}[t]{0.44\linewidth} % width of left subfigure
		\includegraphics[width=\linewidth]{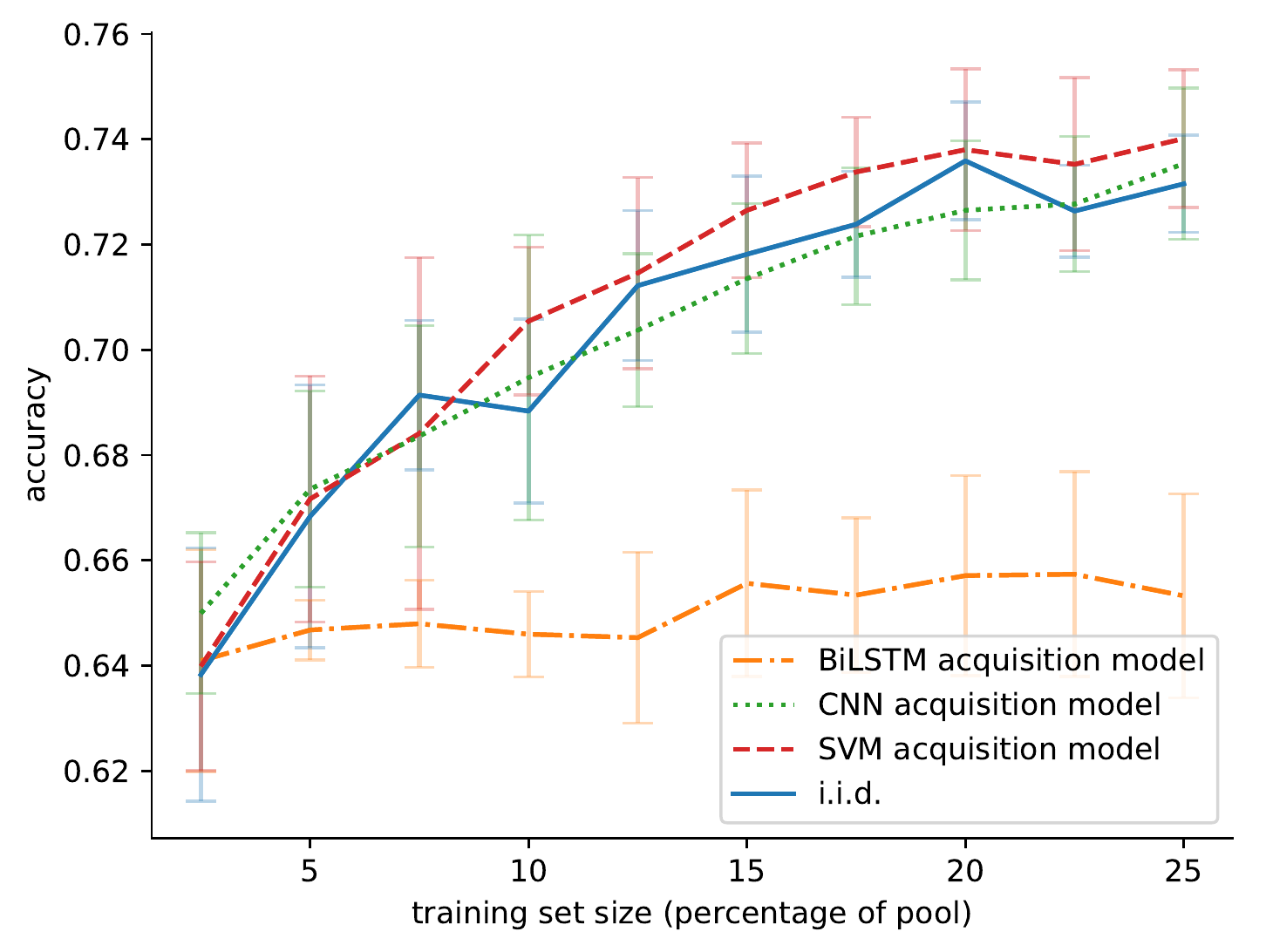}
		\caption{SVM on Customer Review dataset using QBC } % subcaption
  \end{subfigure}
  \begin{subfigure}[t]{0.44\linewidth} % width of left subfigure
		\includegraphics[width=\linewidth]{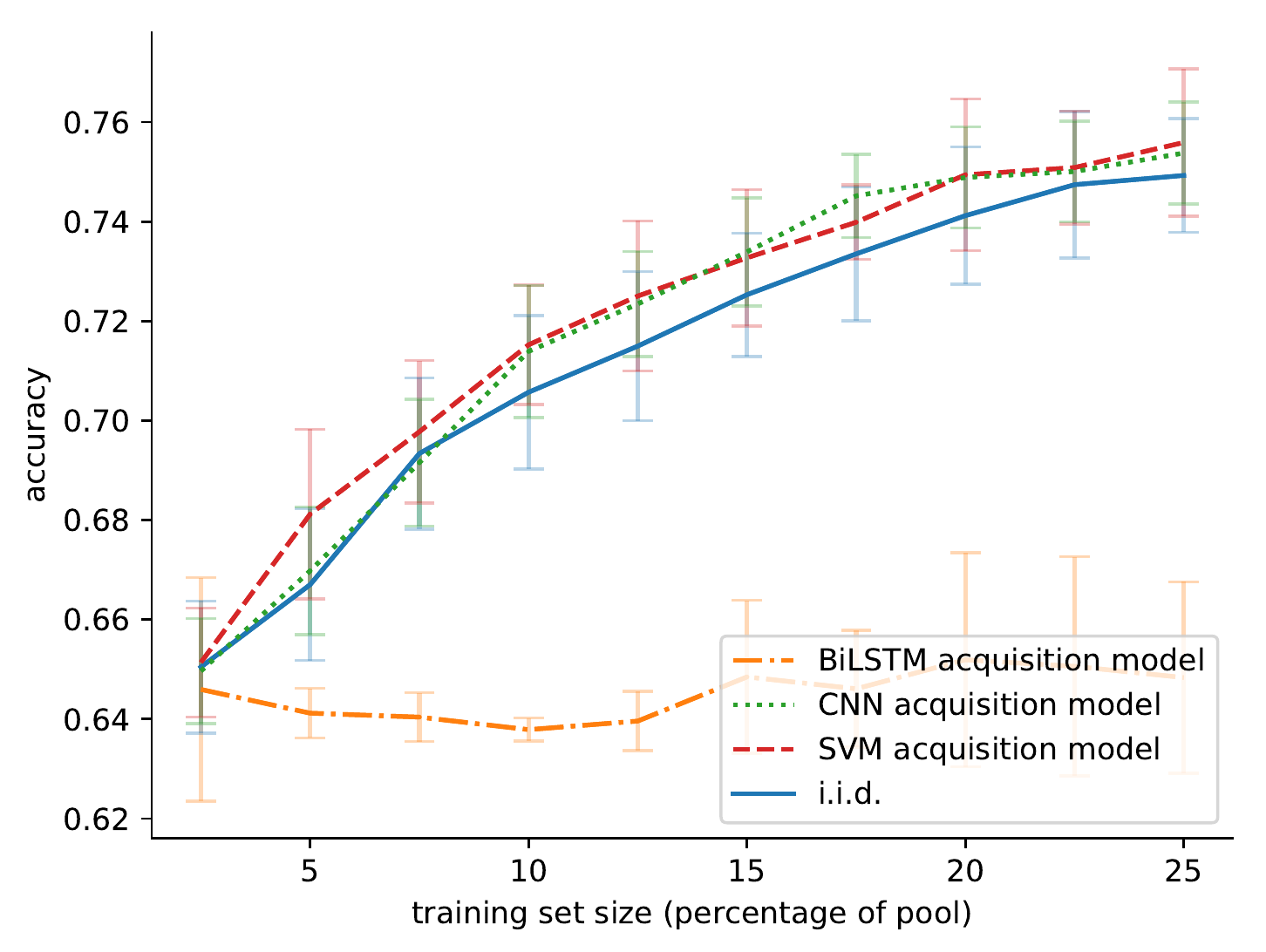}
		\caption{CNN on Customer Review dataset using QBC } % subcaption
  \end{subfigure}\\
  
  \begin{subfigure}[t]{0.44\linewidth} % width of left subfigure
		\includegraphics[width=\linewidth]{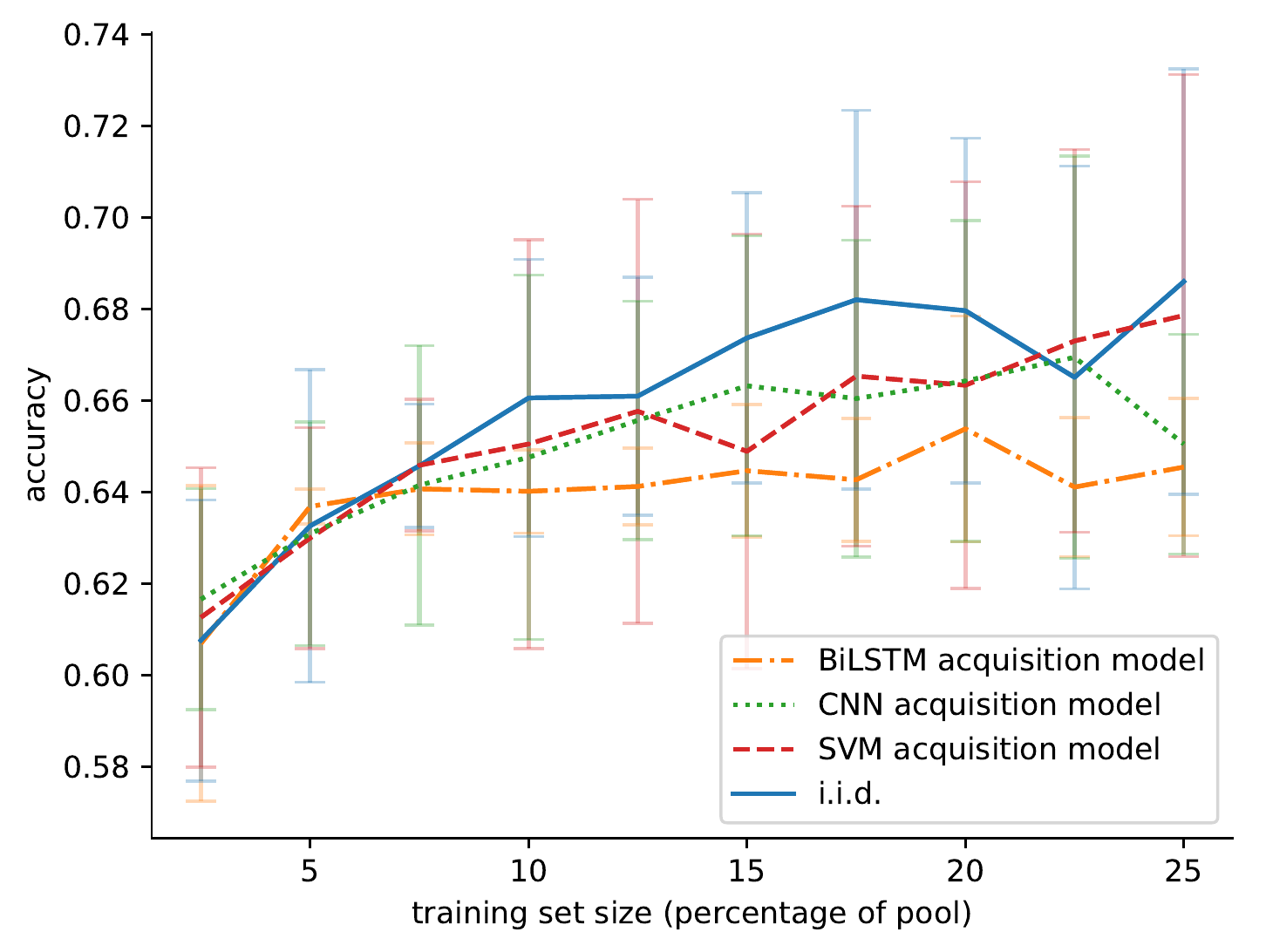}
		\caption{BiLSTM on Customer Review dataset using QBC } % subcaption
  \end{subfigure}
  \begin{subfigure}[t]{0.44\linewidth} % width of left subfigure
		\includegraphics[width=\linewidth]{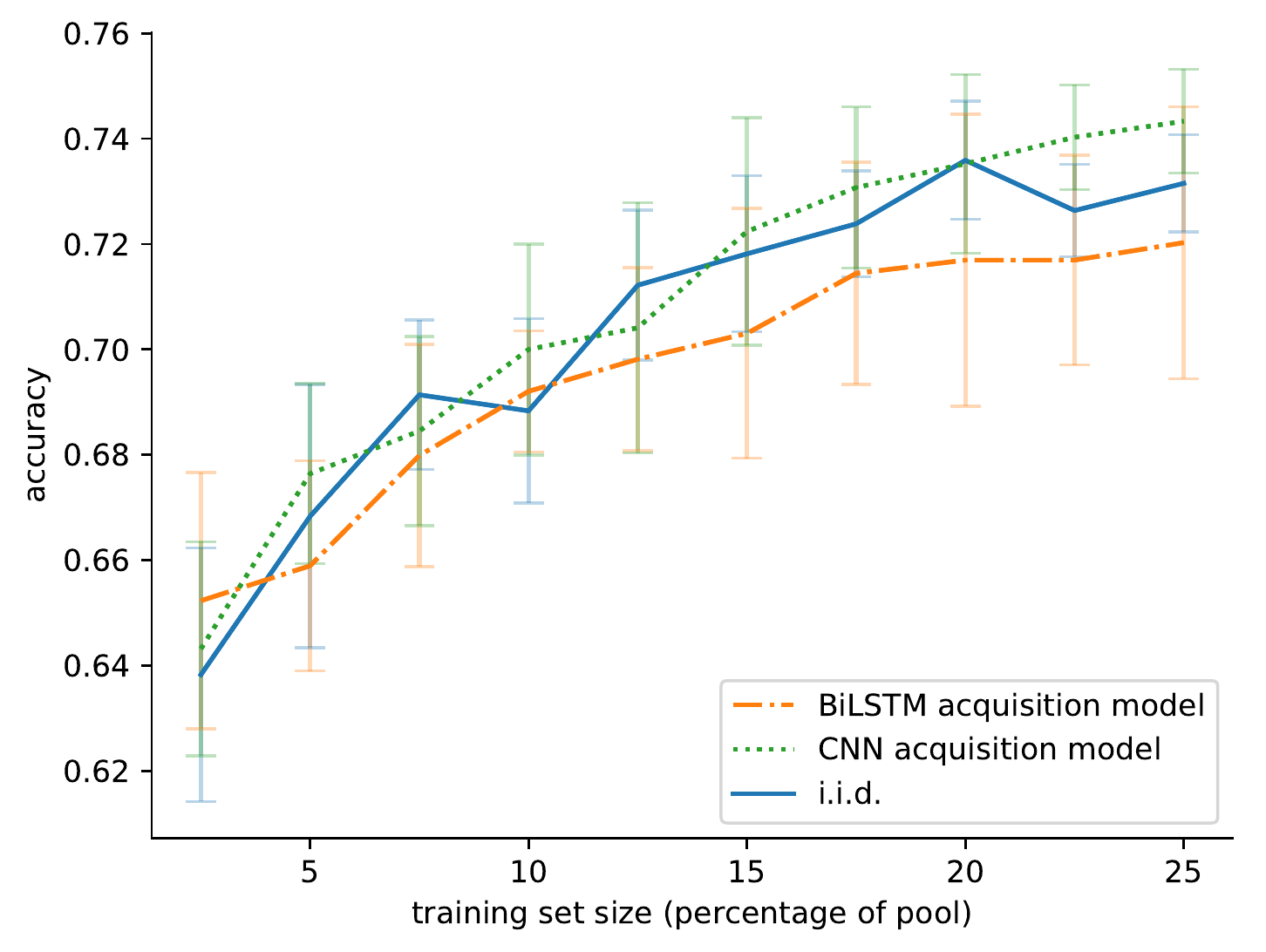}
		\caption{SVM on Customer Review dataset using BALD } % subcaption
  \end{subfigure}\\
  
  \begin{subfigure}[t]{0.44\linewidth} % width of left subfigure
		\includegraphics[width=\linewidth]{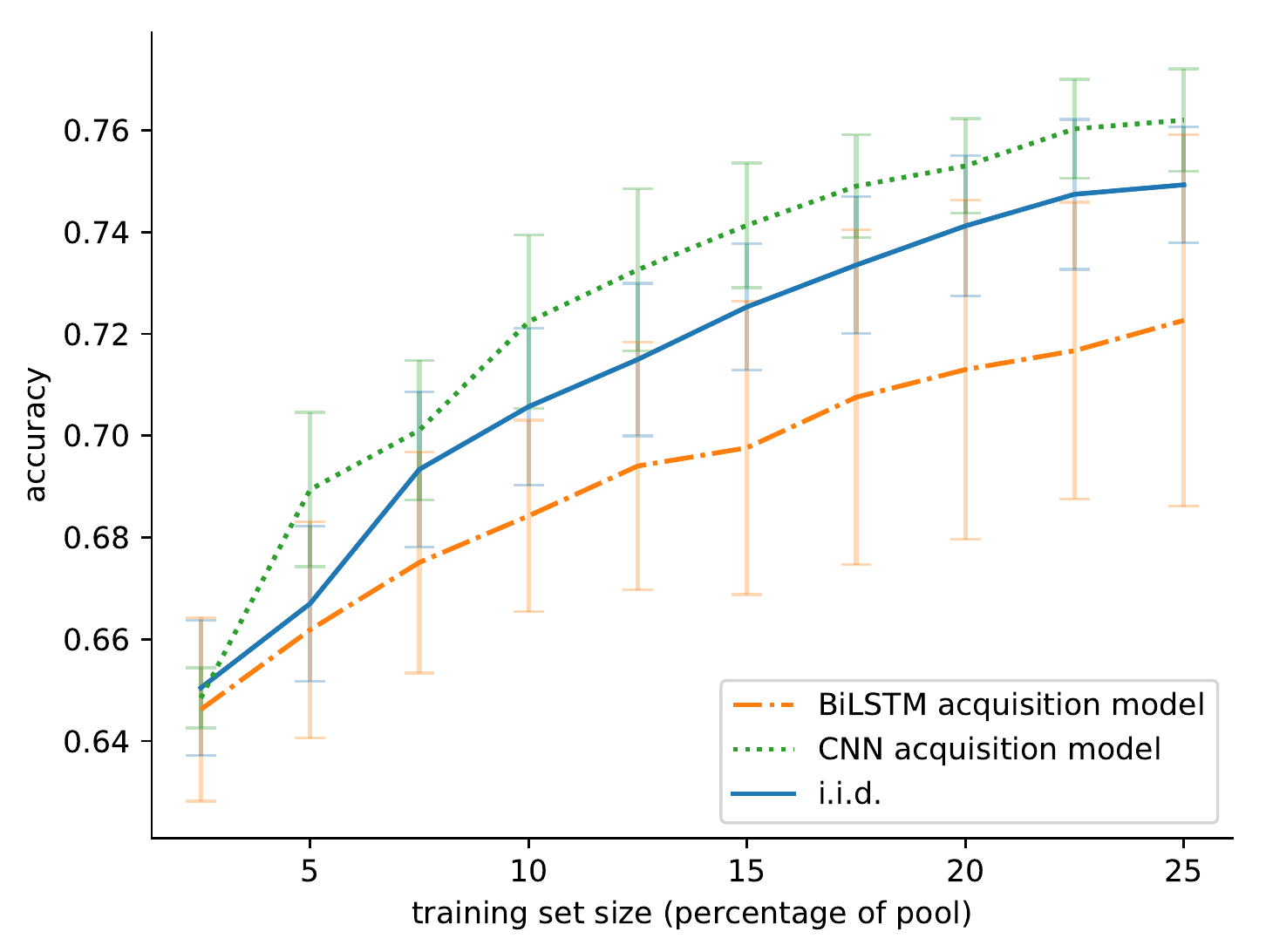}
		\caption{CNN on Customer Review dataset using BALD } % subcaption
  \end{subfigure}
  \begin{subfigure}[t]{0.44\linewidth} % width of left subfigure
		\includegraphics[width=\linewidth]{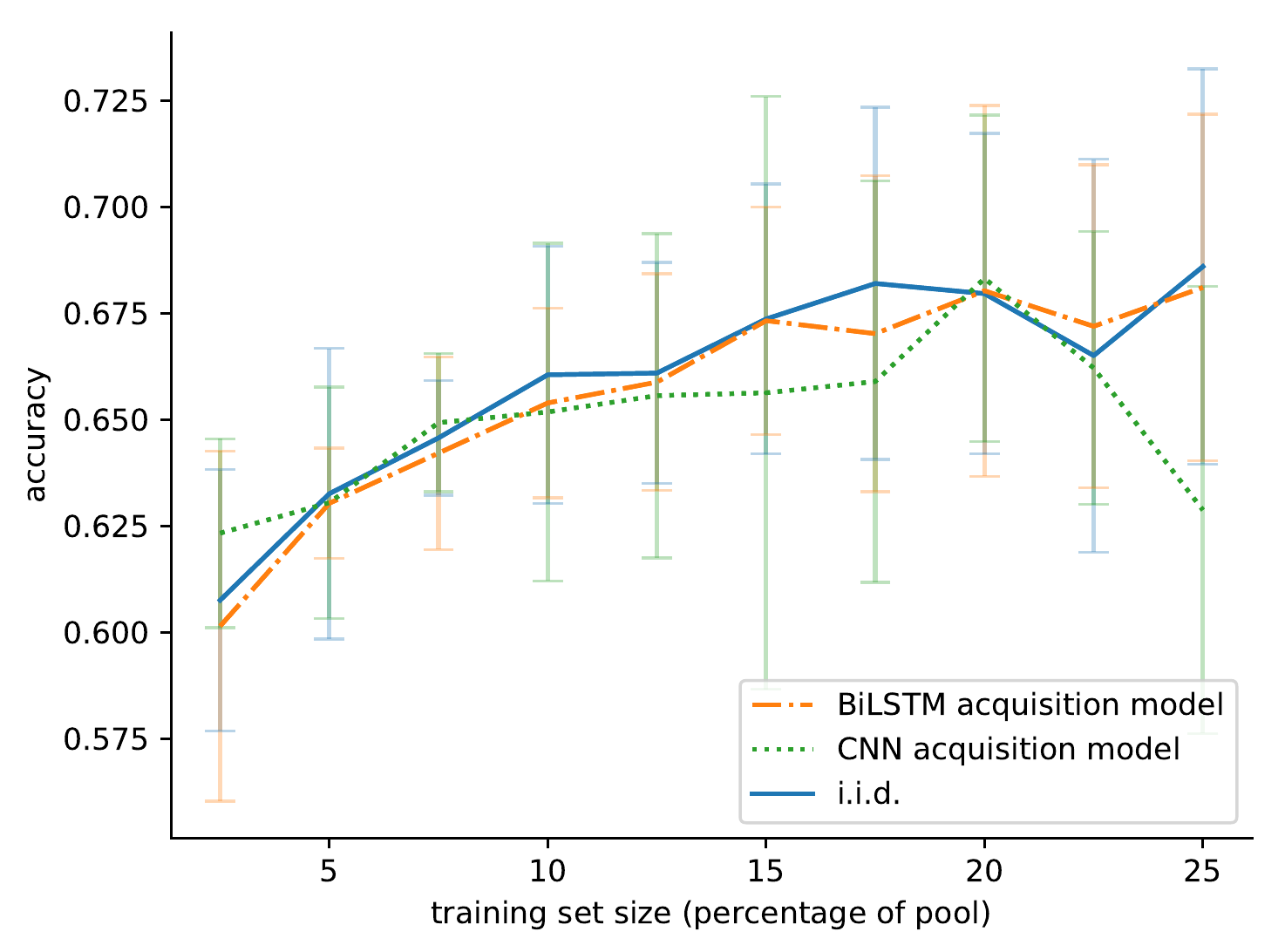}
		\caption{BiLSTM on Customer Review dataset using BALD } % subcaption
  \end{subfigure}\\
 
 \begin{subfigure}[t]{0.44\linewidth} % width of left subfigure
		\includegraphics[width=\linewidth]{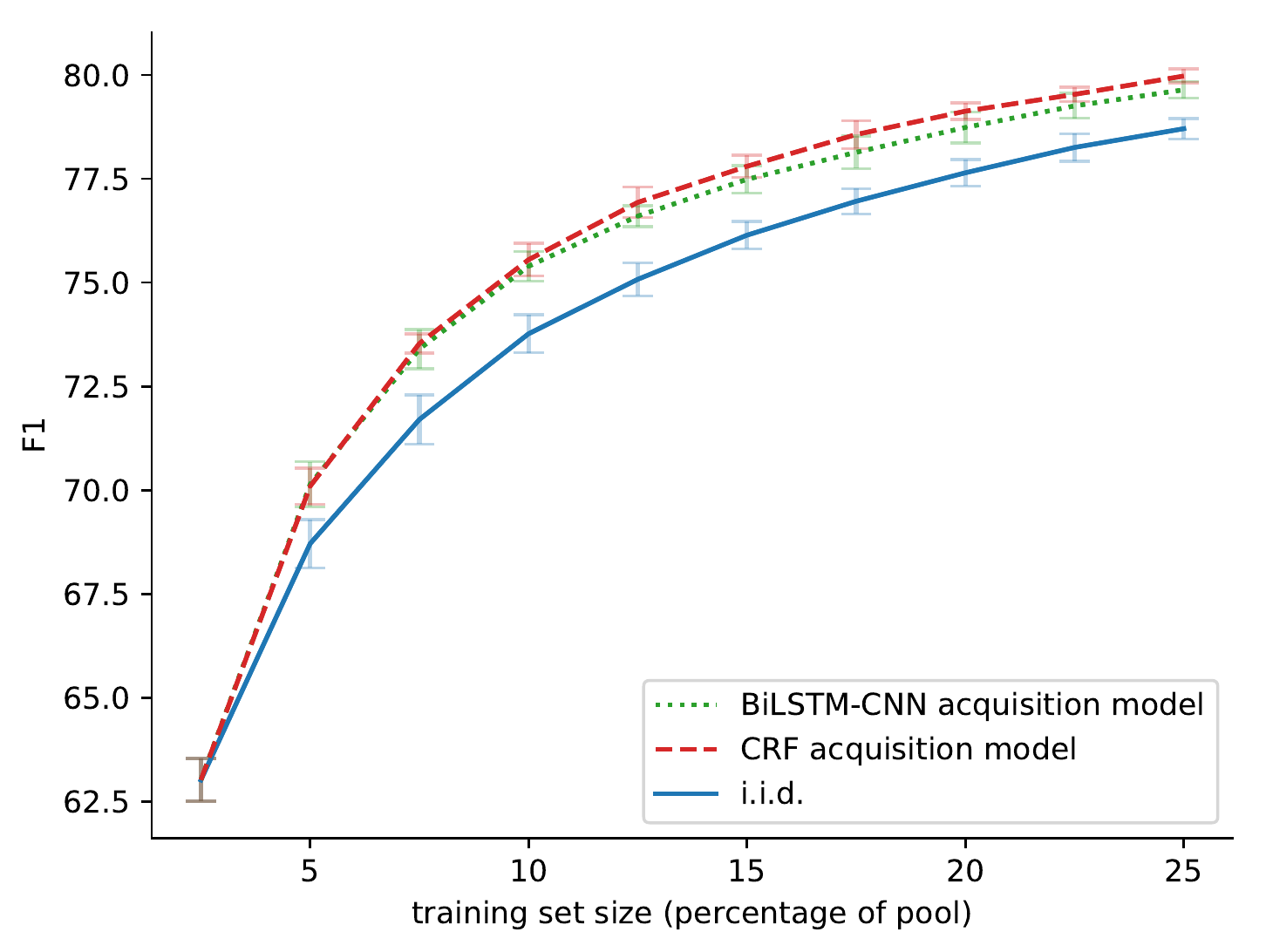}
		\caption{CRF on OntoNotes dataset using max entropy } % subcaption
  \end{subfigure}
  \begin{subfigure}[t]{0.44\linewidth} % width of left subfigure
		\includegraphics[width=\linewidth]{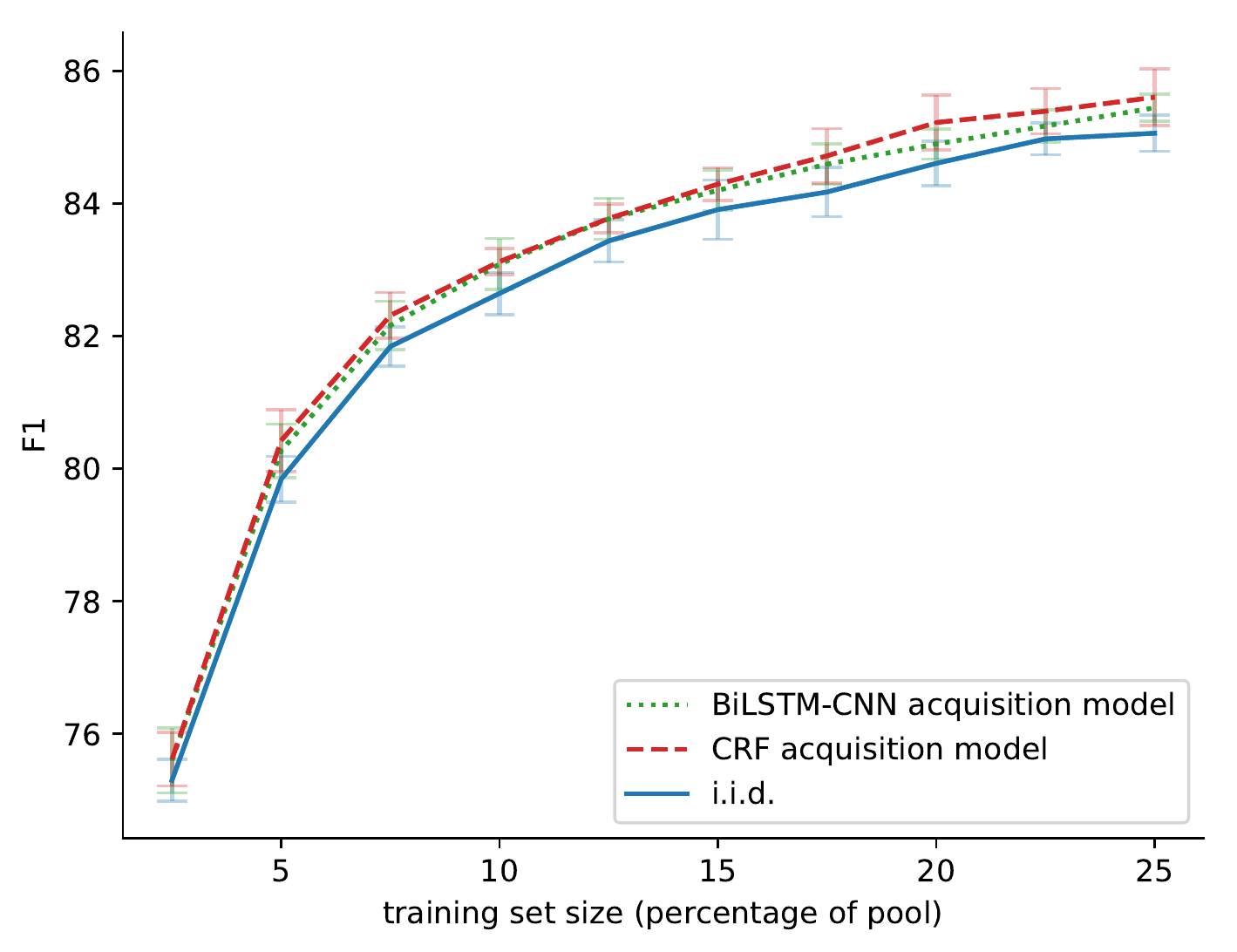}
		\caption{BiLSTM-CNN on OntoNotes dataset using max entropy } % subcaption
  \end{subfigure}\\
  
  \label{fig:appendix_pg5}
\end{figure*}

\begin{figure*}\ContinuedFloat
  \centering
  
  \begin{subfigure}[t]{0.44\linewidth} % width of left subfigure
		\includegraphics[width=\linewidth]{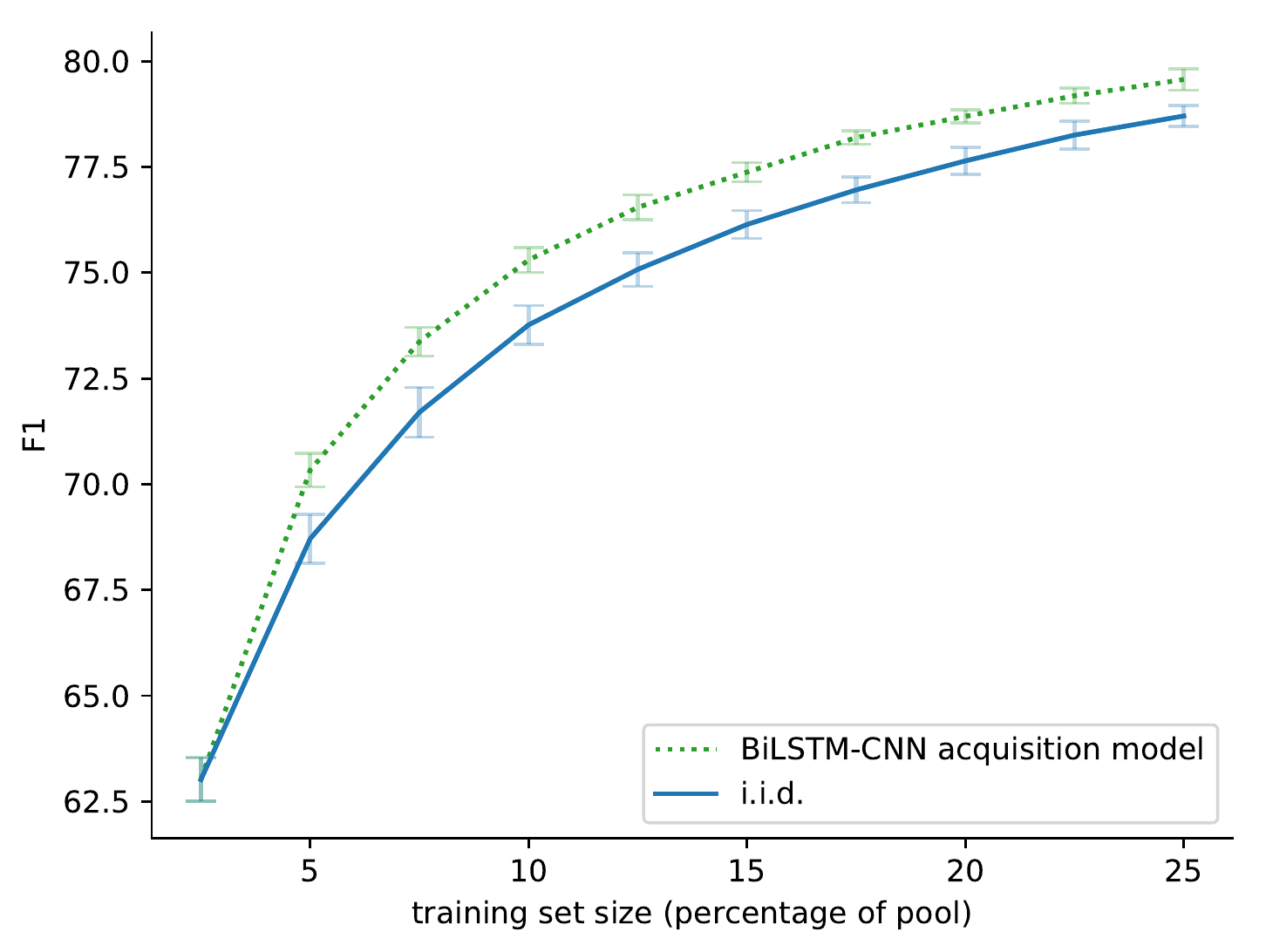}
		\caption{CRF on OntoNotes dataset using BALD } % subcaption
  \end{subfigure}
  \begin{subfigure}[t]{0.44\linewidth} % width of left subfigure
		\includegraphics[width=\linewidth]{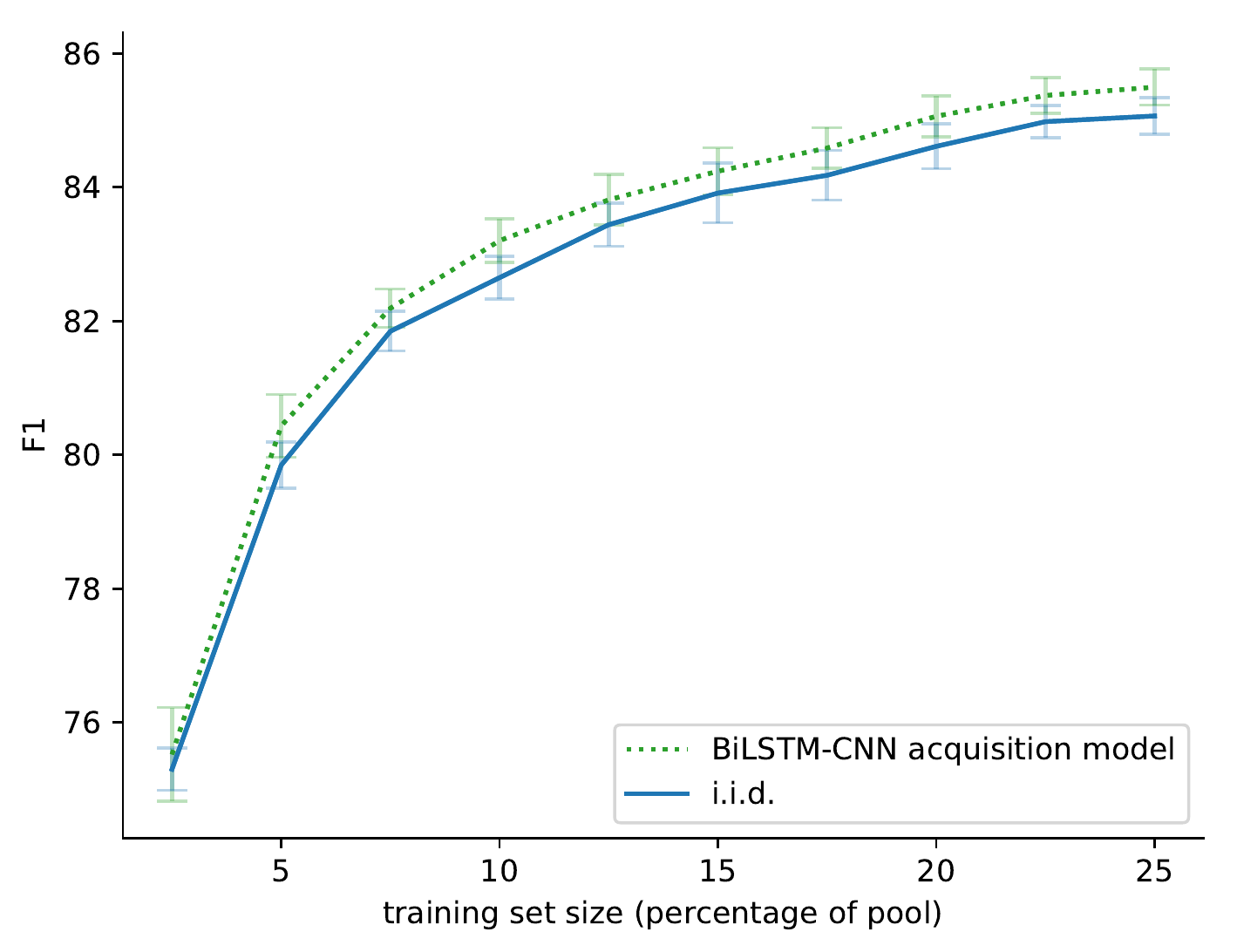}
		\caption{BiLSTM-CNN on OntoNotes dataset using BALD } % subcaption
  \end{subfigure}\\
  
  \begin{subfigure}[t]{0.44\linewidth} % width of left subfigure
		\includegraphics[width=\linewidth]{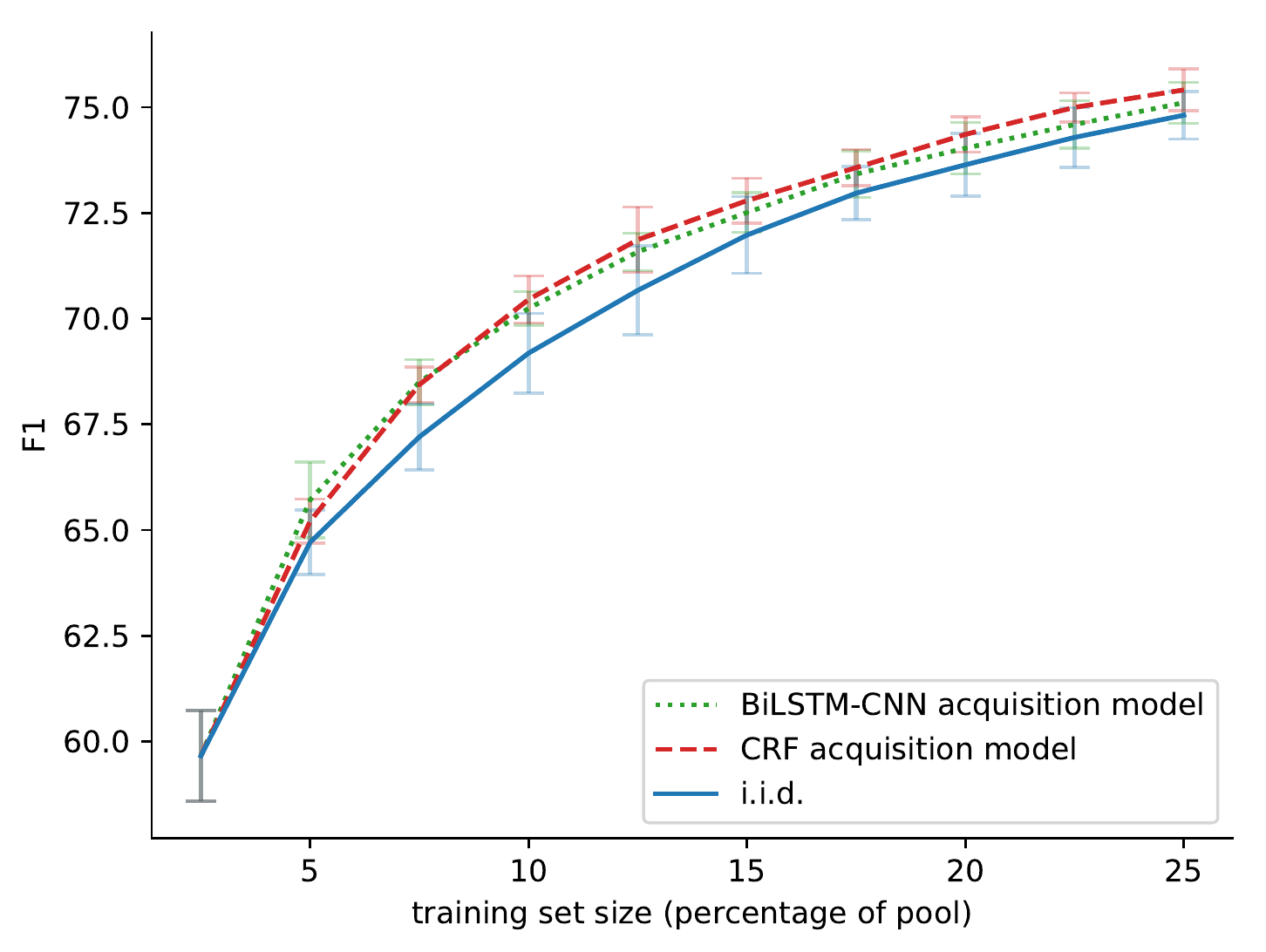}
		\caption{CRF on CoNLL dataset using max entropy } % subcaption
  \end{subfigure}
  \begin{subfigure}[t]{0.44\linewidth} % width of left subfigure
		\includegraphics[width=\linewidth]{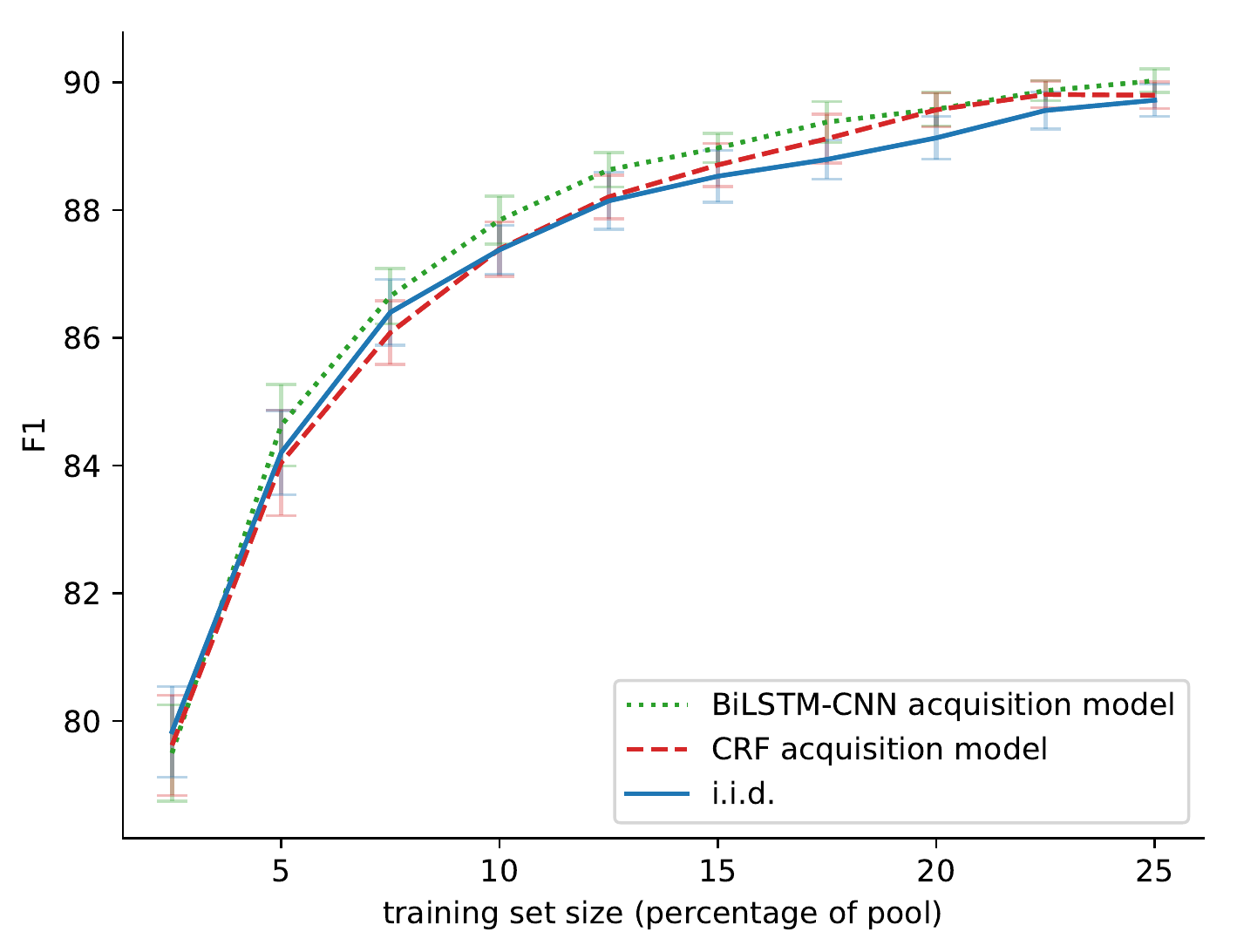}
		\caption{BiLSTM-CNN on CoNLL dataset using max entropy } % subcaption
  \end{subfigure}\\
  
  \begin{subfigure}[t]{0.44\linewidth} % width of left subfigure
		\includegraphics[width=\linewidth]{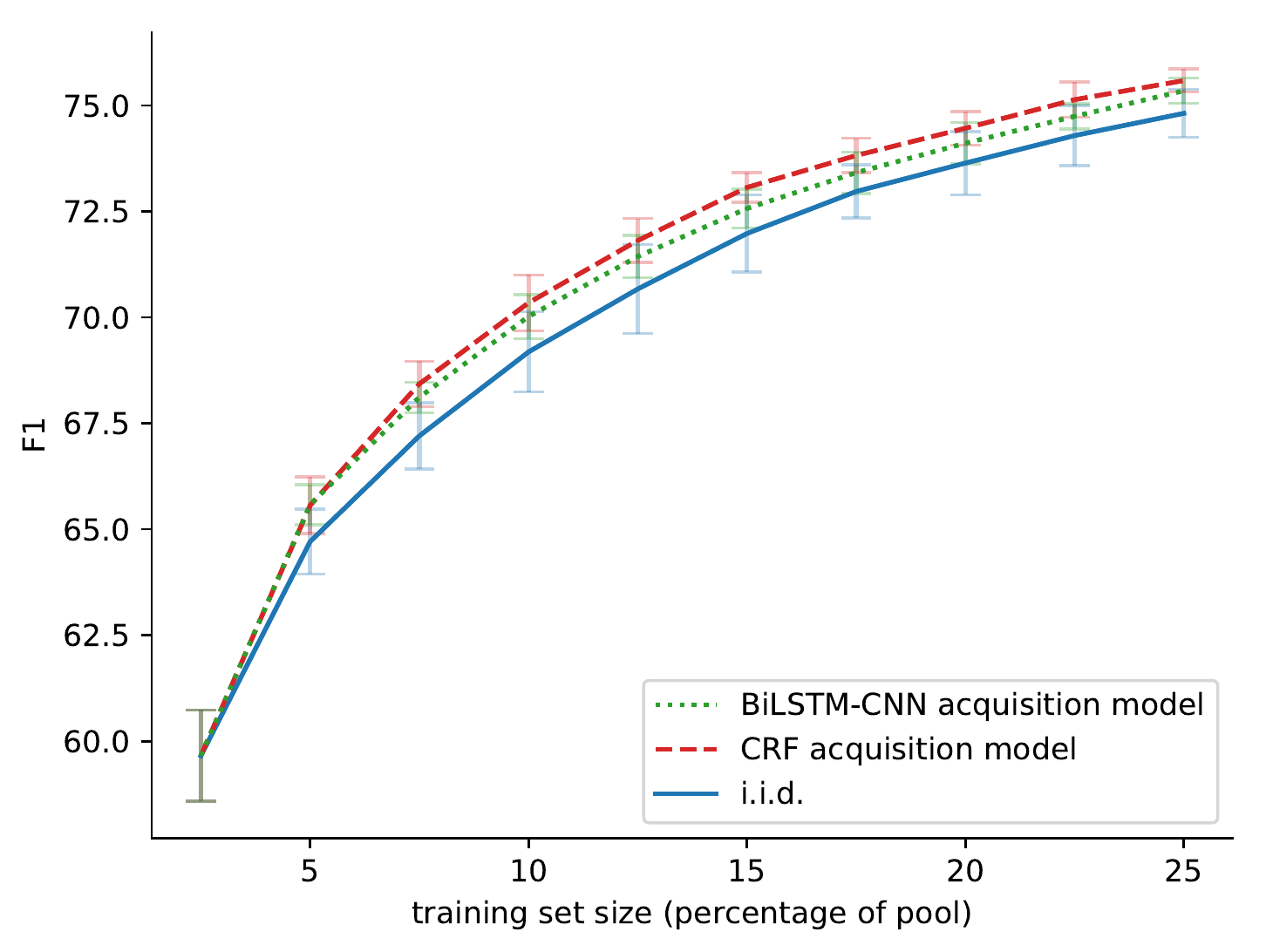}
		\caption{CRF on CoNLL dataset using QBC } % subcaption
  \end{subfigure}
  \begin{subfigure}[t]{0.44\linewidth} % width of left subfigure
		\includegraphics[width=\linewidth]{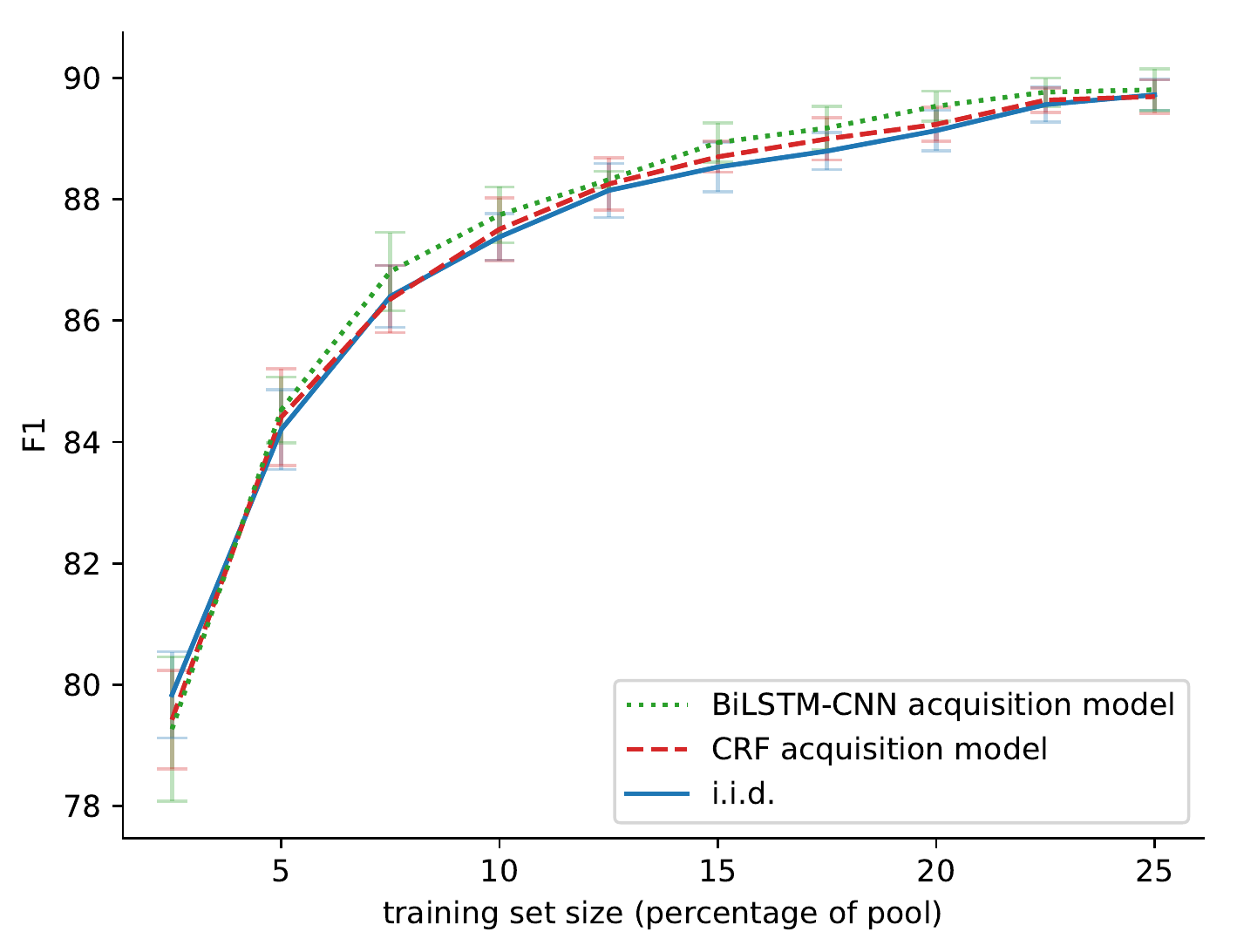}
		\caption{BiLSTM-CNN on CoNLL dataset using QBC } % subcaption
  \end{subfigure}\\
  
  \begin{subfigure}[t]{0.44\linewidth} % width of left subfigure
		\includegraphics[width=\linewidth]{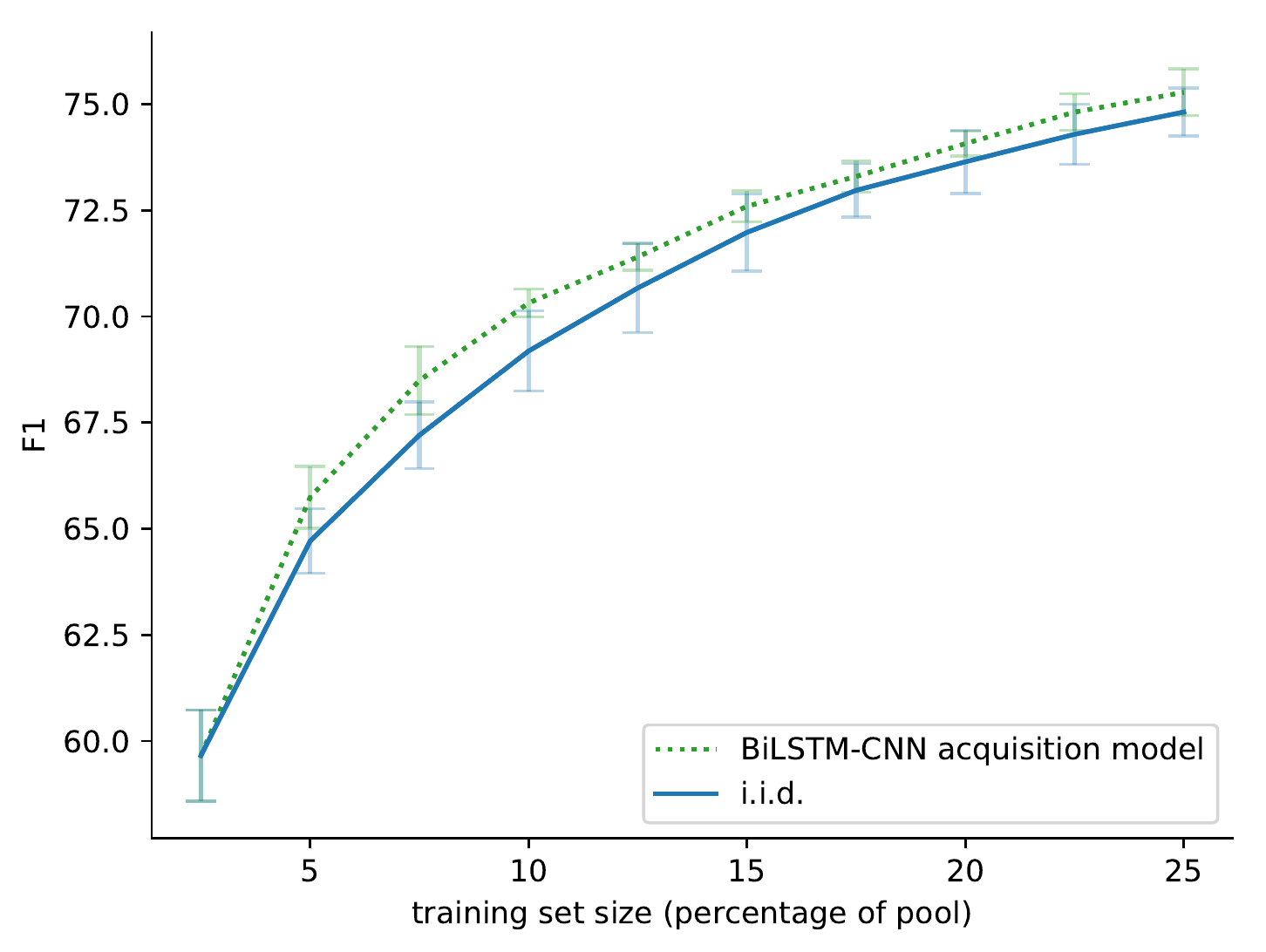}
		\caption{CRF on CoNLL dataset using BALD } % subcaption
  \end{subfigure}
  \begin{subfigure}[t]{0.44\linewidth} % width of left subfigure
		\includegraphics[width=\linewidth]{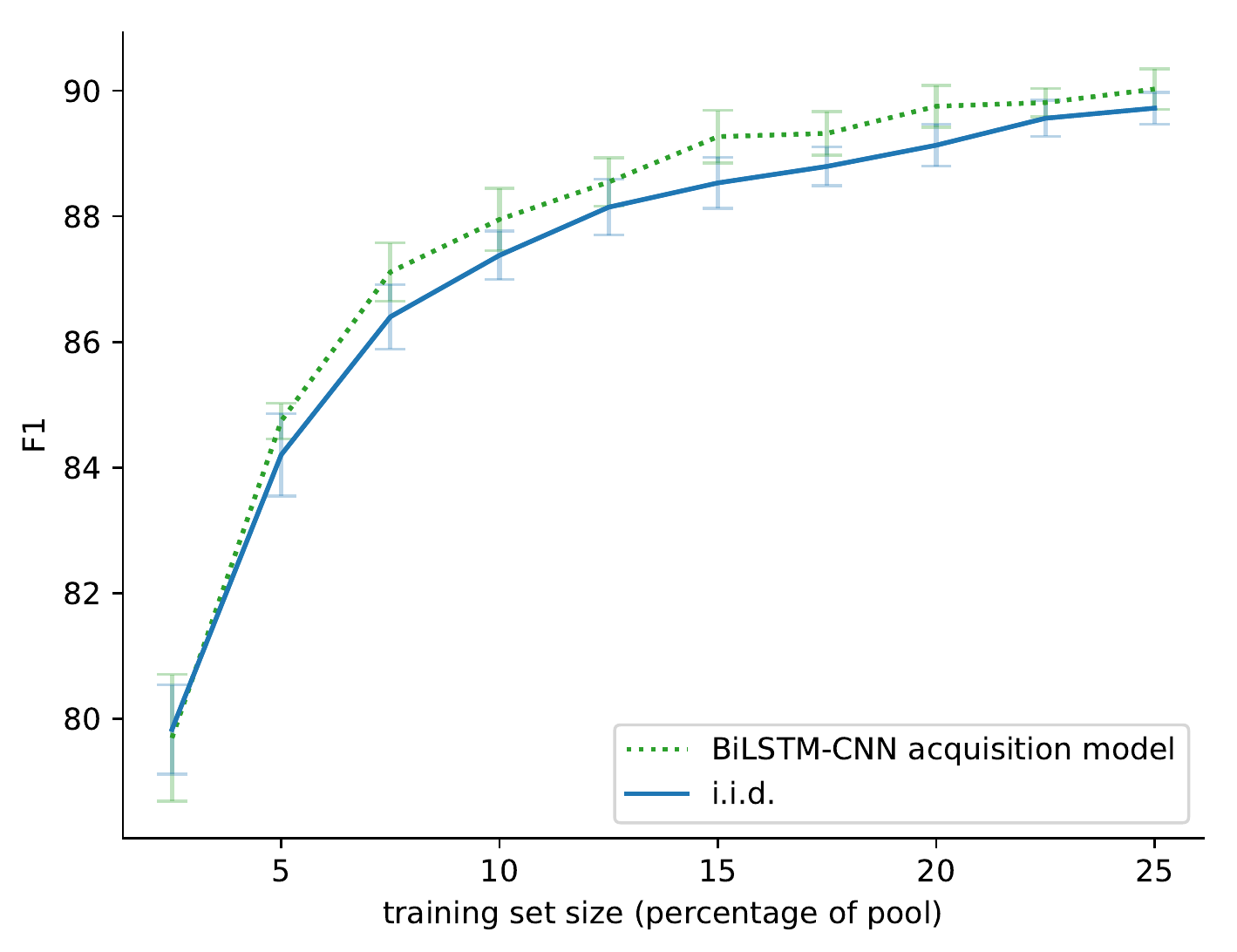}
		\caption{BiLSTM-CNN on CoNLL dataset using BALD } % subcaption
  \end{subfigure}\\
 
 \label{fig:appendix_pg6}
\end{figure*}

\begin{figure*}\ContinuedFloat
  \centering
  
  \begin{subfigure}[t]{0.44\linewidth} % width of left subfigure
		\includegraphics[width=\linewidth]{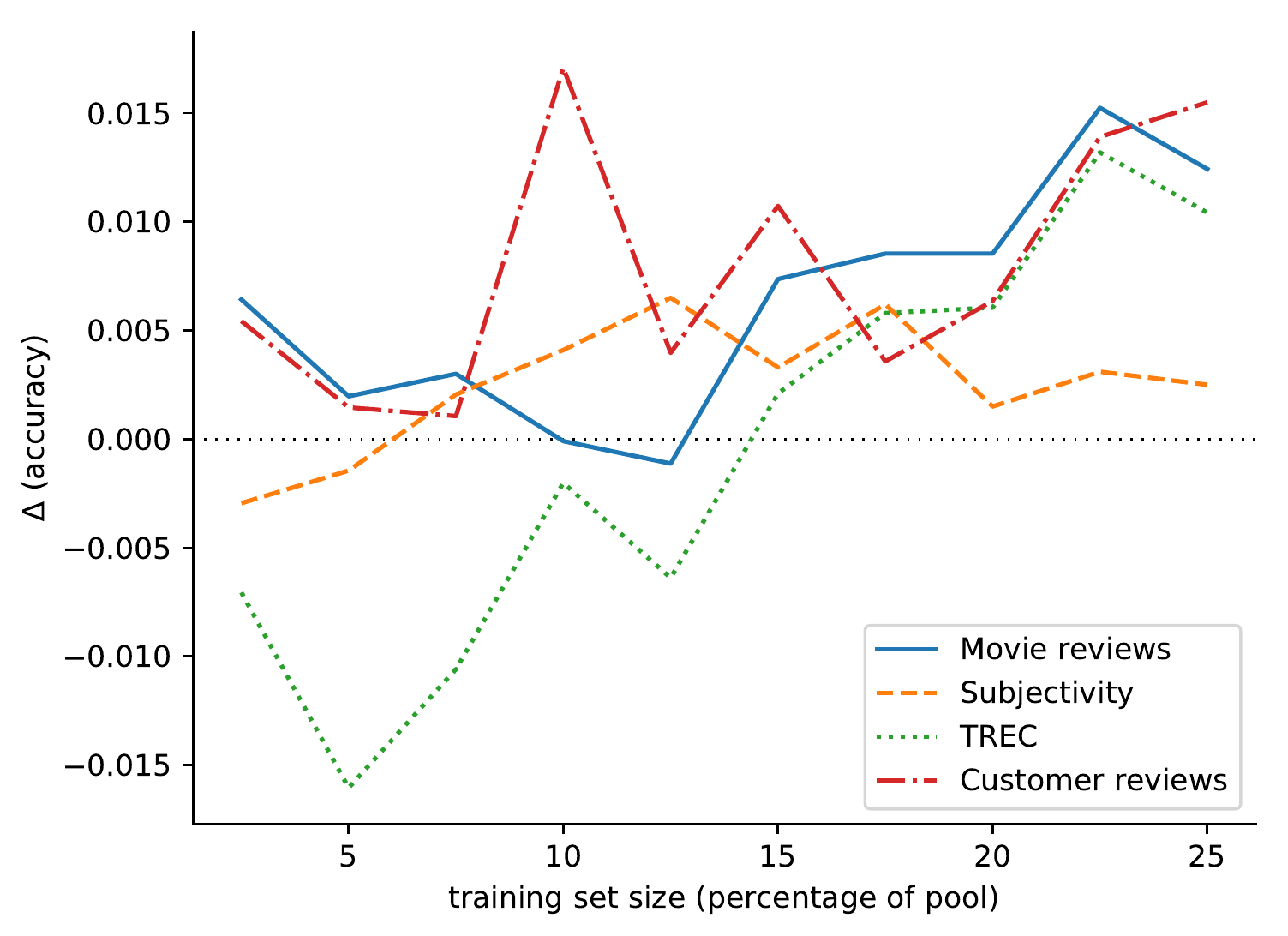}
		\caption{$\Delta$ for SVM using max entropy } % subcaption
  \end{subfigure}
  \begin{subfigure}[t]{0.44\linewidth} % width of left subfigure
		\includegraphics[width=\linewidth]{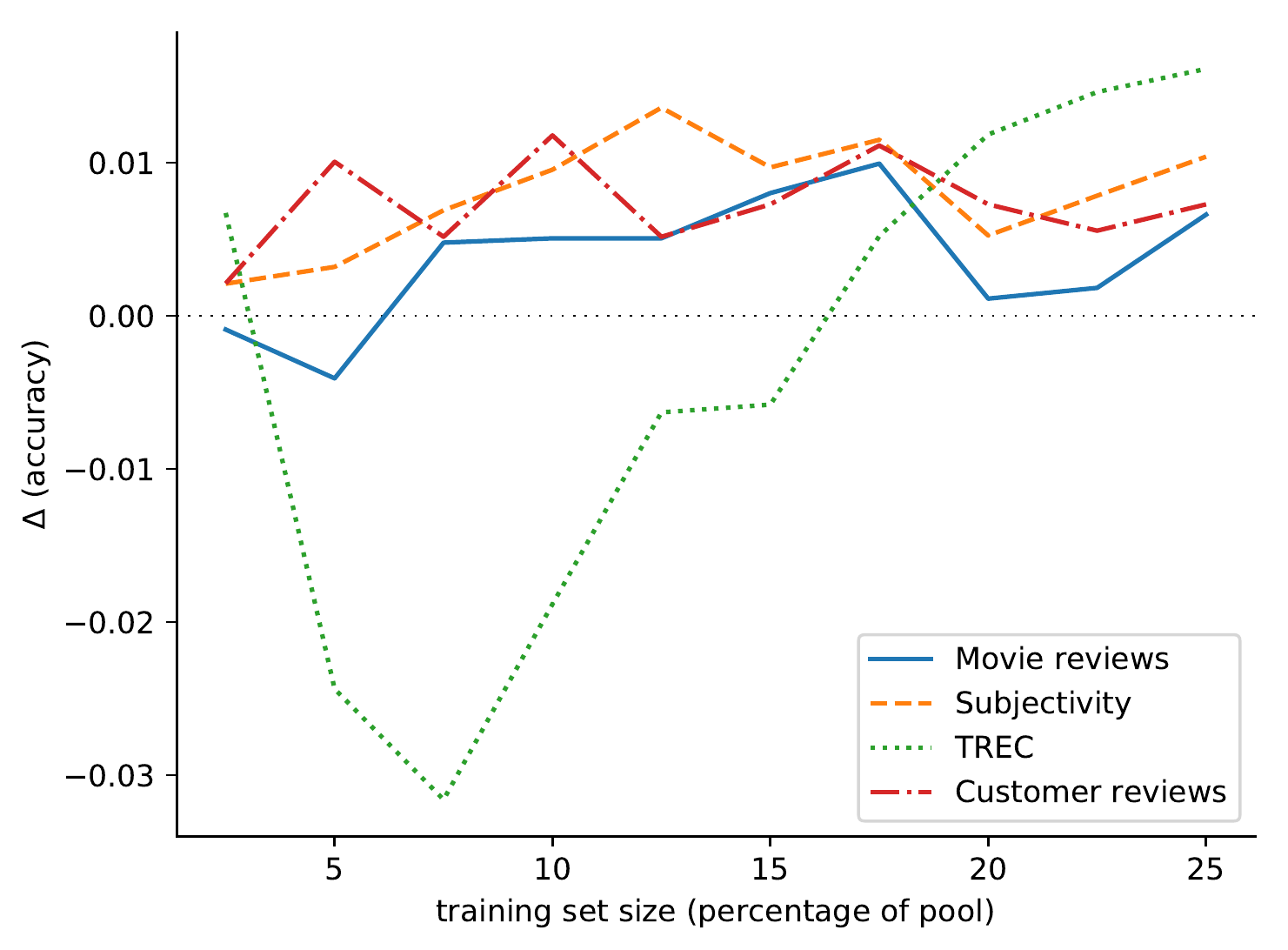}
		\caption{$\Delta$ for CNN using max entropy} % subcaption
  \end{subfigure}\\
  
  \begin{subfigure}[t]{0.44\linewidth} % width of left subfigure
		\includegraphics[width=\linewidth]{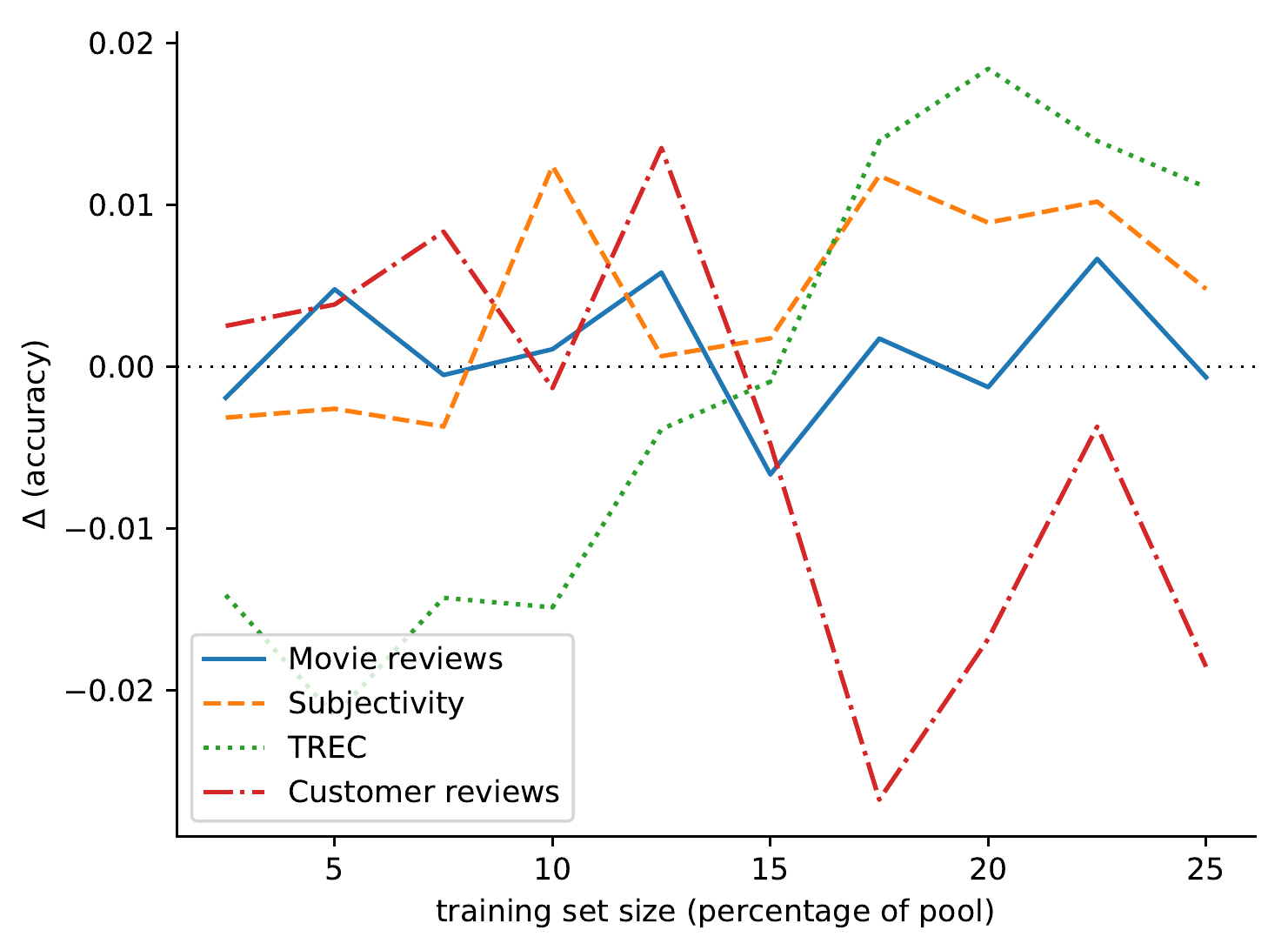}
		\caption{$\Delta$ for BiLSTM using max entropy } % subcaption
  \end{subfigure}
  \begin{subfigure}[t]{0.44\linewidth} % width of left subfigure
		\includegraphics[width=\linewidth]{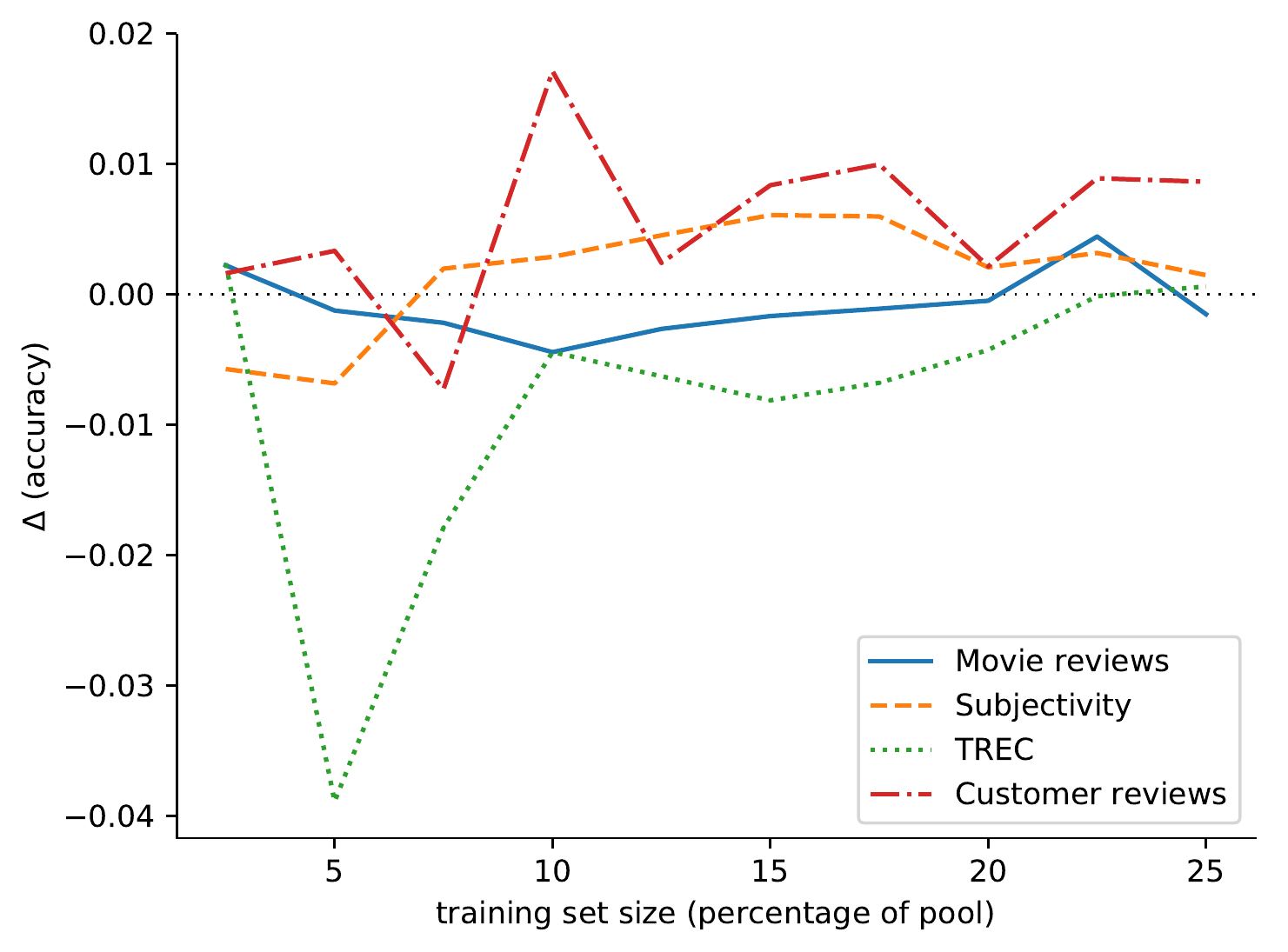}
		\caption{$\Delta$ for SVM using QBC } % subcaption
  \end{subfigure}\\
  
  \begin{subfigure}[t]{0.44\linewidth} % width of left subfigure
		\includegraphics[width=\linewidth]{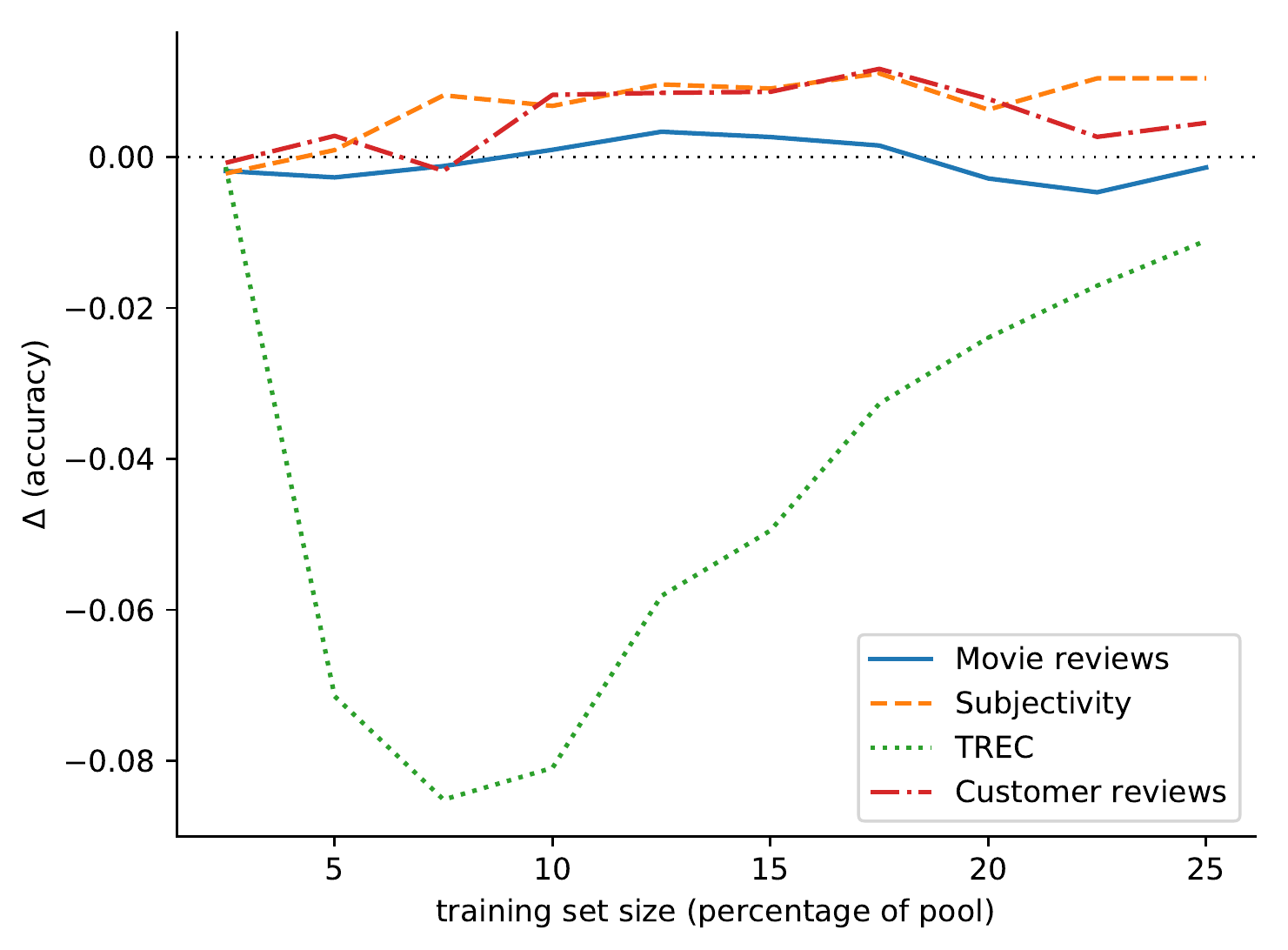}
		\caption{$\Delta$ for CNN using QBC } % subcaption
  \end{subfigure}
  \begin{subfigure}[t]{0.44\linewidth} % width of left subfigure
		\includegraphics[width=\linewidth]{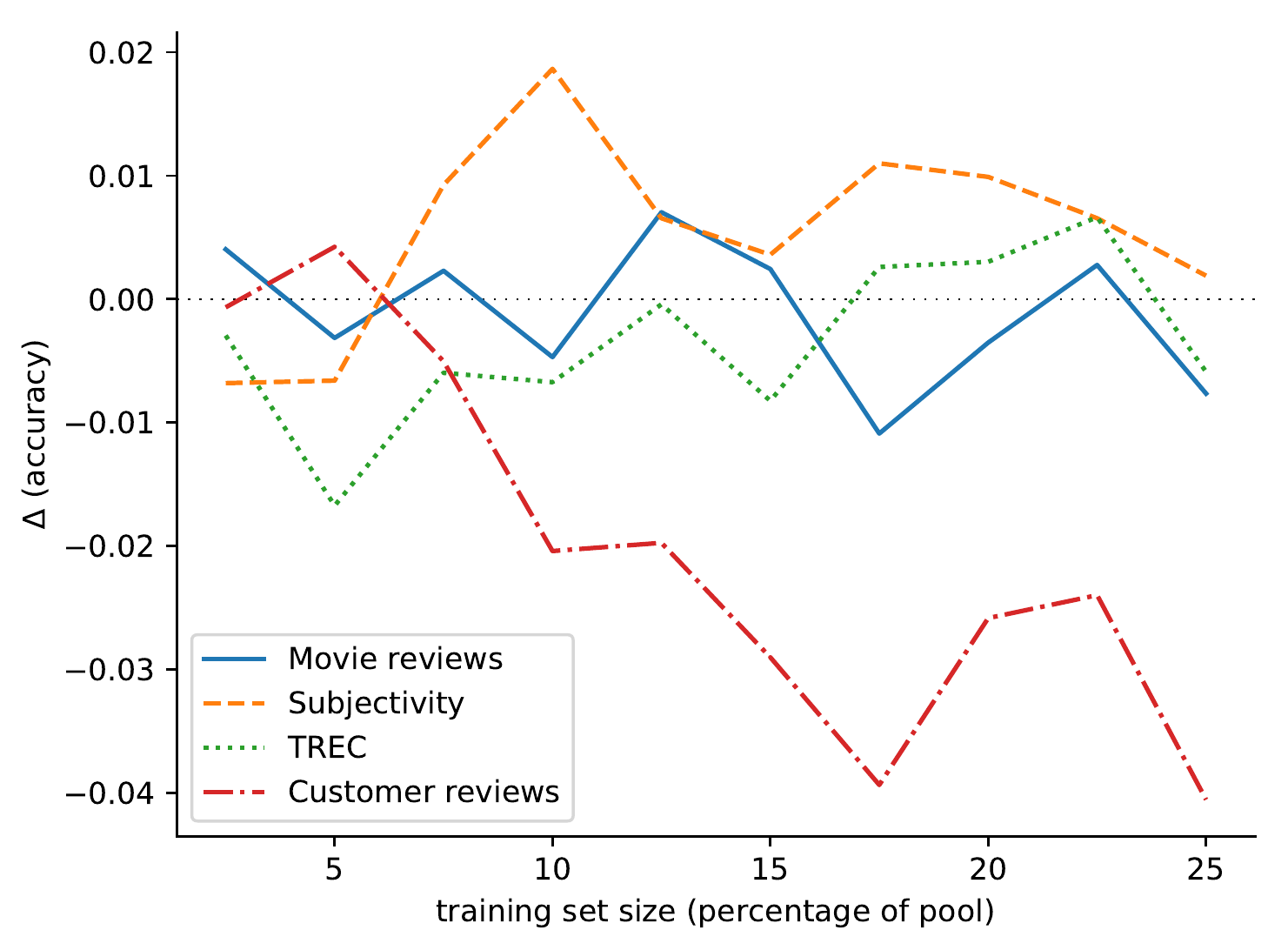}
		\caption{$\Delta$ for BiLSTM using QBC } % subcaption
  \end{subfigure}\\
  
  \begin{subfigure}[t]{0.44\linewidth} % width of left subfigure
		\includegraphics[width=\linewidth]{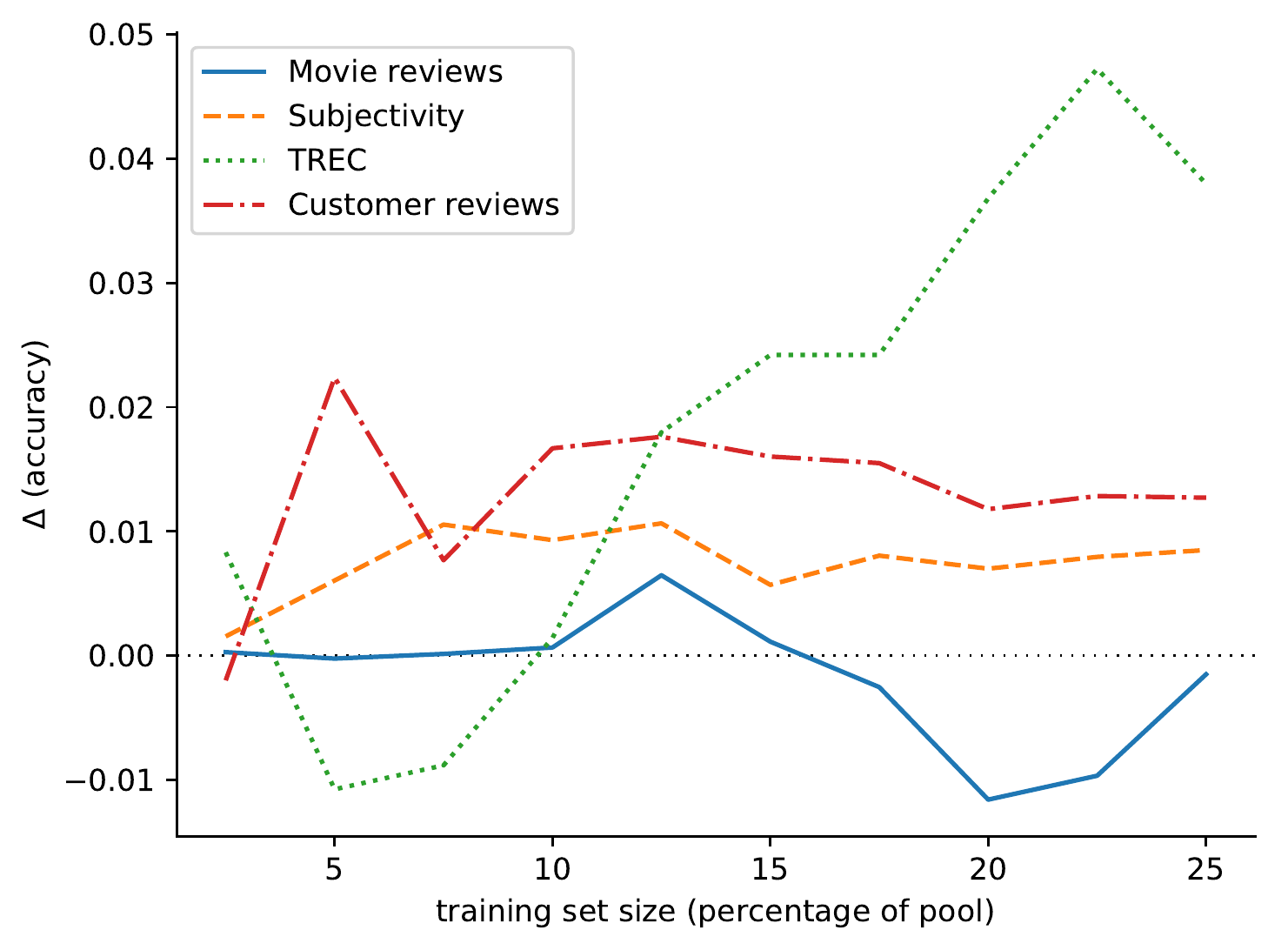}
		\caption{$\Delta$ for CNN using BALD} % subcaption
  \end{subfigure}
  \begin{subfigure}[t]{0.44\linewidth} % width of left subfigure
		\includegraphics[width=\linewidth]{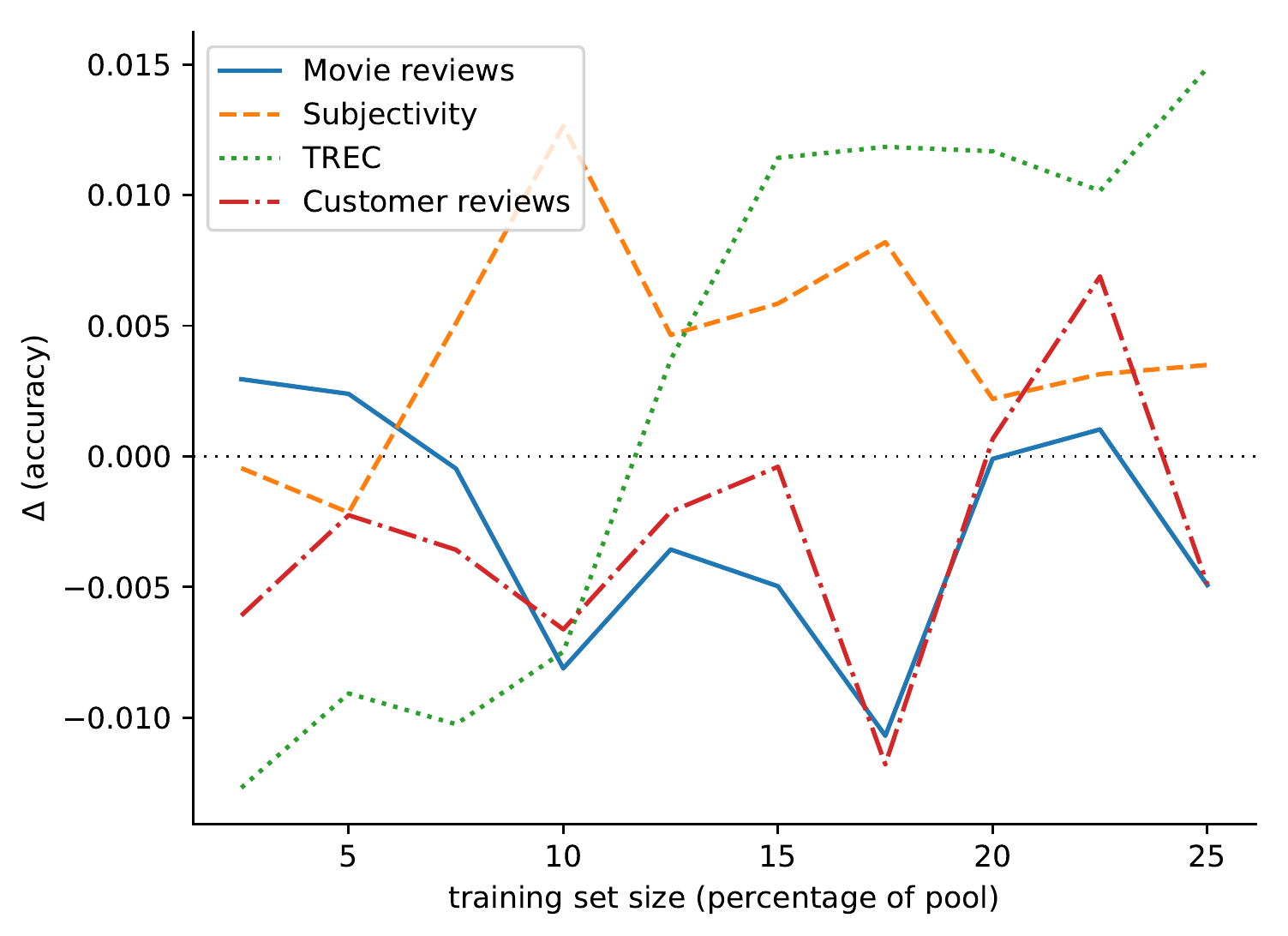}
		\caption{$\Delta$ for BiLSTM using BALD } % subcaption
  \end{subfigure}\\
 
 \label{fig:appendix_pg7}
\end{figure*}

\begin{figure*}\ContinuedFloat
  \centering
  
  \begin{subfigure}[t]{0.44\linewidth} % width of left subfigure
		\includegraphics[width=\linewidth]{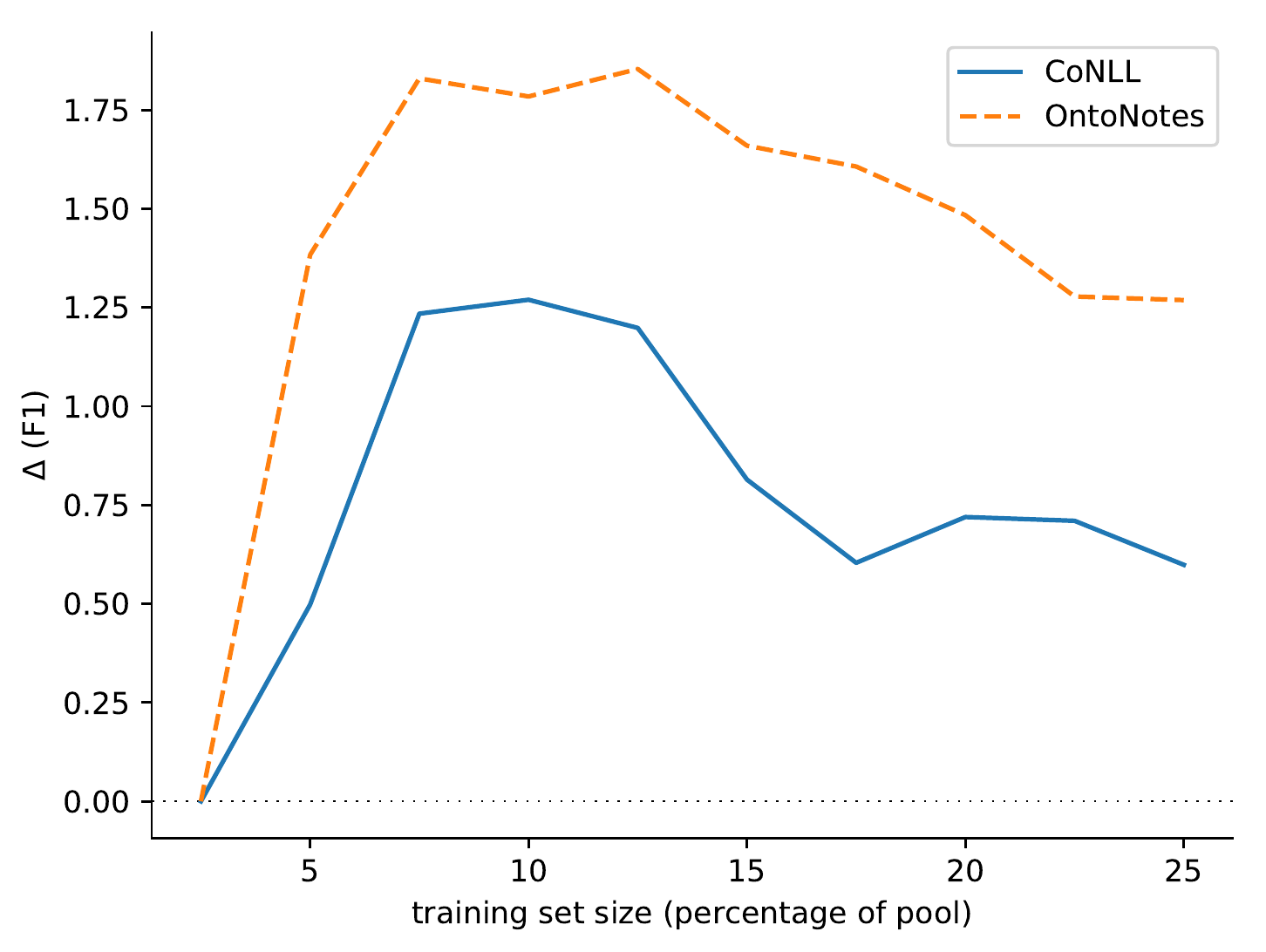}
		\caption{$\Delta$ for CRF using max entropy } % subcaption
  \end{subfigure}
  \begin{subfigure}[t]{0.44\linewidth} % width of left subfigure
		\includegraphics[width=\linewidth]{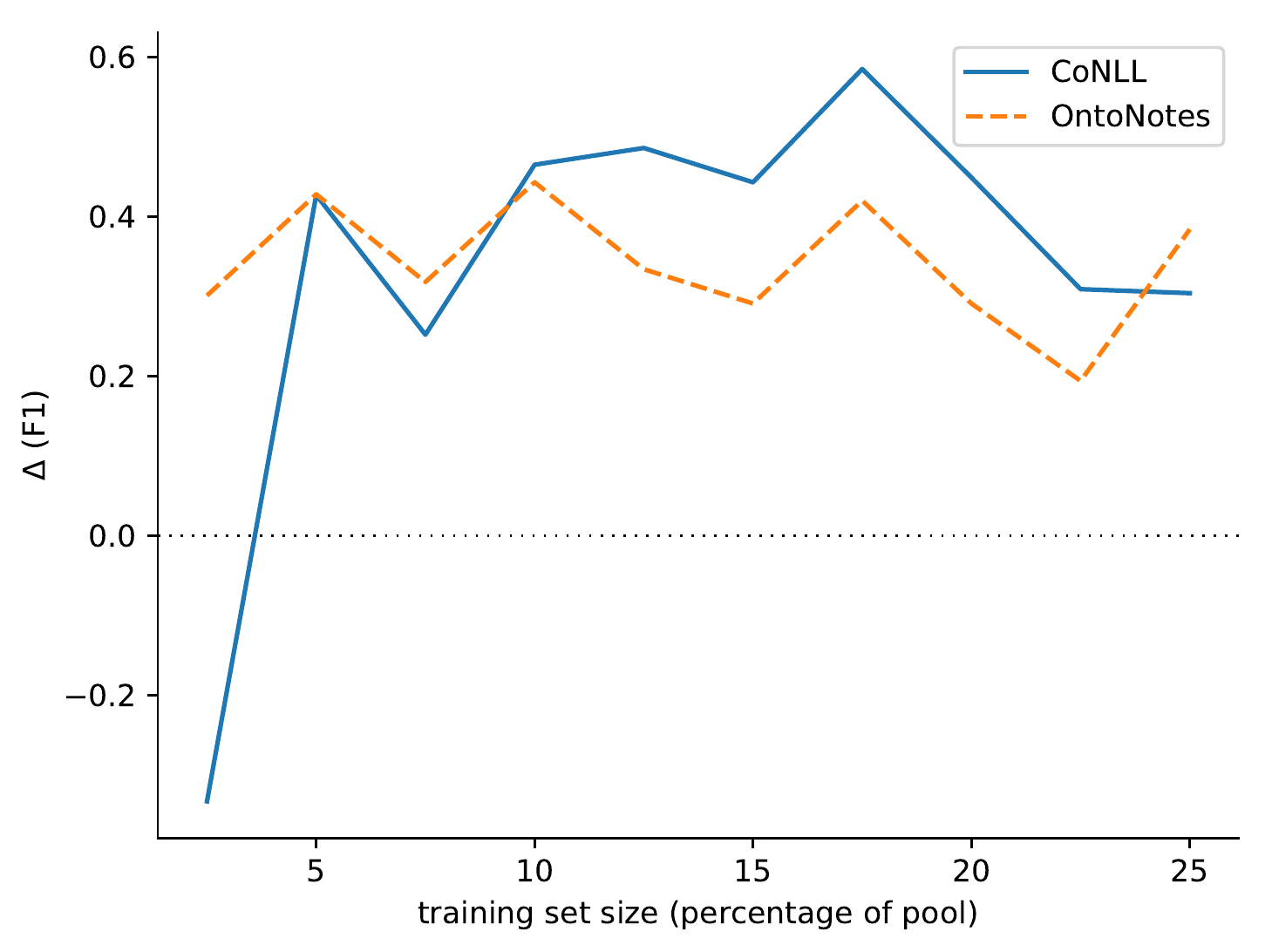}
		\caption{$\Delta$ for BiLSTM-CNN using max entropy} % subcaption
  \end{subfigure}\\
  
  \begin{subfigure}[t]{0.44\linewidth} % width of left subfigure
		\includegraphics[width=\linewidth]{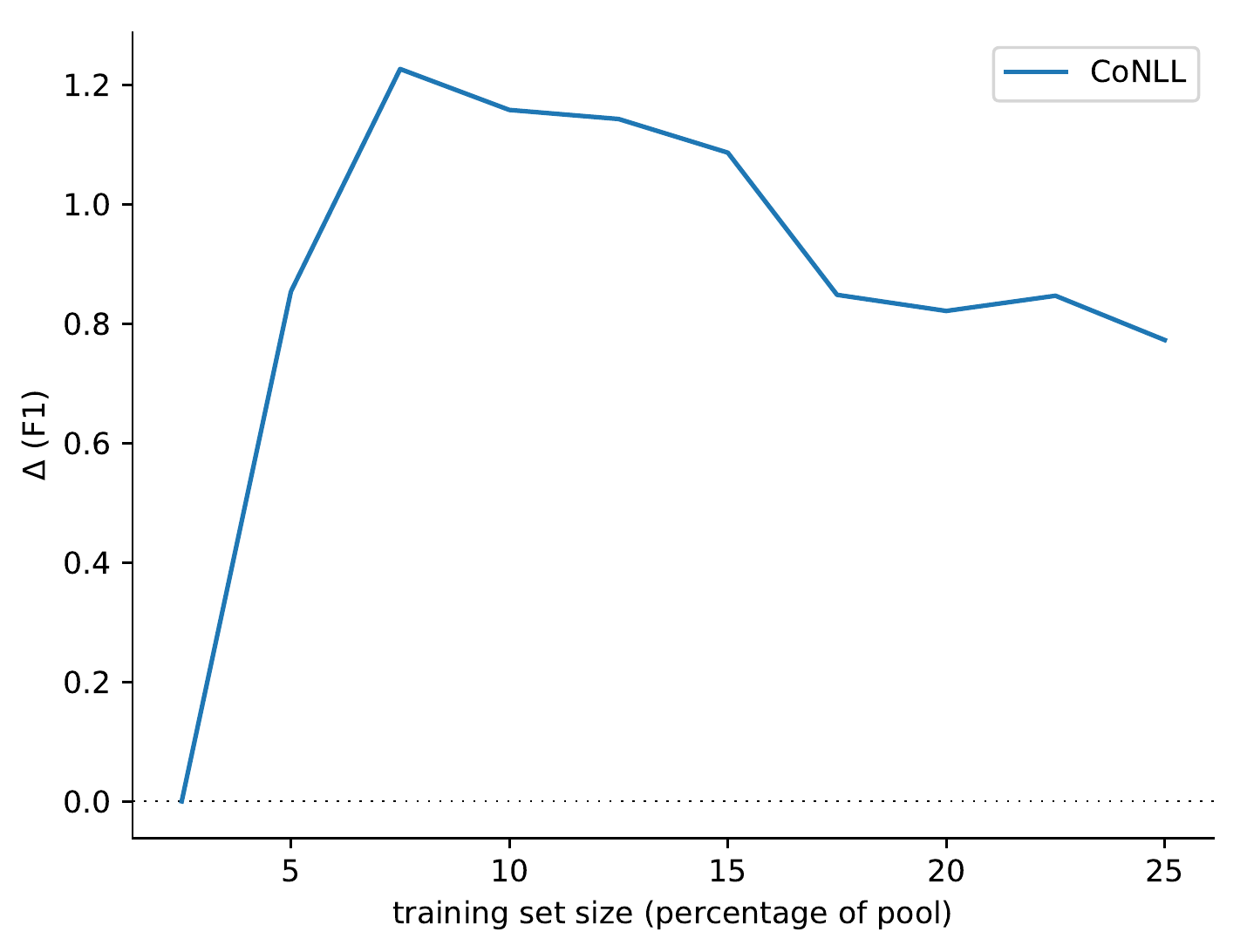}
		\caption{$\Delta$ for CRF using QBC } % subcaption
  \end{subfigure}
  \begin{subfigure}[t]{0.44\linewidth} % width of left subfigure
		\includegraphics[width=\linewidth]{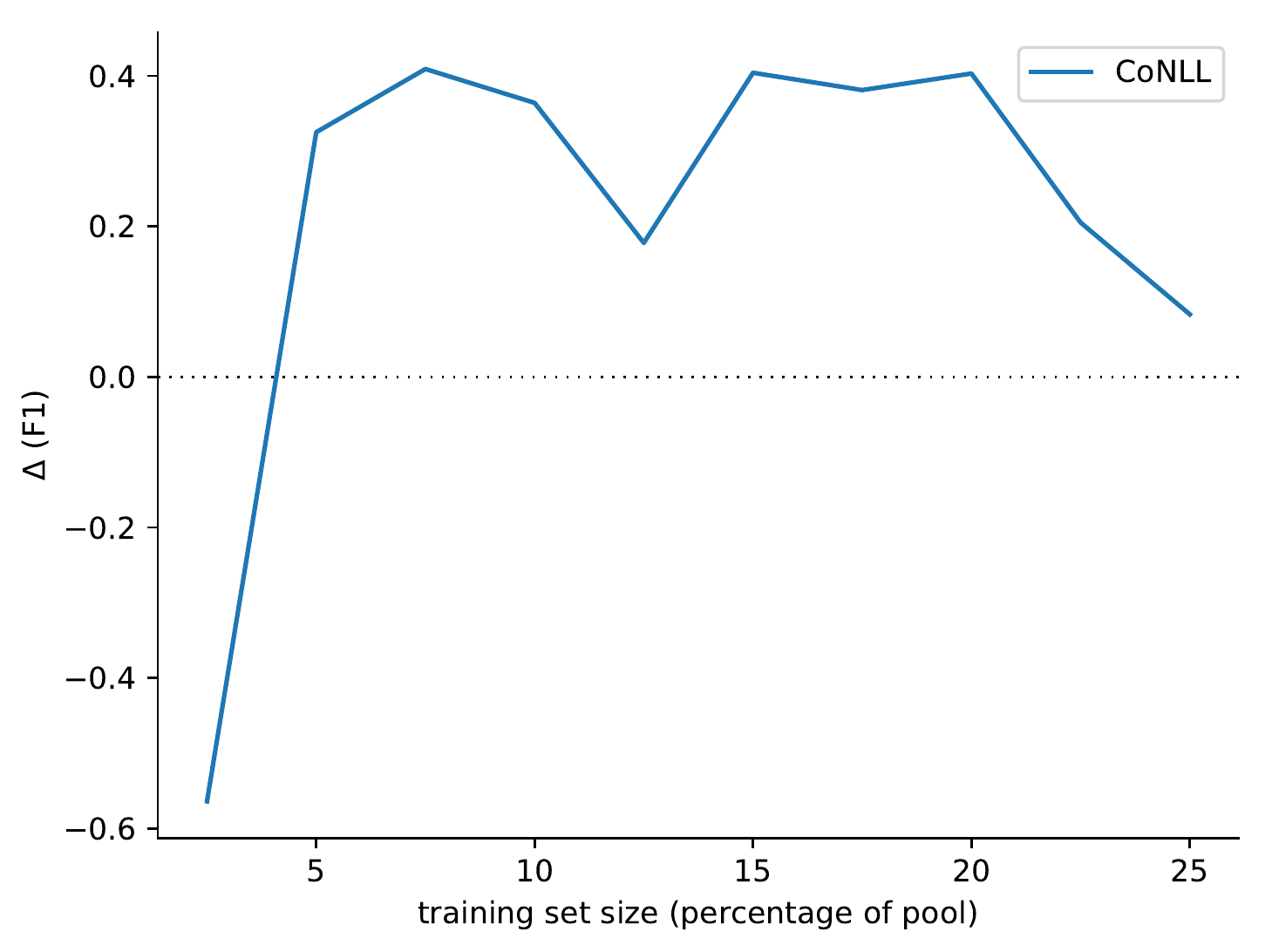}
		\caption{$\Delta$ for BiLSTM-CNN using QBC } % subcaption
  \end{subfigure}\\
  
  \begin{subfigure}[t]{0.44\linewidth} % width of left subfigure
		\includegraphics[width=\linewidth]{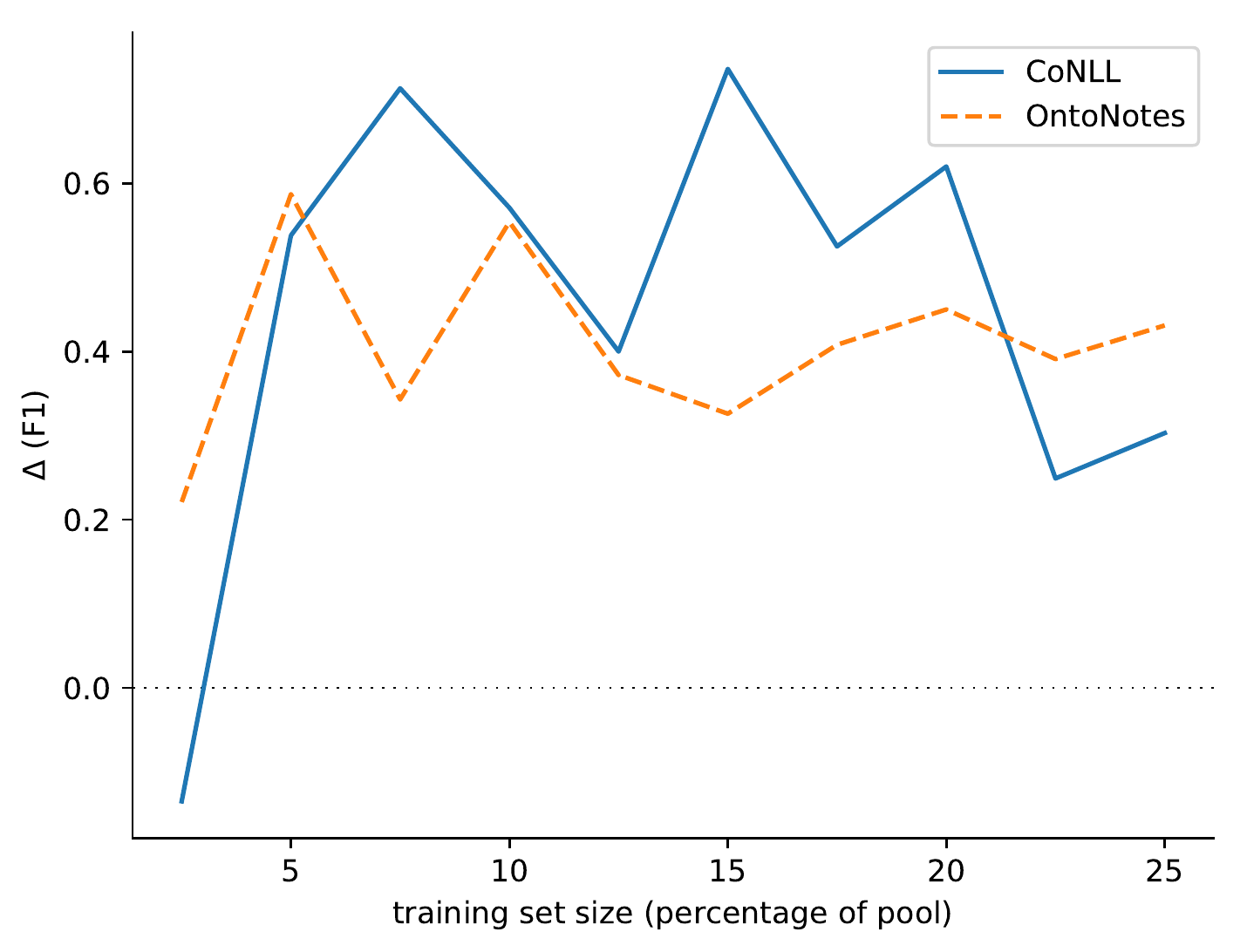}
		\caption{$\Delta$ for BiLSTM-CNN using BALD } % subcaption
  \end{subfigure}\\
 
 \label{fig:appendix_pg8}
\end{figure*}

\begin{table*}
\centering
\vspace{7px}
\begin{tabular}{ l c c c c c c c c} 
 &\multicolumn{8}{c}{Successor}\\
 &\multicolumn{2}{c}{Movie Reiews} & \multicolumn{2}{c}{Subjectivity} & \multicolumn{2}{c}{TREC} & \multicolumn{2}{c}{Customer Reviews} \\
Acquisition Model & CNN & LSTM & CNN & LSTM & CNN & LSTM & CNN & LSTM\\
\toprule
&\multicolumn{8}{c}{Uncertainty Sampling} \\
\midrule
CNN & -- & 0.961 & -- & 0.968 & -- & 0.988 & -- & 0.973 \\
LSTM & 0.989 & -- & 0.996 & -- & 0.992 & -- & 0.980 & -- \\
SVM & 0.991 & 0.961 & 0.997 & 0.970 & 0.990 & 0.987 & 0.991 & 0.974 \\
\midrule
&\multicolumn{8}{c}{QBC} \\
\midrule
CNN & -- & 0.956 & -- & 0.970 & -- & 0.985 & -- & 0.972 \\
LSTM & 0.989 & -- & 0.996 & -- & 0.990 & -- & 0.988 & -- \\
SVM & 0.995 & 0.962 & 0.997 & 0.970 & 0.985 & 0.986 & 0.993 & 0.974 \\
\midrule
&\multicolumn{8}{c}{BALD} \\
\midrule
CNN & -- & 0.963 & -- & 0.969 & -- & 0.988 & -- & 0.974 \\
LSTM & 0.991 & -- & 0.995 & -- & 0.991 & -- & 0.982 & -- \\
\bottomrule
\end{tabular}
\caption{Average Spearman's rank correlation coefficients of cosine distances between test set representations learned with native active learning and distances between those learned with transferred actively acquired datasets.}
\label{tab:spearman_appendix}
\end{table*}